\pdfoutput=1
\documentclass[11pt, a4paper, logo, onecolumn, copyright]{googledeepmind}

\usepackage[authoryear, compress, round]{natbib}
\usepackage{xspace}
\usepackage{siunitx}
\usepackage{placeins}
\usepackage{subfig}
\usepackage{booktabs}
\usepackage{tabularx}
\usepackage{amsmath}
\usepackage{multirow}
\usepackage{enumitem}
\usepackage{xurl}
\usepackage{doi}

\makeatletter
\let\orig@doi\@doi
\def\@doi#1{\orig@doi{#1}\@ifnextchar.{\@gobble}{}}
\makeatother
\usepackage[capitalise,noabbrev]{cleveref}
\crefname{appendix}{Appendix}{Appendices}
\Crefname{appendix}{Appendix}{Appendices}
\crefname{subappendix}{Appendix}{Appendices}
\Crefname{subappendix}{Appendix}{Appendices}
\crefname{subsubappendix}{Appendix}{Appendices}
\Crefname{subsubappendix}{Appendix}{Appendices}

\title{PARDIG: Global Precipitation Estimates Learned from Sparse Satellite Observations}

\author[1*]{Guy Shalev}
\author[1*]{Stephan Rasp}
\author[2*]{Boris Babenko}
\author[1]{Andrew El-Kadi}
\author[1]{Mohammed Alewi Hassen}
\author[1]{Samier Merchant}
\author[2]{Fred Zyda}
\author[1]{Dominic Masters}
\author[1]{Ferran Alet}
\author[2]{Salva R\"uhling Cachay}
\author[1]{Ilan Price}
\author[2]{Akib Uddin}
\author[1]{Nofar Peled Levi}
\author[1]{Ben Gaiarin}
\author[2]{Aaron Bell}
\author[1]{Peter Battaglia}
\author[1*]{Alvaro Sanchez-Gonzalez}

\affil[*]{Equal contribution}
\affil[1]{Google DeepMind}
\affil[2]{Google Research}

\correspondingauthor{{}\{guysha, srasp, alvarosg\}@google.com}

\newcommand{\pardig}{PARDIG\xspace}
\newcommand{\pardignrt}{PARDIG-NRT\xspace}
\newcommand{\pardiglr}{PARDIG-LR\xspace}
\newcommand{\tenthdegree}{\ang{0.1}\xspace}
\newcommand{\quarterdegree}{\ang{0.25}\xspace}

\begin{abstract}
Estimating precipitation at the global scale is vital for Earth system science and
disaster response, yet observational coverage and measurement precision remain constrained.
Existing solutions provide options that trade off desirable traits such as skill, latency,
geographical coverage, temporal coverage, and spatial and temporal resolution. We present \pardig,
an artificial intelligence framework based on a spherical graph-mesh transformer architecture that
produces near-real-time global precipitation estimates. \pardig is trained on dense multi-modal
inputs, including atmospheric state analyses and multi-spectral geostationary satellite imagery, to
predict sparse, high-precision precipitation measurements from spaceborne radar. At inference, the
trained model is applied densely across the globe at \tenthdegree spatial and 15-minute temporal
resolution, producing complete near-real-time precipitation fields from 2016 to the present,
alongside a historical variant extending the record back to 1999.
Evaluation against three independent radar networks (over the US, Europe, and Japan) and against
over 36,000 surface stations worldwide shows that \pardig consistently outperforms IMERG~V07
(Final and Early), MSWEP~V3, ERA5, and HRES across precipitation thresholds and regions, with
the largest improvements at extreme intensities. Case studies of extreme precipitation events
demonstrate that \pardig successfully captures their spatial structure and timing.
The improvements also extend to forecasting: WeatherNext~3, a global weather model, produces
more skillful precipitation forecasts when trained to predict \pardig. Both the historical and
near-real-time \pardig datasets are openly available under a CC~BY~4.0 license.
\end{abstract}

\begin{document}

\maketitle

\section{Introduction}
\label{sec:intro}

Precipitation plays a central role in the Earth system, with far-reaching ecological and societal impacts. It serves as a primary driver for hydrological modeling, agricultural planning, and hazard management. Reliable observation records are therefore essential, yet the episodic and spatially heterogeneous nature of precipitation makes continuous monitoring across the globe exceptionally challenging. Rain gauges provide high-quality observations but are sparse and unequally distributed. A recent study found that only 13.4\% of the global land surface meets the World Meteorological Organization's requirements for annual precipitation
monitoring, with severe gaps across oceans, developing regions, and high latitudes
\citep{su2026precipitation}. Further, many rain gauges measure only daily accumulations, omitting important information about the diurnal cycle and sub-daily extremes. Ground-based weather radars provide yet another popular source of precipitation data: their active sensing delivers high-resolution, low-latency observations for a wide range of applications. However, radars are expensive to set up and maintain, leaving dense, continuous network coverage largely confined to a handful of developed regions and absent over oceans \citep{saltikoff2019overview}. These gaps and uncertainties in precipitation observations are widely recognized as a major bottleneck limiting streamflow simulation and flood prediction accuracy \citep{kavetski2006bayesian, beck2017global, kratzert2021note}.
More broadly, closing these coverage gaps is essential for disaster
risk reduction, climate adaptation, and the improvement of weather
forecast generation and validation.

The community has developed a rich landscape of global precipitation products to
address these gaps, broadly falling into four categories.

\noindent\textbf{Gauge-only products} such as GPCC \citep{schneider2014gpcc} and CPC Unified \citep{chen2008assessing} interpolate station records onto global land grids, offering high accuracy where gauges are dense but inheriting the spatial gaps, ocean absence, and daily resolution of the underlying networks.

\noindent\textbf{Satellite-based products} use passive spaceborne observations from geostationary satellites or low-Earth orbiting microwave radiometers to build global, or near-global, precipitation estimates. The most prominent among those is IMERG \citep{huffman2020imerg, huffman2023imerg} but many others exist \citep{hsu1997persiann, joyce2004cmorph, aonashi2009gsmap, nguyen2020persiann}. While widely used, traditional operational products often struggle to match ground-based observations: methods that primarily rely on passive microwave retrievals like IMERG, for example, require morphing and interpolation algorithms to fill gaps between satellite overpasses, which can smooth fine-scale structure and degrade sub-daily performance \citep{li2023imerg, wang2025hourly}. More recently, deep learning approaches have shown promising improvements by training models to map satellite observations directly to regional ground radar \citep{cannon2026deep} or, as in Oya \citep{brempong2025oya}, to spaceborne GPM retrievals; however, these models operate on local spatial patches and rely solely on satellite observations, without incorporating the three-dimensional atmospheric state.

\noindent\textbf{Reanalysis products} such as ERA5 \citep{hersbach2020era5} and JRA-55 \citep{kobayashi2015jra55} provide globally
complete fields by assimilating observations into physical atmospheric models, but their
precipitation remains a diagnostic byproduct of convective and microphysical parameterization schemes.
As a result, model-based precipitation suffers from persistent, well-documented
biases, including systematic overestimation of light rain frequency, underestimation of
heavy precipitation intensity, and premature triggering of the diurnal convective
cycle \citep{stephens2010dreary, lavers2022evaluation, wang2025hourly, bechtold2014representing}. 

\noindent\textbf{Merged multi-source products} such as MSWEP \citep{beck2019mswep, wang2026mswep} and GMCP \citep{ma2025gmcp}
push skill further by combining satellite, reanalysis, and gauge data.
However, existing merged products post-process intermediate precipitation estimates rather than
learning directly from the underlying multi-spectral imagery and three-dimensional atmospheric state,
thereby inheriting limitations of their inputs. Furthermore, while gauge correction improves accuracy where station networks are dense, it introduces reporting latency, offers limited benefit over oceans and sparsely gauged regions, and complicates independent verification where radar composites cover the same densely gauged regions.

Meanwhile, advances in graph neural networks and transformer architectures have
revolutionized weather forecasting \citep{bi2023pangu, lam2023learning, lang2024aifs,
price2025probabilistic, alet2025skillful}.
Building on this foundation, WeatherNext~3 \citep{wn3} showed that a spherical graph-mesh transformer can take
both atmospheric state analyses and direct, non-analysis observational streams as inputs, and also
learn directly from observational targets. However, these advances have been directed toward
forward weather forecasting, and have not yet been applied to global precipitation estimation.

In this paper, we introduce \pardig (Precipitation AI Reanalysis: Densely Inferred GPM),
a novel approach to global precipitation estimation. \pardig trains a graph-mesh transformer neural network
to predict sparse measurements from the active spaceborne radar onboard the Global Precipitation Measurement (GPM) core observatory, given multi-modal
input observations. Our approach leverages two complementary properties: first, GPM provides
sparse but high-quality precipitation measurements along its orbital swaths that can serve directly as supervision;
second, multi-spectral geostationary satellite imagery and 3D atmospheric state analyses provide globally dense,
complementary signatures of precipitation. By training a neural network to map these dense
multi-modal inputs to the sparse radar targets, \pardig learns this relationship and applies
it densely across the globe at every timestep to produce a complete precipitation field.
Crucially, by relying exclusively on spaceborne GPM retrievals for supervision and calibration,
\pardig never uses ground-based radar or rain gauge observations at any stage, decoupling global
skill from local gauge density and keeping all ground-based networks strictly independent for evaluation.

We produce two complementary datasets: our primary product, \pardignrt (Near-Real-Time), combines ECMWF HRES atmospheric state and surface fields with a 30-minute, 11-channel geostationary satellite mosaic to produce global precipitation estimates at \tenthdegree spatial and 15-minute temporal resolution (2016--present), delivered operationally with an end-to-end latency of 2--8 hours from observation time; and a historical variant, \pardiglr (Long Record), which uses ERA5, GridSat-B1 \citep{knapp2011globally}, and passive microwave sounders to extend coverage back to 1999, demonstrating that learning from sparse radar targets is a general paradigm that readily transfers across input streams.

Through comprehensive evaluation against three independent radar networks spanning three continents (MRMS over the US, EURADCLIM over Europe, and JMA Radar-AMeDAS over Japan), over 36,000 surface stations worldwide, and multiple extreme precipitation events, we demonstrate that \pardig consistently outperforms IMERG~V07 (Final and Early), MSWEP~V3, ERA5, and HRES across precipitation thresholds and regions. The largest improvements occur at extreme intensities where existing products struggle most. Furthermore, the benefits extend downstream: using \pardig as a training target in WeatherNext~3 leads to measurably improved precipitation forecasts compared to training on IMERG Final or analysis precipitation.

\section{Model and Training Data}
\label{sec:data}
\label{sec:methods}

\subsection{Input Data}
\label{sec:input_data}

\pardig ingests multi-modal observations through two primary data streams, atmospheric analysis and geostationary satellite imagery (\Cref{fig:figure_1}). For each reference time $t_0$, the temporal input window spans the 4-hour half-open
interval $[t_0 - 2\,\text{h},\, t_0 + 2\,\text{h})$, comprising four hourly atmospheric
state snapshots ($t_0 - 2\,\text{h}$, $t_0 - 1\,\text{h}$, $t_0$, and $t_0 + 1\,\text{h}$)
and eight 30-minute geostationary frames ($t_0 - 120\,\text{min}$ to $t_0 + 90\,\text{min}$).

\paragraph{Atmospheric state fields.} ECMWF's operational High-Resolution Forecast
(HRES) short-range forecasts (lead times 0--5\,h from each 6-hourly cycle, forming a
continuous hourly stream from 2016) provide upper-air variables (temperature,
geopotential, winds, humidity, vertical velocity) at 13 pressure levels, plus surface
fields (\Cref{tab:input_channels}). We use the pressure level variables at
\quarterdegree resolution and the surface variables at both \tenthdegree and
\quarterdegree.

\paragraph{Geostationary satellite imagery (SatWeave).} \pardignrt ingests
\textbf{SatWeave},\footnote{\url{https://developers.google.com/weathernext/guides/satweave}} a global
composite mosaic of operational geostationary satellites (GOES, Meteosat, and Himawari) released concurrently with
the \pardig datasets and previously used in \citet{agrawal2025nowcasting} and
WeatherNext~3 \citep{wn3}. Out of the 17 spectral channels in the publicly released SatWeave dataset,
\pardignrt ingests 11 channels (spanning visible, near-infrared, shortwave infrared, water vapor, and thermal infrared bands;
\Cref{tab:input_channels}) at a 30-minute cadence from 2016 onward,
regridded to both \tenthdegree and \quarterdegree resolution.

The input streams detailed above correspond to our primary near-real-time product (\pardignrt). For the historical variant (\pardiglr, extending back to 1999 prior to SatWeave availability), HRES is replaced with ERA5, SatWeave with standalone NOAA GridSat-B1, and polar-orbiting passive microwave sounder radiances are incorporated, with all inputs introduced at \quarterdegree resolution. More details can be found in \Cref{sec:supp_variants,sec:supp_inputs}.

\subsection{Training Target: Sparse GPM Data}
\label{sec:target}

\pardig is trained to predict observations from the GPM (Global Precipitation Measurement)
Core Observatory satellite \citep{hou2014global} over 2016--2023 (for \pardignrt).
The target variable is the near-surface precipitation rate from GPM's Combined Radar-Radiometer Algorithm
(CORRA), which merges the Dual-frequency Precipitation Radar (DPR) and the GPM Microwave
Imager (GMI) to produce calibrated precipitation profiles \citep{grecu2016gpm}.
The temporal output (or target) window spans $\pm$1 hours centered on the target time $t_0$,
totalling 2 consecutive hours of sparse GPM data (${\sim}$500,000 footprints per window, or
${\sim}$250,000 per hour).

GPM observations are sparse in both space and time. The raw CORRA product is not a map but
a set of individual radar footprints, each resolving a ground patch of ${\sim}5$\,km across
and carrying a latitude, a longitude, an observation time, and a single near-surface
precipitation rate; successive scans sweep out a swath ${\sim}245$\,km wide along the
satellite's \ang{65}-inclination orbit, providing coverage out to ${\sim}$\ang{67}N/S \citep{iguchi2024gpm}. These footprints supervise the model at their native
positions and times, and are never gridded or interpolated onto a regular map. Because the
Core Observatory revisits a given location only once every few days, any
individual location is unobserved the overwhelming majority of the time.

\subsection{Architecture}
\label{sec:architecture}

\begin{figure}[t!]
    \centering
    \includegraphics[width=\linewidth]{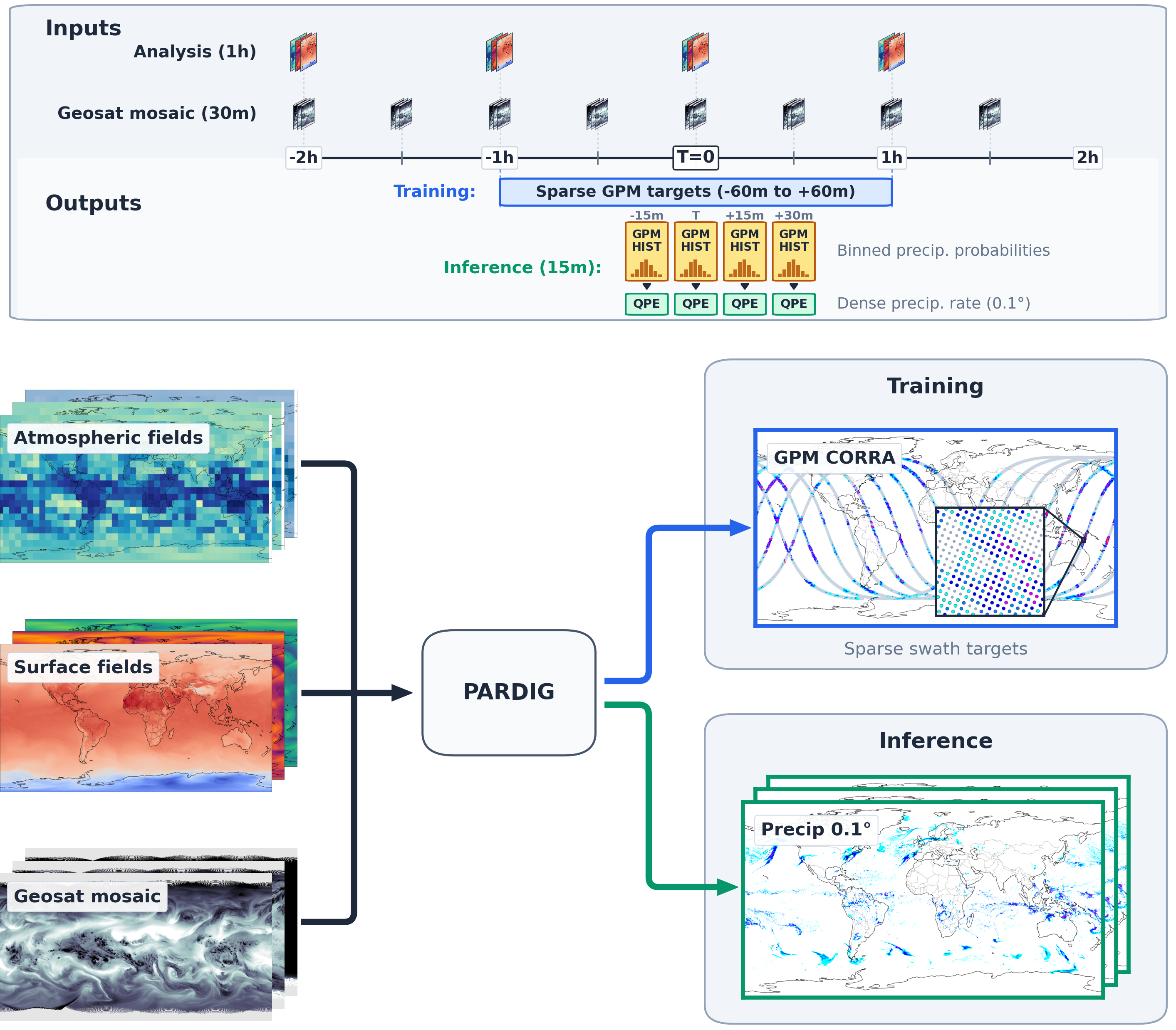}
    \caption{\textbf{Model overview.} Schematic showing the different input and output modalities,
    temporal offsets, and training and inference pathways of \pardig. For each reference
    time $t_0$ (marked $T=0$ on the timeline), \pardig takes two sources of input spanning a 4-hour half-open window ($[-2\,\text{h},\, +2\,\text{h})$):
    four hourly atmospheric analysis frames ($-2\,\text{h}$ to $+1\,\text{h}$), and
    eight geostationary satellite mosaic frames at 30-minute cadence ($-120\,\text{min}$ to $+90\,\text{min}$).
    Upper-air atmospheric fields are ingested at \quarterdegree resolution, while surface fields and the
    geostationary mosaic are provided at both \tenthdegree and \quarterdegree resolution. \pardig's output
    head predicts binned precipitation probabilities at arbitrary spatial and temporal resolution
    (i.e., is continuously queryable in space and time). During training (blue), the model is queried
    at off-grid, sparse GPM CORRA radar swath data within a $\pm 60$-minute window around $t_0$.
    At inference (green), the model is queried on a global \tenthdegree grid at four 15-minute
    timesteps ($-15$, $0$, $+15$, and $+30$\,minutes), and the predicted probability histograms are
    converted via Quantitative Precipitation Estimation (QPE) into deterministic precipitation rate fields.}
    \label{fig:figure_1}
\end{figure}

\pardig adopts the core graph-mesh transformer and continuous coordinate-query architecture of WeatherNext~3 \citep{wn3}, adapting a single step of the model from autoregressive ensemble forecasting to single-pass probabilistic precipitation retrieval (\Cref{fig:figure_1}). Whereas WeatherNext~3 generates ensemble trajectories by rolling the atmospheric state forward in time with injected noise to forecast full atmospheric fields alongside multiple gridded and sparse observational targets, \pardig runs a single noise-free forward pass over $[t_0 - 2\,\text{h},\, t_0 + 2\,\text{h})$ and retains only the continuous coordinate-query output head (used for surface weather stations in WeatherNext~3), supervised here on sparse GPM CORRA footprints to predict discretized marginal precipitation distributions directly. As in WeatherNext~3, the dense multi-modal input grids are encoded onto a global icosahedral mesh ($40{,}962$ nodes, 6 refinement levels), processed through sparse transformer attention, using a smaller capacity of 16 layers and latent dimension 512 (versus 32 layers and dimension 1024 in WeatherNext~3) with no dropout, and decoded back onto a dense latitude-longitude latent grid where features are refined by a shallow convolutional network.
Processing the full sphere on the mesh provides a large spatial receptive field that captures
synoptic-scale context while producing seamless global fields without the patch-boundary artifacts of
regional or tiled models. For \pardignrt, a dedicated \tenthdegree grid encoder-decoder stream processes
the higher-resolution surface and geostationary inputs alongside the \quarterdegree stream to preserve
fine-scale convective structure on the latent grid.

The key mechanism enabling \pardig to learn from sparse orbital swaths and generalize to dense global
maps is a coordinate-query output head that operates on top of this dense latent grid. Given any
spatio-temporal query coordinate $(\text{lat}, \text{lon}, \Delta t)$, where $\Delta t = t - t_0$ is the
time offset relative to the window center, the output head bilinearly interpolates the latent grid to
$(\text{lat}, \text{lon})$, combines the interpolated features with an embedding of $\Delta t$, and
passes the result through a shared multilayer perceptron (MLP) to predict a binned precipitation
probability distribution. Because the encoder and processor always construct a complete global latent
grid from the dense inputs, the output head can be queried at arbitrary locations and times within the
target window. During training, we query the head only at the known positions and timestamps of the
sparse GPM CORRA footprints. At inference, we run one pass per hour $t_0$ and query the same head at
every point on a global \tenthdegree grid across four sub-hourly offsets
($\Delta t \in \{-15, 0, +15, +30\}$ minutes), producing a complete global precipitation field every
15 minutes. Because the output head is queried at continuous coordinates, this \tenthdegree, 15-minute
output grid is a design choice rather than an architectural limitation, and the model can be queried
at finer spatial or temporal resolution if desired. Aside from these adaptations, the underlying encoder, mesh processor, and decoder blocks follow the design of WeatherNext~3; full architectural specifications and training details are provided in
\Cref{sec:supp_training}.

\subsection{Probabilistic Output and Quantitative Precipitation Estimation}
\label{sec:qpe}

Rather than predicting a single deterministic precipitation rate, \pardig outputs
a probability distribution over precipitation intensity through a binned classification
approach, following the methodology introduced by MetNet \citep{sonderby2020metnet}.
For each query coordinate, the model predicts logits over 14 precipitation rate bins
defined by 15 boundaries (in mm/h). Bin probabilities are computed via softmax, and
the training loss is cross-entropy against the discretized GPM observation. This classification formulation avoids
a well-known limitation of regression losses (MSE, MAE): for skewed distributions
like precipitation, regression produces conditional-mean predictions that systematically
overestimate light rain frequency and underestimate heavy precipitation intensity
\citep{ravuri2021skilful, andrychowicz2023deep}.

While the resulting per-bin probabilities provide a marginal probability distribution
at each query point that can directly support probabilistic applications in future work,
in this paper we focus on deriving deterministic precipitation rates from the predicted
distribution. To do so, we first compute the exceedance probability $E_k = P(\text{precip} \ge b_k)$ for each
bin boundary $b_k$ by summing the probabilities of all bins at or above $b_k$. Using a fixed
threshold of $E_k \ge 0.5$ across all bins would correspond to extracting the median of the predicted
distribution; however, because precipitation distributions are skewed toward dry and light-rain bins,
the median systematically underestimates rain occurrence and heavy intensities. Similar to
\citet{andrychowicz2023deep}, we instead apply rate-dependent probability thresholds $\{\tau_k\}$
(\Cref{tab:qpe_thresholds}), chosen so that the resulting precipitation rate frequencies approximately
match the empirical GPM CORRA distribution. Given the highest bin boundary $b_{k^*}$ for which
$E_{k^*} \ge \tau_{k^*}$, we obtain the final deterministic estimate by linearly interpolating between
$b_{k^*}$ and $b_{k^*+1}$ according to how far $E_{k^*}$ exceeds $\tau_{k^*}$ relative to the remaining
gap $\tau_{k^*+1} - E_{k^*+1}$ (or assigning $0$\,mm/h if no threshold is exceeded), producing smooth
rate estimates rather than discrete bin assignments. Full mathematical details and the chosen
probability thresholds are provided in \Cref{sec:supp_qpe}.

\section{Evaluation Setup}
\label{sec:eval_setup}

\subsection{Evaluation Ground Truths}
\label{sec:ground_truth}

We evaluate \pardig against four independent ground truth sources spanning
different continents and observation types, none of which are used to train or
calibrate \pardig. We focus the main text on hourly accumulations, the finest temporal resolution common to all ground-truth datasets and baseline products, while evaluations of native 15-minute instantaneous rates and longer accumulations are presented in \Cref{sec:supp_subhourly,sec:supp_accumulation}. For all evaluations, we bilinearly interpolate the precipitation estimates in space (with no temporal interpolation) to the ground-truth data.

\paragraph{Radar networks.} Three regional radar composites serve as gridded ground truth in
\Cref{fig:figure_2,fig:figure_4}. Over the contiguous United States we use the MRMS
(Multi-Radar Multi-Sensor) Quantitative Precipitation Estimate Pass-2 product, which merges ${\sim}180$ radars with gauge data \citep{zhang2016mrms}. Over Europe
we use EURADCLIM (European Radar Climatology), a gauge-adjusted, retrospectively
reprocessed dataset built by KNMI from OPERA composites
\citep{overeem2023euradclim}. At the time of evaluation, EURADCLIM was only available until the end of 2023. Over Japan we use JMA
Radar-AMeDAS, which merges C-band radar with the AMeDAS gauge network
\citep{makihara1996method}. Evaluations against all three radar composites span 2016--2023 in the main text, with MRMS and JMA Radar-AMeDAS extended through 2025 in the year-by-year evaluation (\Cref{sec:supp_yearly_radar}).

\paragraph{Surface weather stations (rain gauges).} We use $36{,}678$ quality-controlled surface stations reporting hourly rain-gauge measurements from Synoptic's\footnote{\url{https://synopticdata.com/}} derived precipitation service over 2018--2023. To ensure reliable per-station skill estimates and exclude short-lived stations or event-only gauges (which transmit only when precipitation occurs and omit dry hours), we require stations to have at least 750 valid hourly measurements (${>}\,1$\,month) with at least 50 positive ($\ge 1$\,mm/h) and at least 50 negative ($< 1$\,mm/h) observations and at most 50\% positive hours, alongside basic coordinate and duplicate-feed quality control detailed in \Cref{sec:gt_data}.

The two observation types have complementary strengths and limitations. Radar composites
provide spatially continuous pixel averages, but they infer precipitation indirectly from reflectivity and are only available for small parts of the globe. Station gauges measure accumulation directly and cover many regions, albeit unevenly, but provide only point measurements, introducing a spatial representativeness
mismatch against kilometer-scale satellite and model estimates.

In addition to these independent radar and station benchmarks, in the appendix we evaluate \pardig against GPM CORRA observations stratified by year (\Cref{sec:supp_temporal}) and by region (\Cref{sec:supp_gpm_eval}), to assess temporal generalization across training and held-out periods and spatial consistency beyond the radar- and station-covered areas.

\subsection{Baseline Products}
\label{sec:baselines}

We benchmark \pardig against five reanalysis and operational products. Comprehensive multi-product
evaluations generally rank NASA's IMERG among the best satellite-based products,
compared with alternatives such as GSMaP, CMORPH, PERSIANN, and PDIR-Now
\citep{beck2019daily, xu2022era5, wang2025hourly}.
IMERG Final V07,
MSWEP V3, and ERA5 represent leading retrospective satellite, multi-source merged, and reanalysis benchmarks, respectively,
while IMERG Early V07 and HRES serve as operational near-real-time counterparts.

\paragraph{IMERG.} NASA's IMERG provides global precipitation estimates based on data from the GPM satellite constellation, in particular combining low-Earth orbit passive microwave radiometer data and geostationary infrared imagery. IMERG is available at 30 minute temporal and \tenthdegree spatial resolution going back to 1998. IMERG is available in an Early, Late and Final version. Here we evaluate the low-latency Early version, available with a ${\sim}$4-hour latency, and the highest-quality Final version, which is released with a delay of ${\sim}$3.5 months and applies a monthly bias adjustment over land using the gridded GPCC rain-gauge analysis \citep{schneider2014gpcc}. We omit the intermediate Late version for simplicity, as its performance is bounded by the Early and Final versions. At the time of writing the most current IMERG version was V07. An update to V08 is currently in development\footnote{\url{https://gpm.nasa.gov/data/news/update-imerg-v08-transition-schedule-aug-2026}}.

\paragraph{ERA5.} ECMWF's ERA5 \citep{hersbach2020era5} is the most commonly used global atmospheric reanalysis, providing hourly three-dimensional state estimates of the atmosphere going back to 1940 at \quarterdegree. ERA5 is produced by assimilating a wide range of observations into a numerical weather prediction model. Crucially, precipitation is largely not directly assimilated but rather is indirectly predicted by the model physics. Final quality-checked ERA5 fields are released after 2--3 months, while a preliminary low-latency stream (ERA5T) is available with a delay of roughly 5 days.

\paragraph{HRES.} As a low-latency analysis-based precipitation estimate we evaluate short-range forecasts from ECMWF's operational high-resolution forecast model at \tenthdegree. Specifically we evaluate the first 6 hours of every 6-hourly forecast cycle. When compared to ERA5,
HRES has the advantage of higher spatial resolution, and benefits from advancements in model
physics achieved in the past decade, but the disadvantage of an earlier observation cutoff.

\paragraph{MSWEP.} GloH2O's Multi-Source Weighted-Ensemble Precipitation V3 \citep{wang2026mswep} provides global hourly precipitation estimates at \tenthdegree resolution from 1979 onward by blending satellite retrievals, ERA5 reanalysis, and rain-gauge observations using machine learning. MSWEP V3 offers both a retrospective gauge-corrected archive (MSWEP-GC) and a near-real-time stream (MSWEP-NRT). Here we evaluate MSWEP-GC, as it represents MSWEP's highest-accuracy version and archived MSWEP-NRT retains only finalized (${\sim}5$-day) fields rather than initial real-time estimates. Unlike satellite-only or NWP products, MSWEP-GC directly incorporates daily rain-gauge observations and trains its blending model on regional gauge and radar references (including Stage-IV and EURADCLIM).

\subsection{Evaluation Protocol and Metrics}
\label{sec:eval_protocol}

Evaluating precipitation datasets over multi-year records against independent observation
networks is standard practice to adequately sample interannual variability and extreme
events \citep{beck2019daily, xu2022era5}. Our multi-year radar evaluations span 2016--2023
(and the global station evaluations span 2018--2023), which represent the common overlap
periods over which all evaluated products (including \pardiglr, which ends in 2023) and the
respective ground-truth records (in particular, EURADCLIM, which ends in December 2023) are
simultaneously available.

Although these multi-year windows overlap with the \pardignrt training period (2016--2023),
the ground-based benchmarks remain out-of-sample in two respects. First, all ground-based
evaluations are strictly cross-modal: \pardig is supervised exclusively by spaceborne GPM
CORRA retrievals and never uses ground radar or rain gauge observations at any stage. Second,
while evaluating directly against GPM footprints during 2016--2023 is in-sample, ground-based
networks are not: GPM's narrow orbital swath (${\sim}$245\,km width with revisit
times of several days) samples only a few percent of hourly grid-cell observations, so the vast majority of the space-time
points evaluated in the radar and station records had no GPM overpass. In addition, to verify
temporal generalization directly, we evaluate \pardignrt across the strictly held-out post-training years 2024--2025
(\Cref{sec:supp_temporal}), both against GPM CORRA swaths
(\Cref{fig:supp_gpm_yearly}) and against the radar networks
that extend beyond 2023 (MRMS over CONUS and JMA Radar-AMeDAS over Japan;
\Cref{fig:supp_yearly_csi}). Whereas GPM evaluations show the
expected modest difference between in-training (2016--2023) and unseen (2024--2025) satellite
swaths, the independent radar evaluations exhibit no degradation between 2016--2023 and
2024--2025 across any region or threshold, confirming consistent generalization to future
years and an operational near-real-time setting.

We evaluate all products using the Critical Success Index (CSI = hits / (hits + misses + false alarms)),
which penalizes both missed precipitation and false alarms without rewarding correct negatives, an essential property
given that most grid cells are dry at any given time. A suite of
additional categorical metrics ($\text{F}_1$ score, precision, recall, SEDI, frequency bias, and FSS)
is provided in \Cref{sec:supp_metrics}.

\section{Results}
\label{sec:results}

\subsection{Multi-Regional Radar Evaluation}
\label{sec:csi_results}

\begin{figure}[htbp]
    \centering
    \includegraphics[width=\linewidth]{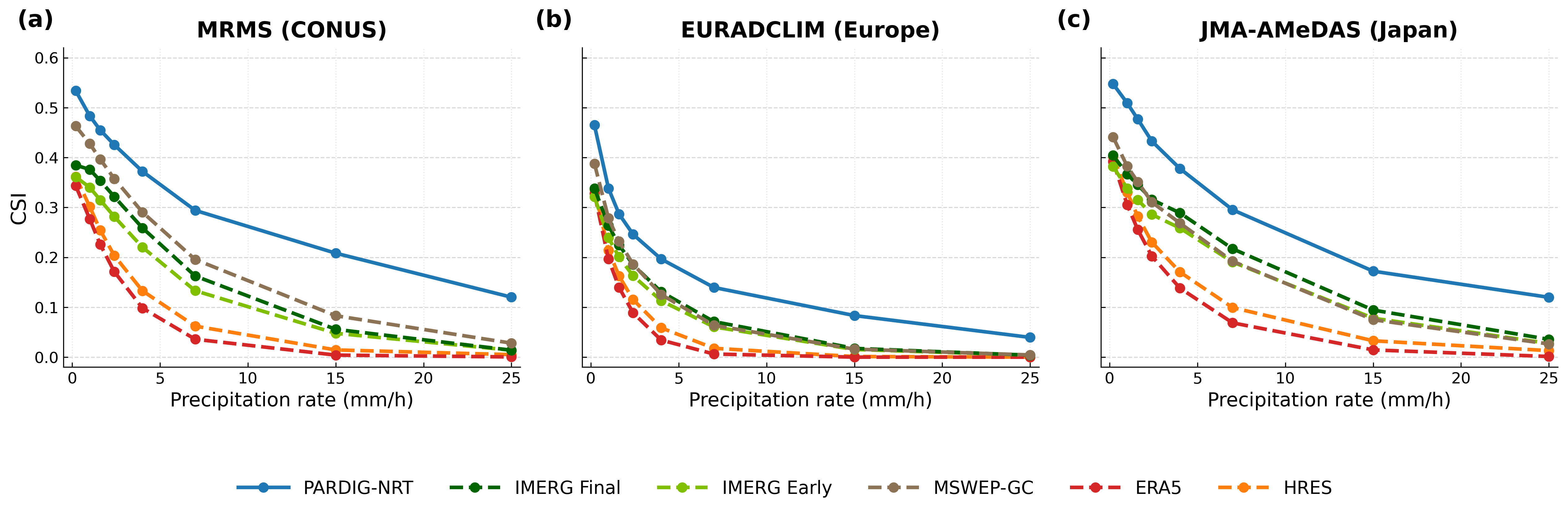}
    \caption{\textbf{Precipitation detection skill across rain rate thresholds.} CSI as a function
    of hourly precipitation rate threshold evaluated against three independent, gauge-adjusted ground
    radar networks (multi-year means): (\textbf{a}) MRMS over the contiguous United States
    (CONUS, 2016--2023), (\textbf{b}) EURADCLIM over Europe
    (2016--2023), and (\textbf{c}) JMA Radar-AMeDAS over Japan
    (2016--2023). \pardignrt (solid blue) consistently achieves
    the highest CSI across all rain rate thresholds in all three regions, with
    the largest relative improvements over the baselines occurring at higher rain rates.
    Numerical CSI values are provided
    in \Cref{tab:csi_all}; see \Cref{fig:supp_yearly_csi}
    for year-by-year stability and \Cref{sec:supp_metrics}
    for additional categorical metrics.}
    \label{fig:figure_2}
\end{figure}

\pardig achieves the highest CSI at every threshold
in all three evaluation regions, with consistent improvements across all precipitation rates (\Cref{fig:figure_2}).
At the 1\,mm/h threshold,
\pardignrt achieves a multi-year mean CSI of $0.48$ over CONUS, compared with $0.43$ for MSWEP-GC,
$0.38$ for IMERG Final, and $0.28$ for ERA5. At 4\,mm/h, the gap widens: \pardignrt scores $0.37$,
while MSWEP-GC falls to $0.29$, IMERG Final to $0.26$, and ERA5 to $0.10$. At 25\,mm/h, baseline products retain
very limited skill ($\text{CSI} \le 0.03$), while \pardignrt achieves $0.12$, a considerable step change.
The pattern holds across the other two radar networks over Europe and Japan.

Year-by-year evaluations (\Cref{fig:supp_yearly_csi}) show strong interannual stability across all products and confirm that \pardignrt's skill carries over without degradation to the held-out post-training years (2024--2025). Over CONUS, held-out CSI reaches $0.48\text{--}0.49$ at 1\,mm/h and $0.24\text{--}0.25$ at 15\,mm/h (versus 2016--2023 means of $0.48$ and $0.21$), with similarly steady generalization over Japan across all thresholds (\Cref{sec:supp_temporal}).

The frequency bias (ratio of predicted to observed precipitation frequency,
\Cref{fig:supp_all_metrics}(d1--d3)) confirms that \pardig is reasonably calibrated across thresholds.
MSWEP-GC, IMERG Final, and IMERG Early are less biased at light-to-moderate rain rates but increasingly underforecast heavy events ($\ge 15$\,mm/h),
whereas ERA5 and HRES systematically overestimate light rain frequency and drop sharply toward zero at higher rates, a well-known limitation of model-based precipitation;
the global precipitation distributions (\Cref{sec:supp_consistency}) further illustrate this contrast, with ERA5 and HRES
predicting precipitation over a substantially larger fraction of global area
than satellite and merged products. \pardig achieves both higher precision and higher recall at nearly all thresholds, highlighting that the CSI improvements come from more accurately estimating the timing and location of precipitation rather than simply predicting more rain or being better calibrated (\Cref{fig:supp_all_metrics}(b1--b3) and \Cref{tab:pr_all}).

Additional radar evaluations at 6-hour and 24-hour accumulation timescales are provided in \Cref{sec:supp_accumulation}, confirming that strong skill extends beyond the hourly resolution to sub-daily and daily precipitation, timescales that are also commonly used in hydrological and weather applications.

\subsection{Global Station Evaluation}
\label{sec:station_results}
\suppressfloats[t]

\begin{figure}[!htb]
    \centering
    \includegraphics[width=\linewidth]{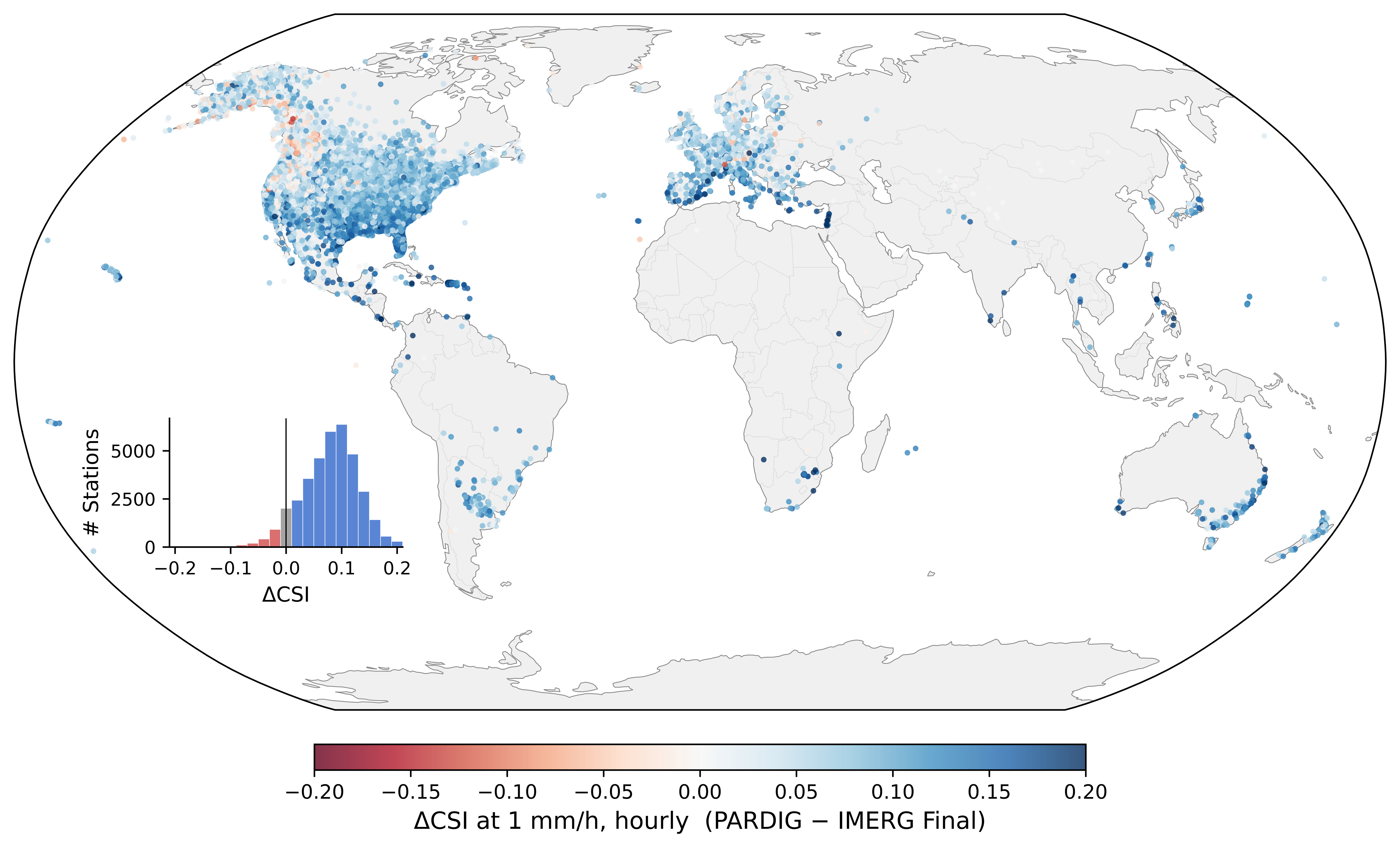}
    \caption{\textbf{Global station-level skill improvement.} Per-station CSI difference ($\Delta\text{CSI}$: \pardignrt minus IMERG Final) across $36{,}678$ independent quality-controlled Synoptic surface rain-gauge stations worldwide (2018--2023), evaluated at the 1\,mm/h hourly accumulation threshold. Blue circles indicate stations where \pardignrt achieves higher CSI; red circles indicate stations where IMERG Final achieves higher CSI (color scale clipped to $[-0.20, +0.20]$). The inset histogram shows the distribution of per-station $\Delta\text{CSI}$ values across all $36{,}678$ stations (with the neutral center bin $|\Delta\text{CSI}| < 0.01$ shaded grey).}\label{fig:figure_3}
\end{figure}

\Cref{fig:figure_3} presents a global map of per-station $\Delta$CSI (\pardignrt minus IMERG Final)
across $36{,}678$ quality-controlled surface gauge stations worldwide (\Cref{sec:ground_truth}),
evaluated at the 1\,mm/h hourly threshold.

Defining stations within the neutral center bin ($|\Delta\text{CSI}| < 0.01$) as tied, \pardignrt outperforms IMERG Final at $89.9\%$ of stations worldwide, whereas IMERG Final is better at $4.6\%$ and $5.5\%$ are tied. Across all $36{,}678$ stations, \pardignrt achieves a mean CSI of $0.36$ compared with $0.28$ for IMERG Final, a mean improvement of $+0.08$ CSI points ($+28.5\%$ relative). The inset histogram is heavily right-skewed, peaking at $\Delta\text{CSI} = 0.10$, with over one-third of all stations ($35.9\%$) exhibiting a gain of $\Delta\text{CSI} \ge +0.10$.

The improvement spans all continents with observing stations. Although gauge coverage in the tropical and subtropical belt ($\ang{30}$S--$\ang{30}$N) is sparser than in the Northern Hemisphere mid-latitudes, across the $2{,}841$ stations in this band \pardignrt outperforms IMERG Final at $96.8\%$ of stations (vs.\ $0.8\%$ for IMERG Final and $2.5\%$ tied), with a mean gain of $+0.12$ CSI points ($0.36$ vs.\ $0.24$, $+52\%$ relative), compared with $89.3\%$ of stations and a mean gain of $+0.08$ CSI points ($0.36$ vs.\ $0.28$, $+27\%$ relative) poleward of $\ang{30}$. Conversely, the small fraction of stations where IMERG Final scores higher ($4.6\%$ globally) occurs primarily in high-elevation mountainous terrain at mid-latitudes, such as parts of the North American Rockies and the European Alps.

A similar per-station comparison between \pardignrt and MSWEP-GC, along with station-averaged CSI across all hourly thresholds ($0.2$--$25$\,mm/h) for all seven products (including \pardiglr), is provided in \Cref{sec:supp_station_csi}. Finally, to complement these station evaluations in regions where ground-based networks are sparse or absent, we also evaluate \pardignrt against held-out GPM CORRA swaths across 16 regional domains (\Cref{sec:supp_gpm_eval}), confirming strong skill over tropical and southern-hemisphere landmasses.

\subsection{Extreme Precipitation Case Studies}
\label{sec:case_studies}

\begin{figure}[p]
    \centering
    \makebox[\linewidth][c]{\includegraphics[width=1.08\linewidth, height=0.80\textheight, keepaspectratio]{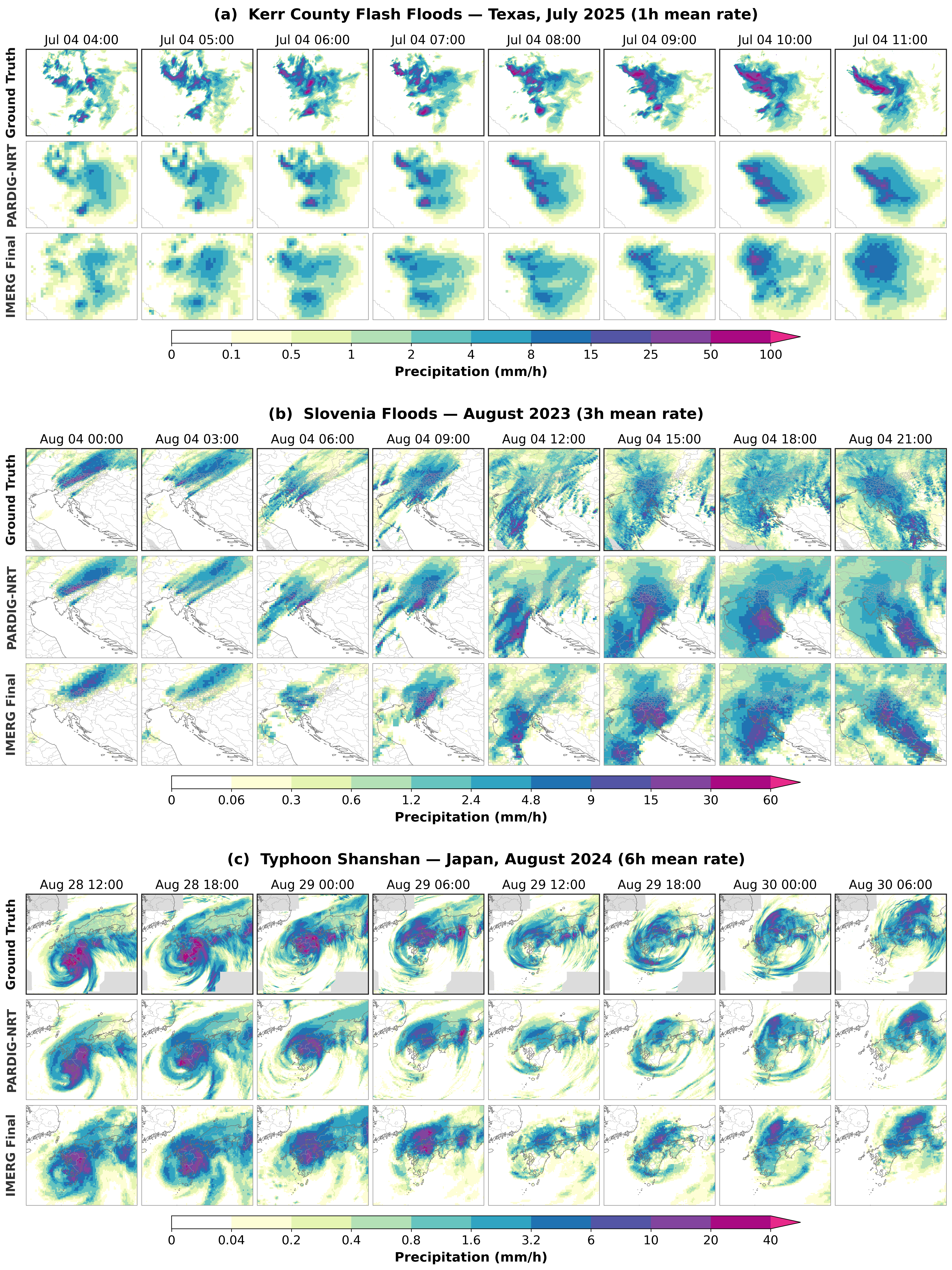}}
    \caption{\textbf{High-resolution case studies of extreme precipitation events.}
    Sequential 8-step progressions for (\textbf{a}) Kerr County convective flash floods,
    Texas (July 2025, 1h mean rate, MRMS), (\textbf{b}) Orographic floods, Slovenia
    (August 2023, 3h mean rate, EURADCLIM), and (\textbf{c}) Typhoon Shanshan landfall,
    Japan (August 2024, 6h mean rate, JMA Radar-AMeDAS). Rows display gauge-adjusted radar
    ground truth (top, bold frame), \pardignrt (middle), and IMERG Final V07 (bottom) in mm/h,
    each shown at its native spatial resolution ($\ang{0.01}$ for MRMS and JMA Radar-AMeDAS,
    $\ang{0.02}$ for EURADCLIM, and $\ang{0.1}$ for \pardignrt and IMERG Final).
    Column headers denote accumulation interval start times in UTC ($[t, t + \Delta t)$);
    light grey shading in the ground-truth panels indicates regions outside radar coverage.
    Numerical CSI verification across thresholds is detailed in \Cref{tab:case_study_csi}.}
    \label{fig:figure_4}
\end{figure}

To illustrate how these multi-year statistical results manifest at the storm scale,
\Cref{fig:figure_4} examines one recent high-impact precipitation event from each of our
three radar benchmark regions, chosen to span three distinct dynamical regimes and
accumulation windows: convective flash floods in central Texas (Kerr County, July 2025, 1h),
frontal and orographic floods in Slovenia (August 2023, 3h), and tropical cyclone landfall in
Japan (Typhoon Shanshan, August 2024, 6h). The Texas and Japan events occurred after the
2016--2023 training period, and the Slovenia event occurred in the final year of EURADCLIM
availability. For each event, we show 8 sequential time steps comparing gauge-adjusted radar
ground truth, \pardignrt, and IMERG Final.

\paragraph{Kerr County flash floods, Texas (\Cref{fig:figure_4}(a), July 2025, 1h accumulation).} An
intense mesoscale convective system produced localized extreme precipitation over central Texas.
At the 1-hour accumulation window,
\pardig resolves individual convective cells rather than smoothing them into a broad,
diffuse precipitation field, capturing the rapid onset and movement of the system with high
fidelity against MRMS radar. The advantage over IMERG is most pronounced at extreme
thresholds: at $15$\,mm/h and $25$\,mm/h, \pardig achieves event-level CSIs of $0.29$ and $0.10$, compared
to $0.08$ and $0.02$ for IMERG Final.

\paragraph{Slovenia floods (\Cref{fig:figure_4}(b), August 2023, 3h accumulation).} A slow-moving frontal
system produced sustained heavy precipitation over the Julian Alps and surrounding valleys,
triggering the most extreme flood event in Slovenia in recent decades \citep{bezak2023first}.
\pardig accurately captures the spatial structure, location, and progression of the precipitation bands observed in the EURADCLIM radar composite, though it overestimates precipitation intensity during the final three frames as the system shifts southward. Despite this late-stage intensity bias, \pardig outperforms IMERG Final across all thresholds over the event, achieving a CSI of $0.38$ versus $0.32$ at a $4$\,mm/h mean rate ($12$\,mm/3h) and $0.22$ versus $0.16$ at $7$\,mm/h ($21$\,mm/3h).

\paragraph{Typhoon Shanshan (\Cref{fig:figure_4}(c), August 2024, 6h accumulation).} Typhoon Shanshan
made landfall in southwestern Japan with extreme rainfall. \pardig reproduces the asymmetric precipitation distribution characteristic of
tropical cyclones (heavy rainfall concentrated in the northeast quadrant relative
to the storm center) as well as the feeder bands extending hundreds of kilometers
from the core. Over the 48-hour cyclone passage, \pardig achieves slightly higher skill than IMERG Final across all thresholds, with an event CSI of $0.46$ versus $0.43$ at $24$\,mm/6h ($4$\,mm/h mean rate) and $0.38$ versus $0.36$ at $42$\,mm/6h ($7$\,mm/h mean rate).

Two additional case studies from 2017 and 2023 that also include \pardiglr and MSWEP-GC are
shown in \Cref{fig:supp_case_study}. Although a handful of storms cannot by themselves
establish general performance, the sharper rainband structure and heavy-rain skill seen
across these examples align with the multi-year radar and gauge evaluations in
\Cref{sec:csi_results,sec:station_results,sec:supp_accumulation}.

\subsection{Spatiotemporal Structure and Physical Consistency}
\label{sec:physical_consistency}

Beyond grid-point categorical skill, \pardig preserves fine-scale spatial structure and
sub-hourly temporal evolution without over-smoothing precipitation fields in space or time.
Globally, \pardig retains higher spherical harmonic variance across $50\text{--}200$\,km
wavelengths than MSWEP-GC, IMERG Final, or HRES (\Cref{sec:supp_spectral}), and CONUS spatial
semivariograms against MRMS radar confirm that this variance reflects a more realistic spatial
correlation decay ($240\text{--}250$\,km decorrelation length vs.\ $150\text{--}170$\,km for MRMS,
$290\text{--}300$\,km for HRES and IMERG Final, and $330$\,km for MSWEP-GC; \Cref{sec:supp_variogram}).
At its native 15-minute cadence, \pardignrt maintains uniform skill across all intra-hour
offsets and exhibits rapid frame-to-frame decorrelation approaching MRMS instantaneous radar,
avoiding the slower decorrelation of morphed satellite retrievals (\Cref{sec:supp_subhourly}).

Across regional and latitudinal climatologies, \pardig reproduces observed diurnal and
seasonal regimes (\Cref{sec:diurnal_seasonal}), capturing the nocturnal convective peak over
the US Great Plains and matching JMA radar over Kyushu, while overestimating seasonal cycle
amplitude over the European Alps (underestimating wintertime relative to summertime
precipitation). Across tropical regions without ground radar (South America, Central Africa,
and South Asia), \pardig agrees closely with IMERG Final in seasonal cycle and peaks between ERA5 (which triggers convection too early) and IMERG Final in diurnal phase. Globally, its zonal-mean profile closely tracks
IMERG Final across the Intertropical Convergence Zone, subtropical dry zones, and
mid-latitude storm tracks, while estimating lower mean rates toward the poles
(\Cref{sec:supp_zonal}).

Finally, \pardignrt and the historical \pardiglr variant remain closely aligned across their
2016--2023 overlap, exhibiting near-identical global precipitation distributions with area-weighted
global means within 2\% and aligned upper tails through $p_{99.999}$ (\Cref{sec:supp_overlap_consistency}).
Extending further back in time, \pardiglr maintains stable annual means from 2008 through
2023, whereas sparser early-2000s satellite observations induce a distributional shift prior
to 2008, evidenced by lower annual mean rates (\Cref{sec:supp_long_term_trends}). Across the
independent radar and station benchmarks, \pardiglr closely tracks \pardignrt, outperforming
all external baselines at 1-hour and 6-hour accumulations, while MSWEP-GC is comparable or
slightly better in some of the 24-hour evaluations
(\Cref{sec:supp_metrics,sec:supp_station_csi,sec:supp_accumulation}).

\subsection{Improving Precipitation Forecasts}
\label{sec:forecasting}

\pardig has already been used to improve precipitation forecasts in the recently released WeatherNext 3 (WN3) model \citep{wn3}. WN3 is a state-of-the-art global, probabilistic, medium-range forecast model with two precipitation specific output heads, one trained against IMERG Final data and one against \pardig. The \pardig head consistently achieves better skill when evaluated against independent radar observations. This holds across precipitation intensities and temporal accumulations. As an example, at early lead times (6h), CRPS when evaluated against 6h MRMS accumulations is approximately 15\% lower compared to the IMERG head and 30\% lower compared to the analysis-based precipitation head.\footnote{The WeatherNext~3 training used a slightly earlier internal version of \pardig with a closely similar architecture and input configuration.}

\FloatBarrier
\section{Discussion}
\label{sec:discussion}

We have presented \pardig, a near-real-time global precipitation estimation system
that maps multi-modal atmospheric and satellite inputs to dense global precipitation
fields by learning from sparse, high-quality spaceborne radar observations.

Comprehensive evaluation across three independent radar networks on three continents
demonstrates that \pardig consistently outperforms IMERG~V07 (Final and Early), MSWEP-GC, ERA5, and
HRES across all precipitation thresholds, with the largest improvements at extreme intensities
(\Cref{fig:figure_2}). This advantage is confirmed across over 36,000 independent surface stations
worldwide (\Cref{fig:figure_3}) and through high-resolution case studies of extreme convective, frontal,
and cyclonic events (\Cref{fig:figure_4}). Furthermore, these improvements translate into measurably
more skillful precipitation forecasts when used to train the WeatherNext~3 global
weather model (\Cref{sec:forecasting}).

Notably, \pardig achieves these gains over monthly gauge-adjusted IMERG Final and gauge-corrected MSWEP without relying on
rain gauge inputs or gauge-based bias correction. Because the satellite and reanalysis inputs
are globally continuous and GPM samples precipitation along its orbital swaths without
preferential geographic coverage over land or ocean, model skill is not tied to local ground
infrastructure. While our evaluations are necessarily limited to regions with independent
radar and station observations, this spatial uniformity gives reason to expect more reliable
generalization to sparsely gauged land regions and oceans than for gauge-adjusted products or
models trained on gauge-dependent targets.

The companion \pardiglr dataset extends global coverage back to 1999, demonstrating that
the sparse-to-dense architecture functions as a general framework adaptable to coarser historical
reanalysis, geostationary, and passive microwave streams. Because \pardiglr maintains high skill,
stable annual means from 2008 onward, and closely aligned global distributions with \pardignrt
across their overlap period, the two records can be concatenated into a consistent series from
2008 to the present; prior to 2008, however, sparser early-2000s satellite observations induce a
distributional shift in \pardiglr, warranting caution when using these earlier years
(\Cref{sec:supp_long_term_trends}).

Several limitations of the current system warrant consideration. First, although \pardig outperforms baseline products overall across Europe and North America, it shows smaller margins or occasional station-level deficits relative to IMERG Final at some high-elevation mountainous sites (\Cref{fig:figure_3}) and underestimates wintertime relative to summertime precipitation over the European Alps (\Cref{sec:diurnal_seasonal}). While the causes of these localized differences are not
fully understood (and likely reflect in part the general difficulty of resolving complex
orographic precipitation in atmospheric analyses and satellite retrievals), one plausible
contributing factor to explore in future work is the GPM training target itself, as GPM's
DPR has documented sensitivity limitations for light snowfall and drizzle \citep{hamada2016improvements, chase2022direct}.
Second, because GPM's orbital coverage ends at ${\sim}\ang{67}$N/S, estimates at polar
latitudes rely on unconstrained extrapolation from the dense inputs and should be treated
with caution (\Cref{sec:supp_zonal}).

Looking ahead, this framework can naturally incorporate additional or higher-resolution observational streams and richer training targets to further improve retrieval skill, particularly in challenging regimes such as cold-season precipitation. More broadly, the same sparse-to-dense learning paradigm, using continuous multi-modal context to densely infer sparse, high-precision measurements, offers a general approach for other sparsely observed Earth system variables.

Both the \pardignrt and \pardiglr datasets are openly released,
with the \pardignrt archive continuously updated to support near-real-time operational workflows and research applications.

\section*{Data Availability}

\pardig precipitation estimates at \tenthdegree spatial resolution are openly available under a CC~BY~4.0 license on
Google Cloud Storage as Zarr datasets at \texttt{gs://pardig\_1/NRT/} (\pardignrt historical archive and
continuously updated near-real-time stream) and \texttt{gs://pardig\_1/LR/} (\pardiglr historical archive, 1999--2023),
with dual chunking optimized for global maps and time series analysis. The historical archives are additionally available in the
Google Earth Engine Data Catalog at
\url{https://developers.google.com/earth-engine/datasets/catalog/projects_gcp-public-data-weathernext_assets_PARDIG_1_NRT} (\pardignrt) and
\url{https://developers.google.com/earth-engine/datasets/catalog/projects_gcp-public-data-weathernext_assets_PARDIG_1_LR} (\pardiglr). Full documentation and access instructions are available at \url{https://developers.google.com/weathernext/guides/pardig}.

Operational availability of the \pardignrt stream is governed by the 6-hourly
ECMWF HRES forecast initialization cycles and near-real-time SatWeave composite
latency, producing 15-minute global precipitation fields at \tenthdegree
resolution with end-to-end latencies of approximately 2--8 hours from observation
time to estimate availability, depending on where the observation falls within the 6-hourly HRES cycle; full pipeline details are provided in
\Cref{sec:supp_operational}.

The evaluation ground truth and baseline datasets used in this study are available from their respective providers: MRMS from NOAA NSSL, EURADCLIM from KNMI, JMA Radar-AMeDAS from JMA, IMERG from NASA GES DISC, MSWEP from GloH2O (gloh2o.org/mswep), and surface weather station observations from Synoptic Data PBC (synopticdata.com). ERA5 data are available from the Copernicus Climate Data Store and HRES data from ECMWF.

\section*{Acknowledgments}

In alphabetical order, we thank Adriana Fernandez Lara, Assaf Shmuel, Avinatan Hassidim, Chris Kwaku-Abdullahi, Darshan Prajapati, Devaja Shah, Drew Purves, Elinor Kruse, Frederik Kratzert, Grey Nearing, Jash Rana, Kasia Mohammed, Keita Saito, Kunal Shah, Marc Deisenroth, Martin Gauch, Matthew Willson, Mavish Mahomed-Silva, Oren Gilon, Patrick Ebel, Piyush Ingale, Rachel Stigler, Rahul Mahrsee, Raia Hadsell, Remi Lam, Reuven Sayag, Shail Parekh, Shreya Agrawal, Shubham Joshi, Shubham Kumar, Stephanie Rees, Stephanie Sanchez, Sunny Mak, Tom Andersson, Yuval Yevnin, and Zoubin Ghahramani.

We also thank ECMWF, EUMETSAT, GloH2O, JAXA, JMA, KNMI, NASA, and NOAA for providing invaluable datasets to the research community.

\section*{Disclaimers}

Experimental Estimates: \pardig is an automated, experimental AI system under active research.
Estimates are provided ``as-is'' for informational purposes and are not official severe weather
warnings. For protection of life and property, always defer to official alerts from your local
emergency authorities and national meteorological services. Use of the \pardig datasets is
subject to a Creative Commons Attribution 4.0 International (CC~BY~4.0) license
(\url{https://creativecommons.org/licenses/by/4.0/}).

This document is based on ERA5 and HRES data and products of the European Centre for Medium-Range Weather Forecasts (ECMWF). Copyright \textcopyright{} 2026 European Centre for Medium-Range Weather Forecasts (ECMWF). Source: \url{https://www.ecmwf.int}. This data is published under a Creative Commons Attribution 4.0 International (CC~BY~4.0) license (\url{https://creativecommons.org/licenses/by/4.0/}). The original ECMWF data was modified by regridding and subsetting it for use as inputs in an AI precipitation model. ECMWF does not accept any liability whatsoever for any error or omission in the data, their availability, or for any loss or damage arising from their use.

Based on MSWEP precipitation data, \textcopyright{} GloH2O.

\bibliography{main}

\clearpage
\appendix
\crefalias{section}{appendix}
\crefalias{subsection}{appendix}
\renewcommand{\thesection}{A.\arabic{section}}
\renewcommand{\theHsection}{A.\arabic{section}}
\setcounter{section}{0}
\numberwithin{figure}{section}
\numberwithin{table}{section}
\numberwithin{equation}{section}

\section*{Appendix}

\section{Model Variants and Observational Strategy}
\label{sec:supp_variants}

The primary \pardig product presented in the main text (denoted \textbf{NRT}, Near-Real-Time) is designed for maximal precipitation retrieval quality in the modern satellite era and operates with low latency for near-real-time global monitoring. However, a central strength of the \pardig framework is its modularity: the same underlying architecture can be paired with alternative observational streams to address different operational and scientific needs, such as providing long historical records for training and calibrating downstream models (denoted \textbf{LR}, Long Record).

\subsubsection*{Design Principles and the Dual-Variant Strategy}

The choice of input variables in \pardig is guided by their direct physical relevance to precipitation processes, balanced against three practical considerations:
\begin{enumerate}[label=(\arabic*), leftmargin=*, itemsep=2pt, topsep=2pt]
    \item \textbf{Data quality and detail}: Providing sufficient spatial and spectral resolution to capture cloud structure, atmospheric moisture, and localized storm development.
    \item \textbf{Historical record length}: Extending the precipitation record across multiple decades.
    \item \textbf{Real-time operational availability}: Ensuring inputs are available with minimal latency for near-real-time global monitoring and forecasting.
\end{enumerate}

Because no single observational suite simultaneously satisfies all three requirements, we develop two complementary model configurations (\Cref{tab:variants}):
\begin{itemize}[leftmargin=*, itemsep=2pt, topsep=2pt]
    \item \textbf{\pardignrt (HRES + SatWeave)} is optimized for maximum real-time skill and fine spatial detail. It ingests high-resolution operational ECMWF HRES short-range forecast fields (\tenthdegree and \quarterdegree surface grids, \quarterdegree upper-air) alongside modern 11-channel SatWeave geostationary satellite composites at 30-minute cadence (covering visible, water vapor, and thermal infrared channels), available from 2016 to the present.
    \item \textbf{\pardiglr (ERA5 + GridSat-B1 + Passive Microwave)} extends the inferred global precipitation record back to 1999. It couples ERA5 reanalysis fields with NOAA GridSat-B1 3-channel geostationary imagery and polar-orbiting passive microwave sounder observations. While \pardiglr is trained in the GPM era (2014--2023, when GPM CORRA target observations are available), all of its input streams are available during 1999--2023 (constrained by the coverage of our microwave sounder record), allowing the trained model to be applied retrospectively at inference across the 25-year period. \pardiglr supports historical reanalysis, hydrological, and model-training applications requiring extended records, though multi-decadal shifts in early satellite constellation coverage induce a distributional shift prior to 2008 that should be accounted for when using these earlier years (\Cref{sec:supp_long_term_trends}).
\end{itemize}

\begin{table}[htbp]
\centering
\caption{\textbf{Overview of \pardig model variants.}}
\label{tab:variants}
\resizebox{\textwidth}{!}{%
\begin{tabular}{@{}lllll@{}}
\toprule
Variant & Analysis & Geostationary & Microwave Sounders & Period \\
\midrule
\textbf{\pardignrt} & HRES (\quarterdegree upper-air, \tenthdegree+\quarterdegree surface) & SatWeave (11 ch, \tenthdegree+\quarterdegree, 30-min) & --- & 2016--present \\
\textbf{\pardiglr} & ERA5 (\quarterdegree) & GridSat-B1 (3 ch, \quarterdegree, 3-hourly) & Hourly mosaic (\quarterdegree) & 1999--2023 \\
\bottomrule
\end{tabular}%
}
\end{table}

\subsubsection*{Shared Architecture and Modeling Foundations}

Despite their distinct input configurations, \pardignrt and \pardiglr share an identical core learning pipeline:
\begin{itemize}[leftmargin=*, itemsep=2pt, topsep=2pt]
    \item Both models use the identical Graph-Mesh Transformer architecture and are trained independently on the same GPM CORRA target observations (with \pardiglr trained on scenes from March 2014 onwards).
    \item Both variants utilize the same categorical cross-entropy loss and Quantitative Precipitation Estimation (QPE) formulation, converting predicted probability distributions over 14 intensity bins into deterministic rain rates (with variant-specific calibrated exceedance thresholds detailed in \Cref{sec:supp_qpe}).
    \item For \pardiglr, all inputs are ingested at \quarterdegree resolution without a separate \tenthdegree input stream; dense inference is still performed on a \tenthdegree grid to ensure spatial consistency with the primary \pardignrt dataset.
\end{itemize}

\subsubsection*{Exclusion of Existing Precipitation Products and Ground Networks}

A natural design question is why existing satellite precipitation products are not used as input. Because \pardig is trained exclusively on GPM CORRA data (each co-located with a GMI overpass), IMERG achieves its highest accuracy precisely where \pardig's training targets are available but not at the arbitrary inference query points, when GMI is typically absent. Ingesting raw radiances instead (e.g., ATMS and AMSU-A/MHS in \pardiglr) lets the model learn an end-to-end mapping from primary observations to precipitation, avoiding this co-location bias. In preliminary experiments, including IMERG as an input indeed led the model to over-rely on it during training, thereby improving performance against the GPM target while degrading skill against independent benchmarks.

A distinct design decision is the exclusion of ground-based radars and gauges from both model inputs and training targets. Because ground networks cover only a small fraction of the globe and virtually no ocean regions, supervising a model on ground radar or gauge targets either restricts the domain regionally, inherits cross-network calibration differences, or requires the model to extrapolate spatially to oceans and climate regimes where such targets are never observed \citep{cannon2026deep}. Excluding ground observations from inputs and post-hoc adjustments as well, and relying instead on spaceborne radar alongside globally continuous satellite and reanalysis inputs, provides a uniform formulation across land and sea while preserving all ground-based networks as independent evaluation benchmarks.

\section{Observational Data Streams and Input Channel Specifications}
\label{sec:supp_inputs}

This section details the primary data streams ingested by the \pardignrt and \pardiglr model variants, summarized in \Cref{tab:input_channels}.

\subsubsection*{Input Dataset Descriptions and Processing}

\textbf{Atmospheric state fields.} For \pardignrt, we use upper-air fields across 13 pressure levels at \quarterdegree resolution (1-hour cadence) from ECMWF operational HRES short-range forecasts, alongside dynamic surface variables at both \quarterdegree and \tenthdegree. For \pardiglr, ERA5 reanalysis \citep{hersbach2020era5} provides the same upper-air state and surface fields at \quarterdegree, supplemented by five additional single-level fields (TCWV, CAPE, surface pressure, SST, and total precipitation). While these additional five variables are readily available in the retrospective ERA5 archive (and help compensate for \pardiglr's coarser satellite inputs), they are omitted in \pardignrt so that the model relies only on instantaneous, globally continuous state fields that are consistently available across our existing operational HRES stream. Both variants also ingest four static surface fields from the respective source datasets (\Cref{tab:input_channels}).

\textbf{Geostationary satellite imagery.} \pardignrt ingests 11 spectral channels from \textbf{SatWeave} (\url{https://developers.google.com/weathernext/guides/satweave}) at 30-minute cadence, spanning visible through thermal infrared wavelengths (\Cref{tab:input_channels}) and regridded to both \quarterdegree and \tenthdegree from 2016 to the present. SatWeave merges the calibrated imagery available from operational geostationary satellites (GOES, Meteosat, and Himawari) into a single global mosaic. Note that the SatWeave configuration ingested by \pardignrt differs from the concurrently released public SatWeave dataset (\tenthdegree, 17 channels; 30-minute cadence for 2016--2025 and 1-hour cadence from 2026 onward): \pardignrt uses 30-minute imagery on dual grids (\tenthdegree and \quarterdegree) and an 11-channel subset. Because SatWeave is available from 2016 onward, \pardiglr instead uses standalone NOAA GridSat-B1 \citep{knapp2011globally}, which provides 3-channel geostationary imagery (visible, water vapor, and thermal infrared; native $\ang{0.07}$ resolution, regridded to \quarterdegree) at 3-hourly intervals extending back to the early satellite era.

\textbf{Polar-orbiting passive microwave sounder mosaic.} \pardiglr ingests hourly global composite mosaics of brightness temperatures from polar-orbiting microwave sounders (ATMS on Suomi-NPP and NOAA-20/21 \citep{weng2012atms}, and AMSU-A \citep{mo1996amsu} and MHS \citep{bonsignori2007mhs} on MetOp and NOAA POES), providing direct hydrometeor and temperature/moisture profile information via 17 combined sounding and window channels (\Cref{tab:input_channels}). Microwave mosaics are not included in \pardignrt; their real-time integration will require additional infrastructure work.

\begin{table}[htbp]
\centering
\small
\caption{\textbf{Input channel specifications for \pardig model variants.} Summary of atmospheric state analyses, static surface fields, and satellite observations ingested by \pardignrt and \pardiglr. Upper-air fields span 13 standard pressure levels (50--1000\,hPa).}
\label{tab:input_channels}
\setlength{\tabcolsep}{5pt}
\begin{tabularx}{\textwidth}{@{} >{\raggedright\arraybackslash}p{2.5cm} >{\raggedright\arraybackslash}X >{\raggedright\arraybackslash}p{2.3cm} >{\raggedright\arraybackslash}p{2.6cm} cc @{}}
\toprule
\textbf{Variable Group} & \textbf{Variables / Channels} & \textbf{Source Dataset} & \textbf{Input Resolution} & \textbf{NRT} & \textbf{LR} \\
\midrule
\multicolumn{6}{@{}l}{\textbf{Atmospheric State Analyses}} \\
Upper-Air State (13 levels) & \textit{z}, \textit{t}, \textit{q}, \textit{u}, \textit{v}, \textit{w}\newline \textit{Levels: 50, 100, 150, 200, 250, 300, 400, 500, 600, 700, 850, 925, 1000\,hPa} & HRES / ERA5 & \quarterdegree, 1-hour & \checkmark & \checkmark \\
\addlinespace[3pt]
Surface State \& Forcing & \textit{2t}, \textit{2d}, \textit{10u}, \textit{10v}, \textit{100u}, \textit{100v}, \textit{msl}, \textit{tcc}, \textit{hcc}, \textit{mcc}, \textit{lcc}, \textit{tisr}, diurnal \& seasonal progress ($\sin/\cos$) & HRES / ERA5 & \tenthdegree+\quarterdegree / \quarterdegree, 1-hour & \checkmark & \checkmark \\
\addlinespace[3pt]
Additional ERA5 Surface Fields & \textit{tcwv}, \textit{cape}, \textit{sp}, \textit{sst}, \textit{tp} & ERA5 & \quarterdegree, 1-hour & --- & \checkmark \\
\midrule
\multicolumn{6}{@{}l}{\textbf{Static Surface Fields}} \\
Topography \& Land Cover & Surface Geopotential / Elevation, Land-Sea Mask & HRES / ERA5 & \tenthdegree+\quarterdegree / \quarterdegree, static & \checkmark & \checkmark \\
\addlinespace[3pt]
Sub-grid Orography \& Lakes & Orography Std Dev ($\sigma_{\text{orog}}$), Lake Cover fraction & HRES / ERA5 & \quarterdegree, static & \checkmark & \checkmark \\
\midrule
\multicolumn{6}{@{}l}{\textbf{Satellite Observations}} \\
SatWeave Geostationary Mosaic & 11 channels (0.64, 0.86, 1.61, 3.83, 6.20, 7.33, 8.59, 9.62, 11.20, 12.40, 13.30\,$\mu$m) & SatWeave & \tenthdegree+\quarterdegree, 30m & \checkmark & --- \\
\addlinespace[3pt]
GridSat-B1 Imagery & 3 channels (0.6\,$\mu$m VIS, 6.7\,$\mu$m WV, 11.0\,$\mu$m IR) & GridSat-B1 & \quarterdegree, 3-hourly & --- & \checkmark \\
\addlinespace[3pt]
Microwave Sounder Mosaic & 17 combined sounding and window channels (23.8, 31.4\,GHz; 12 channels at 50.3--57.3\,GHz $\text{O}_2$; 88.2/89.0\,GHz; $183.31\pm 1.0$, $183.31\pm 3.0$\,GHz $\text{H}_2\text{O}$) & ATMS, AMSU-A, MHS & \quarterdegree, 1-hour & --- & \checkmark \\
\bottomrule
\end{tabularx}
\end{table}

\section{Quantitative Precipitation Estimation (QPE) Details}
\label{sec:supp_qpe}

The \pardig model outputs a categorical probability distribution over $K = 14$ precipitation
intensity bins rather than directly regressing scalar rain rates. The Quantitative
Precipitation Estimation (QPE) procedure converts these predicted discrete probability
distributions into deterministic precipitation rate estimates (in mm/h).

\subsection{Bin Discretization and Classification Formulation}

During training, GPM precipitation intensity is discretized into $K = 14$ bins using
15 bin boundaries $\mathbf{b} = [b_0, \dots, b_{14}]$, defined as:
\begin{equation}
\mathbf{b} = [0,\ 0.01,\ 0.1,\ 0.2,\ 1.0,\ 1.6,\ 2.4,\ 4.0,\ 7.0,\ 15.0,\ 25.0,\ 35.0,\ 50.0,\ 75.0,\ 100.0]\quad\text{(in mm/h)}.
\end{equation}
Bin $i \in \{0, \dots, 13\}$ corresponds to the precipitation rate interval $[b_i, b_{i+1})$, where $i = 0$ ($[0, 0.01)$\,mm/h) represents the dry regime and observations exceeding $100.0$\,mm/h are assigned to the uppermost bin ($i = 13$).

For any query location $x$ and timestep $t$, the decoder produces unnormalized logits 
$\mathbf{z}(x, t) = [z_0, \dots, z_{13}] \in \mathbb{R}^{14}$. Categorical class probabilities
$\mathbf{p}(x, t) = [p_0, \dots, p_{13}]$ are computed via the softmax function:
\begin{equation}
p_i(x, t) = \frac{\exp(z_i(x, t))}{\sum_{j=0}^{13} \exp(z_j(x, t))}.
\end{equation}
During training, the objective is standard multi-class cross-entropy loss against the
one-hot encoded ground truth bin index corresponding to the observed GPM CORRA surface rate.

\subsection{Exceedance Probabilities and Calibrated Decision Thresholds}

To convert the predicted bin probabilities into deterministic rates without underestimating heavy precipitation, \pardig uses an exceedance probability formulation. Exceedance probabilities are defined for the $13$ interior bin boundaries $b_k$ ($k \in \{1, \dots, 13\}$, spanning $0.01$ to $75.0$\,mm/h), where $E_k(x, t) = P(\text{precip} \ge b_k)$ is computed as the tail sum:
\begin{equation}
E_k(x, t) = \sum_{j=k}^{13} p_j(x, t).
\end{equation}
By definition, $1 \ge E_1 \ge E_2 \ge \dots \ge E_{13} \ge 0$; if none of these thresholds is crossed, the pixel is assigned a precipitation rate of $0$\,mm/h.

Applying a fixed probability threshold of $0.5$ across all bins corresponds to extracting the (interpolated) median of the predicted distribution. However, because the predicted distributions are heavily skewed toward dry and light-rain bins, exceedance probabilities rarely reach $0.5$ (especially at higher rain rates), so the median systematically underestimates both rain occurrence and intensity. Instead, a set of calibrated probability decision thresholds $\boldsymbol{\tau} = [\tau_1, \tau_2, \dots, \tau_{13}]$ with $\tau_k \in (0, 1)$ is defined for each model variant (\Cref{tab:qpe_thresholds}). Within the 1--25\,mm/h range, these thresholds are chosen so that the reconstructed precipitation rate distribution approximately matches the GPM CORRA ground-truth distribution while maintaining distributional consistency between \pardignrt and \pardiglr (\Cref{sec:supp_consistency}). Outside this range, we keep the boundary threshold values constant ($\tau_k = \tau_{\text{1\,mm/h}}$ for $b_k < 1$\,mm/h, and $\tau_k = \tau_{\text{25\,mm/h}}$ for $b_k > 25$\,mm/h) as a reasonable first approximation: below 1\,mm/h, GPM is known to underestimate light drizzle \citep{hamada2016improvements, iguchi2024gpm}, whereas instantaneous rates above 25\,mm/h are sufficiently rare that direct distributional matching becomes noisy.

Many other options for translating the categorical probability distribution to a deterministic
rate exist (e.g., taking the distribution mean, or a fixed percentile). We leave the
investigation of such approaches, along with their potential benefits and drawbacks, as future work.

\begin{table}[htbp]
\centering
\footnotesize
\setlength{\tabcolsep}{3.5pt}
\caption{\textbf{Calibrated exceedance probability decision thresholds ($\tau_k$) for \pardig model variants.} Thresholds are chosen to align the reconstructed precipitation rate distribution with GPM CORRA observations across 1--25\,mm/h rain rates while maintaining distributional consistency between \pardignrt and \pardiglr.}
\label{tab:qpe_thresholds}
\begin{tabular*}{\textwidth}{@{\extracolsep{\fill}} l *{13}{c} @{}}
\toprule
\textbf{Variant} & \multicolumn{13}{c}{\textbf{Precipitation Rate Threshold $b_k$ (mm/h)}} \\
\cmidrule{2-14}
& 0.01 & 0.1 & 0.2 & 1.0 & 1.6 & 2.4 & 4.0 & 7.0 & 15.0 & 25.0 & 35.0 & 50.0 & 75.0 \\
\midrule
\pardignrt & 0.32 & 0.32 & 0.32 & 0.32 & 0.32 & 0.34 & 0.34 & 0.30 & 0.22 & 0.16 & 0.16 & 0.16 & 0.16 \\
\pardiglr  & 0.28 & 0.28 & 0.28 & 0.28 & 0.28 & 0.28 & 0.30 & 0.26 & 0.16 & 0.14 & 0.14 & 0.14 & 0.14 \\
\bottomrule
\end{tabular*}
\end{table}

\subsection{Deterministic Rate Reconstruction}

Given the computed exceedance probabilities $\{E_k\}_{k=1}^{13}$ and calibration
thresholds $\{\tau_k\}_{k=1}^{13}$ (\Cref{tab:qpe_thresholds}), the deterministic precipitation rate $r(x, t)$ is reconstructed as follows:
\begin{enumerate}[leftmargin=*, itemsep=2pt]
    \item \textbf{Highest exceeded threshold determination:} Identify the highest
    threshold index $k^*$ for which the exceedance probability equals or exceeds
    its calibrated decision threshold:
    \begin{equation}
    k^* = \max \left\{ k \in \{1, \dots, 13\} \;\middle|\; E_k(x, t) \ge \tau_k \right\}.
    \end{equation}
    Because $\tau_k$ is not fixed, a higher bin can occasionally exceed its threshold even if a lower bin does not. Taking the maximum index $k^*$ resolves this non-monotonicity. If no threshold is exceeded ($E_k < \tau_k$ for all $k \ge 1$), the pixel is classified as dry, and $r(x, t) = 0$\,mm/h.
    
    \item \textbf{Continuous interval interpolation:} When $1 \le k^* \le 12$, the continuous rate is
    computed by interpolating between bin boundary $b_{k^*}$ and the next boundary $b_{k^*+1}$.
    The interpolation weight $\alpha_{k^*}$ balances the surplus probability of
    bin $k^*$ ($\Delta_{\text{over}} = E_{k^*} - \tau_{k^*}$) against the probability
    deficit of the subsequent bin $k^*+1$ ($\Delta_{\text{under}} = \tau_{k^*+1} - E_{k^*+1}$):
    \begin{equation}
    \alpha_{k^*} = \frac{\Delta_{\text{over}}}{\Delta_{\text{over}} + \Delta_{\text{under}}} = \frac{E_{k^*} - \tau_{k^*}}{(E_{k^*} - \tau_{k^*}) + (\tau_{k^*+1} - E_{k^*+1})},
    \end{equation}
    yielding the reconstructed precipitation rate:
    \begin{equation}
    r(x, t) = b_{k^*} + \alpha_{k^*} \cdot (b_{k^*+1} - b_{k^*}).
    \end{equation}
    Intuitively, if the exceedance probability barely clears the threshold ($\Delta_{\text{over}} \approx 0$), $\alpha_{k^*} \approx 0$ and the rate sits near the lower boundary $b_{k^*}$; conversely, if the probability is close to triggering the next threshold ($\Delta_{\text{under}} \approx 0$), $\alpha_{k^*} \approx 1$ and the estimate approaches $b_{k^*+1}$.
    If the uppermost threshold ($k^* = 13$, $b_{13} = 75.0$\,mm/h) is exceeded, we do not interpolate further and simply cap the reconstructed instantaneous rate at $r(x, t) = 75.0$\,mm/h (note that point-level quantitative skill at such extreme instantaneous rates is limited in any case).

\end{enumerate}

\subsection{Sub-Hourly Inference and Temporal Accumulation}

Although the coordinate-query decoder is trained on GPM observations across $[t_0 - 60\,\text{min},\, t_0 + 60\,\text{min}]$ and can be sampled at arbitrary relative times within that window, at each hourly inference cycle initialized at $t_0$ we generate dense instantaneous precipitation fields at the four 15-minute offsets closest to $t_0$:
\begin{equation}
t \in \{t_0 - 15\,\text{min},\ t_0,\ t_0 + 15\,\text{min},\ t_0 + 30\,\text{min}\}.
\end{equation}
Querying this central 1-hour sub-window tiles successive hourly cycles continuously at 15-minute cadence while avoiding minor skill degradation toward the $\pm 60\,\text{min}$ edges of the training window. Between the two boundary candidates ($-30\,\text{min}$ and $+30\,\text{min}$), both yield nearly identical skill, so we select $+30\,\text{min}$ to extend each cycle's outputs 15\,minutes further forward in real time and reduce operational latency; retrieval skill remains uniform across all four selected offsets (\Cref{sec:supp_subhourly_offsets}).

To compute hourly precipitation accumulations ($R_{\text{1h}}$ in mm) for the 1-hourly evaluation,
\pardig averages four consecutive 15-minute instantaneous rates using a right-labeled convention:
\begin{equation}
R_{\text{1h}}(x, t_0) = \frac{1}{4}\bigl(r(x,\, t_0 - 45\,\text{min}) + r(x,\, t_0 - 30\,\text{min}) + r(x,\, t_0 - 15\,\text{min}) + r(x,\, t_0)\bigr),
\end{equation}
representing accumulated precipitation over the preceding interval $(t_0 - 1\,\text{h},\, t_0]$.\footnote{Other aggregation weights over the 15-minute instantaneous estimates can also be used; while all hourly evaluations here use this equal-weighted average over $\{-45, -30, -15, 0\}$\,min, we noted in later checks that trapezoidal weighting ($\tfrac{1}{8}[1, 2, 2, 2, 1]$ over $\{-60, -45, -30, -15, 0\}$\,min) can yield modestly better hourly scores.}
Because of this right-labeled definition, each hourly accumulation integrates
data across two successive inference cycles: the earlier two timestamps
($t_0 - 45\,\text{min}$ and $t_0 - 30\,\text{min}$) originate from the cycle
initialized at $t_0 - 1\,\text{h}$, while the latter two
($t_0 - 15\,\text{min}$ and $t_0$) originate from the cycle initialized at $t_0$.

\section{Model Architecture and Training Details}
\label{sec:supp_training}

\subsection{Graph-Mesh Transformer Architecture}

\pardig uses the same core encoder--processor--decoder architecture as a single step of WeatherNext~3 \citep{wn3}, but predicts marginal probability distributions directly over discrete intensity bins via cross-entropy in a single noise-free forward pass (rather than generating ensemble samples via Functional Generative Network noise conditioning, epistemic dropout, and CRPS), uses a smaller model capacity ($16$ mesh transformer layers with latent dimension $d = 512$, compared to $32$ layers with $d = 1024$ in WeatherNext~3, and width $512$ instead of $768$ in the convolutional and query-head networks), omits gridded and cyclone output heads while repurposing the station coordinate-query output head to predict 14-bin GPM CORRA precipitation marginals (conditioned at the query point only on coordinates and relative time offset rather than station elevation and land/sea metadata), and is trained from scratch in a single 20,000-step stage rather than through WeatherNext~3's multi-stage resolution and autoregressive rollout curriculum. To summarize, the core network operates on a level-6 refined icosahedral mesh ($40{,}962$ nodes) and processes data through three sequential stages:
\begin{enumerate}[leftmargin=*, itemsep=4pt]
    \item \textbf{Encoder (Grid $\to$ Mesh):} Regular latitude-longitude grid inputs at
    \quarterdegree (and \tenthdegree for \pardignrt) are mapped
    onto the icosahedral mesh using a single-step message-passing graph neural network (GNN). Each grid
    point connects to nearby mesh nodes via ball-query search with a radius of $0.6\times$ the longest mesh edge length, and edge features (encoded into a 32-dimensional latent vector)
    capture the 3D scalar and vector distances and relative latitude/longitude displacements between grid and mesh nodes. For the \pardignrt variant,
    a separate \tenthdegree grid encoder processes the higher-resolution surface and geostationary
    channels, providing additional spatial detail for convective-scale features.

    \item \textbf{Processor (Mesh $\to$ Mesh):} The mesh representation is refined through
    16 layers of sparse transformer attention ($4$ attention heads, head dimension $128$, feed-forward hidden dimension $2048$, no dropout). Each layer's attention spans a $k$-hop mesh neighborhood with $k = 2^{6-1} = 32$ hops (at a mean mesh edge length of ${\sim}$110\,km, corresponding to ${\sim}3{,}500$\,km per layer), giving each layer access to a synoptic-scale neighborhood and yielding a global receptive field across the 16 layers.

    \item \textbf{Decoder (Mesh $\to$ Latent Grid $\to$ Points):} Following WeatherNext~3, the processed
    mesh representation is first decoded back onto a latitude-longitude latent grid (\tenthdegree resolution for \pardignrt and \quarterdegree for \pardiglr) via a single-step GNN using in-triangle connectivity (where each grid point receives messages from the three vertices of its enclosing mesh triangle, with 32-dimensional edge encodings), combined
    with the corresponding grid-level skip connections from the encoder, and refined on the grid by a 4-layer convolutional neural network ($3 \times 3$ kernels, hidden dimension $512$). To predict precipitation at any continuous
    query coordinate $(x, t)$, the output head bilinearly interpolates this refined latent grid to the spatial
    location $x$, combines the interpolated features with an embedding of the relative query time
    offset, and maps the result through a 2-hidden-layer multilayer perceptron (MLP; hidden dimension $512$, Swish activations) to produce 14 bin logits (bounded to $(0, 10)$ via a scaled sigmoid before the softmax for training stability). During training, query coordinates are the sparse GPM footprint positions and times;
    at inference, the same output head is queried at every point on a regular \tenthdegree grid and
    15-minute offset, producing a dense global field.
\end{enumerate}

\subsection{Training Setup and Optimization}

\pardig is trained on Google Cloud TPU v5p accelerators (also used for near-real-time inference) with
a global batch size of 32 for 20,000 steps (${\sim}640,000$ training examples).
Models are trained with the Muon optimizer \citep{jordan2024muon} for weight matrices ($\ge$2D parameters) and Adam for biases and layer-norm scales (1D parameters), with $\beta_1 = 0.9$, $\beta_2 = 0.95$, weight decay 0.1, and global gradient norm clipping at 32.0. The learning rate is warmed up linearly over 1,000 steps to a peak of $10^{-3}$ and decayed to 0 following a cosine schedule. The objective is standard categorical cross-entropy over the 14 precipitation intensity bins against one-hot encoded GPM CORRA surface rain rates.

\section{Operational Pipeline and Latency}
\label{sec:supp_operational}
\suppressfloats[t]

A single \pardig inference pass at reference time $t_0$ produces four instantaneous
precipitation frames at $t_0 - 15$\,min, $t_0$, $t_0 + 15$\,min, and $t_0 + 30$\,min (\Cref{fig:figure_1}).
Each pass requires two categories of input:
(i)~four hourly HRES fields corresponding to
$t_0 - 2$\,h, $t_0 - 1$\,h, $t_0$, and $t_0 + 1$\,h (drawn from lead times 0--5\,h of the 6-hourly HRES forecast cycles;
see Section~\ref{sec:data}),
and
(ii)~eight SatWeave geostationary satellite mosaic frames
spanning $t_0 - 120$\,min to $t_0 + 90$\,min at 30-minute cadence.
The model outputs are binned probability histograms over precipitation
intensity; conversion to deterministic rain-rate estimates is a separate
post-processing step described in \Cref{sec:supp_qpe}.

\begin{figure}[!htb]
    \centering
    \includegraphics[width=0.88\linewidth]{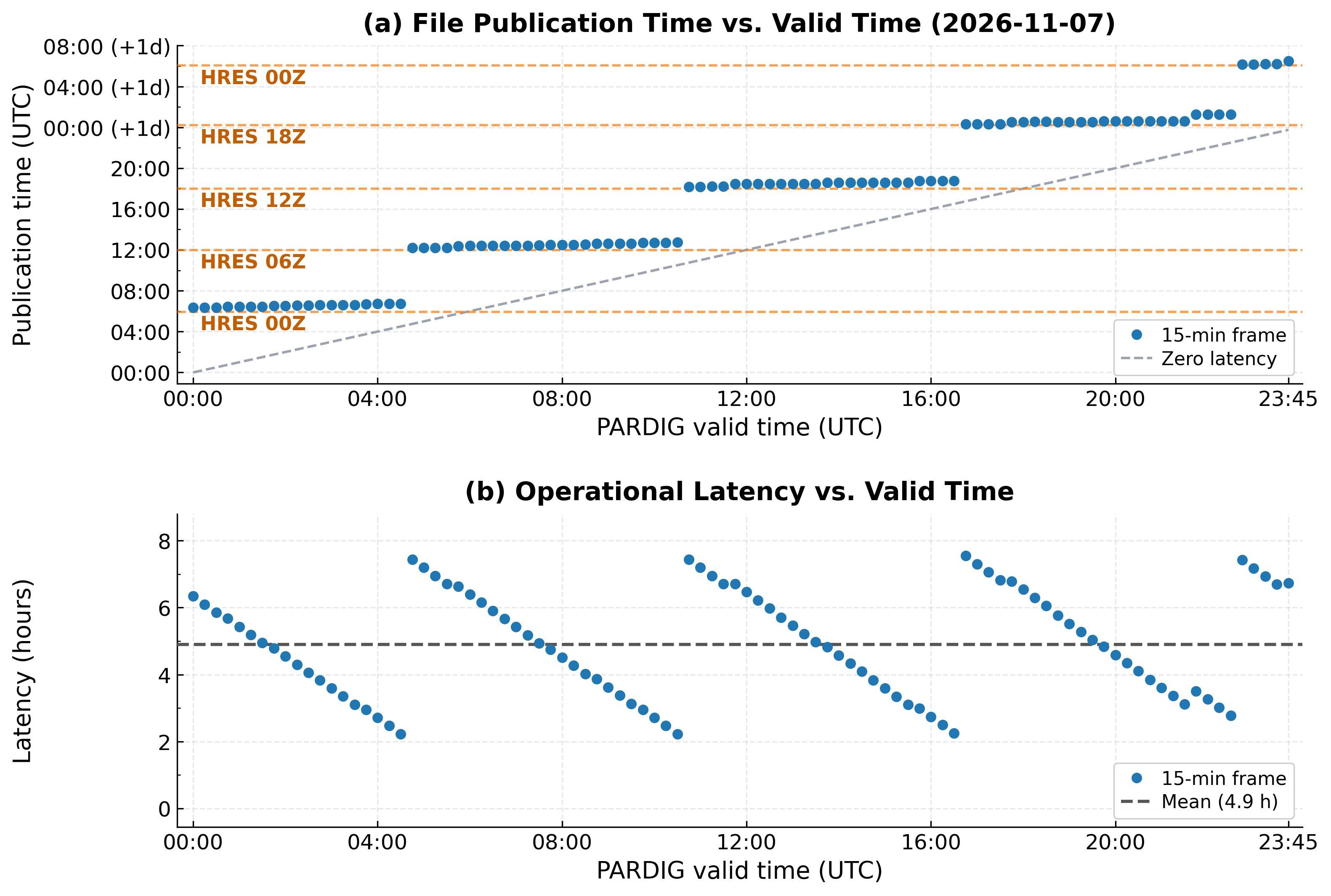}
    \caption{\textbf{Operational per-frame latency of \pardignrt on 7 November 2026.} (a)~Publication time in the public bucket versus valid time for all 96 15-minute frames, with 6-hourly HRES cycle publication times (orange dashed). (b)~Resulting latency versus valid time (mean 4.9~h); the small bump at 21:45--22:30\,UTC is a single inference run ($t_0 = \text{22:00}$\,UTC) that waited on slower-than-usual SatWeave processing.}
    \label{fig:supp_daily_latency}
\end{figure}

In the operational configuration, \pardignrt can run for any reference time $t_0$
the moment both required input streams become available, so operational
availability is governed by the four daily HRES forecast initialization cycles
(00, 06, 12, and 18\,UTC) and the latency of the SatWeave composites.
Once the inputs are present, the computation itself is fast: on a single Google Cloud TPU v5p chip, the forward pass for all four 15-minute global \tenthdegree fields takes ${\sim}3$~seconds of accelerator compute (${\sim}2\text{--}2.5$~minutes including loading model weights, reading the global HRES and SatWeave inputs, and writing the raw outputs), and with on-demand job scheduling, QPE post-processing, and Cloud Storage export, each run completes in roughly $10$~minutes end-to-end.
As an illustrative example, when the 06\,UTC HRES forecast becomes available
around 12:00\,UTC, SatWeave composites are typically available up to
approximately 11:30\,UTC. This, in combination with data from the previous HRES
cycle, provides all required inputs for the six consecutive reference times from
$t_0 = \text{05:00}$ through $t_0 = \text{10:00}$\,UTC, allowing six parallel inference passes to
produce $6 \times 4 = 24$ frames of 15-minute precipitation estimates covering
04:45 to 10:30\,UTC.
Taking all of the above into account, the end-to-end latency of \pardignrt (from
observation time to estimate availability) ranges from roughly 2 to 8~hours,
depending on where a given output falls within the HRES window. In principle,
lower latency could be achieved by using longer HRES forecast lead times (up to
12~hours from the preceding initialization cycle) alongside real-time SatWeave
imagery, trading some accuracy for faster availability.
Measured over a representative day (7 November 2026), per-frame latency from valid time to publication in the public bucket ranged from 2.2 to 7.6~hours (mean 4.9~hours), following a sawtooth set by the HRES publication times (\Cref{fig:supp_daily_latency}).

\FloatBarrier
\section{Ground truth data}
\label{sec:gt_data}

\subsection{MRMS}
We use the Multi-Radar Multi-Sensor (MRMS) Quantitative Precipitation Estimate Pass-2 product for independent evaluation \citep{zhang2016mrms}. This product provides hourly precipitation fields that have additionally been calibrated against rain gauges. We exclude areas with a low radar quality for evaluation to ensure we only evaluate against high-quality data. We subsample the native $\ang{0.01}$ data to $\ang{0.05}$ for computational reasons. For the sub-hourly evaluations in \Cref{sec:supp_subhourly}, we additionally use the MRMS Instantaneous Precipitation Rate product, which provides radar-only instantaneous surface precipitation rates; throughout the paper, we refer to the hourly gauge-adjusted Pass-2 product simply as MRMS and the instantaneous product as MRMS Instantaneous Precipitation Rate (or MRMS Rate).

\subsection{EURADCLIM}
EURADCLIM provides calibrated radar observations over Europe \citep{overeem2023euradclim}. The native ${\sim}2$\,km grid is subsampled to $\ang{0.04}$. At the time of evaluation, EURADCLIM was only available until the end of 2023.

\subsection{JMA Radar-AMeDAS}
JMA Radar-AMeDAS provides calibrated observations over Japan \citep{makihara1996method}. Although JMA issues 1-hour accumulations at 30-minute intervals, we retain only the non-overlapping top-of-the-hour analyses. The native ${\sim}1$\,km grid is subsampled to $\ang{0.05}$.

\subsection{Rain gauges}
Rain gauge data over 2018--2023 was downloaded from Synoptic's\footnote{\url{https://synopticdata.com/}} derived precipitation service, which aggregates multi-network surface observations (including national meteorological services and citizen weather station networks) and accumulates measurements to 1-hour intervals. Because multi-network station archives include short-lived deployments, co-located duplicate feeds, event-only reporting sensors (which transmit only when precipitation occurs and omit dry hours), and stations with misconfigured metadata coordinates, we apply five quality-control filters defined strictly on the station metadata and ground-truth observation counts (starting from $46{,}298$ stations with valid coordinates):
\begin{enumerate}[leftmargin=*, itemsep=2pt]
    \item \textbf{Minimum positive and negative event counts:} At least 50 positive ($\ge 1$\,mm/h) and at least 50 negative ($< 1$\,mm/h) hourly observations ($37{,}718$ stations retained), ensuring sufficient sample size for per-station CSI and excluding dry-locked or purely event-triggered gauges.
    \item \textbf{Land and island mask:} Located on or within $\ang{0.10}$ (${\sim}10$\,km, to accommodate coastal stations and rounded coordinates) of Natural Earth\footnote{\url{https://www.naturalearthdata.com/}} 10m land or minor-island polygons, and excluding $(0, 0)$ placeholder coordinates ($37{,}674$ stations retained; $44$ open-ocean/mislocated stations removed, such as amateur stations with sign-flipped longitudes).
    \item \textbf{Coordinate deduplication:} For co-located station feeds sharing identical coordinates rounded to $\ang{0.001}$ (${\sim}100$\,m), only the single station feed with the largest number of valid hourly observations is retained ($37{,}485$ stations retained; $189$ duplicate feeds removed).
    \item \textbf{Minimum record length:} At least 750 valid hourly measurements (${>}\,1$\,month of observations; $37{,}009$ stations retained; $476$ short-lived stations removed).
    \item \textbf{Conservative upper bound on wet-hour fraction:} At most $50\%$ of valid hourly observations exceeding the $1$\,mm/h threshold ($N_{\text{pos}} / N_{\text{valid}} \le 0.50$; $36{,}678$ stations retained; $331$ gauges that report primarily during wet periods removed).
\end{enumerate}
For the $4$\,mm/h per-station CSI cumulative distribution function (\Cref{fig:supp_station_csi}c), applying the same sample-size and wet-fraction criteria at the $4$\,mm/h threshold ($\ge 50$ positive, $\ge 50$ negative, and $\le 50\%$ positive fraction at $4$\,mm/h) retains $26{,}957$ stations.

\FloatBarrier

\section{Global Precipitation Distributions and \pardignrt--\pardiglr Consistency}
\label{sec:supp_consistency}

\subsection{Overlap Period Consistency (2016--2023)}
\label{sec:supp_overlap_consistency}

\begin{figure}[t]
    \centering
    \includegraphics[width=\linewidth]{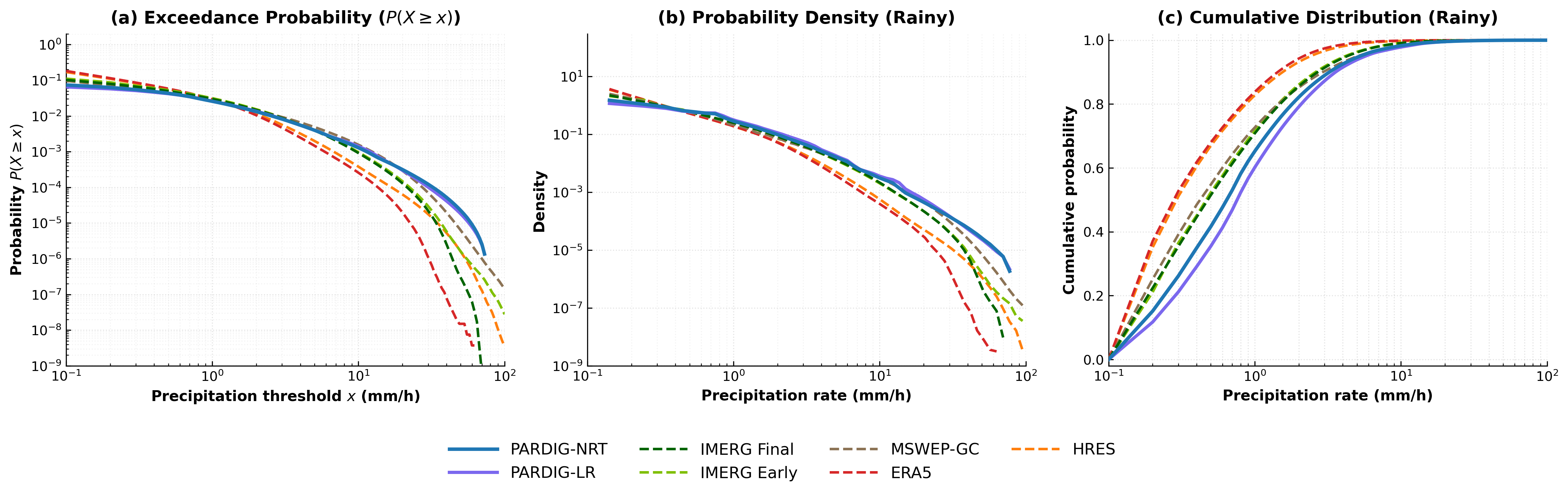}
    \caption{\textbf{Global hourly precipitation distributions (2016--2023).} Area-weighted distributions across $\ang{85}\text{S}$--$\ang{85}\text{N}$ for \pardignrt, \pardiglr, and baselines: (\textbf{a})~unconditional exceedance probability $P(X \ge x)$, and (\textbf{b})~probability density function (PDF) and (\textbf{c})~cumulative distribution function (CDF) conditioned on rainy cells ($\ge 0.1$\,mm/h). Summary statistics and percentiles are given in \Cref{tab:distribution_stats}.}
    \label{fig:supp_distribution_consistency}
\end{figure}

To compare global hourly precipitation distributions across \pardignrt, \pardiglr, and the external baselines over their shared 2016--2023 period, we sample 400 hourly timestamps uniformly at random across 2016--2023 over $\ang{85}\text{S}$--$\ang{85}\text{N}$ (${\sim}99.6\%$ of Earth's surface area, avoiding polar-cap missing values in IMERG). All summary statistics, percentiles ($p_{90}$--$p_{99.999}$), and distribution curves weight grid cells by $\cos(\text{latitude})$ so that values represent fractions of surface area rather than raw grid-cell counts. \Cref{tab:distribution_stats} reports the resulting summary statistics and upper-tail percentiles, and \Cref{fig:supp_distribution_consistency} compares the unconditional exceedance probability $P(X \ge x)$ alongside the rainy-cell ($\ge 0.1$\,mm/h) PDF and CDF.

\textbf{Distributional agreement between \pardignrt and \pardiglr.} The two variants show close agreement across all summary statistics (\Cref{tab:distribution_stats}): area-weighted global mean precipitation rates differ by less than 2\%, area-weighted rainy percentages are within one percentage point, and upper-tail percentiles align tightly through $p_{99.999}$. As shown in \Cref{fig:supp_distribution_consistency}, the exceedance, PDF, and CDF curves overlap closely from light rain ($0.1$\,mm/h) through extreme convective intensities ($>50$\,mm/h), confirming that the two variants map disparate input streams into closely aligned precipitation distributions.

\begin{table}[htbp]
\centering
\small
\setlength{\tabcolsep}{3.5pt}
\caption{\textbf{Area-weighted global precipitation distribution statistics and percentiles} (2016--2023 record, 1h accumulation). All statistics, rainy percentages ($\ge 0.1$\,mm/h), and percentiles ($p_{90}$--$p_{99.999}$) are $\cos(\text{lat})$ area-weighted across $\ang{85}\text{S}$--$\ang{85}\text{N}$; rates in mm/h.}
\label{tab:distribution_stats}
\begin{tabular}{@{}lccccccccc@{}}
\toprule
Product & Mean & Std & Rainy\,\% & $p_{90}$ & $p_{95}$ & $p_{99}$ & $p_{99.9}$ & $p_{99.99}$ & $p_{99.999}$ \\
\midrule
\pardignrt & 0.108 & 0.874 & 7.4\% & 0.00 & 0.37 & 2.55 & 11.39 & 32.36 & 58.9 \\
\pardiglr & 0.106 & 0.866 & 6.5\% & 0.00 & 0.32 & 2.60 & 12.00 & 30.14 & 57.0 \\
IMERG Final & 0.114 & 0.714 & 10.2\% & 0.10 & 0.45 & 2.70 & 9.60 & 21.50 & 34.0 \\
IMERG Early & 0.117 & 0.735 & 11.1\% & 0.10 & 0.50 & 2.80 & 9.80 & 22.39 & 36.7 \\
MSWEP-GC & 0.121 & 0.871 & 9.6\% & 0.09 & 0.40 & 2.81 & 12.42 & 27.39 & 45.8 \\
ERA5 & 0.123 & 0.484 & 18.2\% & 0.24 & 0.60 & 2.05 & 5.87 & 13.61 & 22.6 \\
HRES & 0.124 & 0.560 & 16.9\% & 0.23 & 0.57 & 2.18 & 6.77 & 16.84 & 34.8 \\
\bottomrule
\end{tabular}
\end{table}

\textbf{Comparison with other precipitation products.} Across product families (\Cref{tab:distribution_stats} and \Cref{fig:supp_distribution_consistency}), ERA5 and HRES predict rain ($\ge 0.1$\,mm/h) over $17\text{--}18\%$ of global area, more than double \pardig ($6.5\text{--}7.4\%$) and well above MSWEP-GC and IMERG ($9.6\text{--}11.1\%$), and concentrate $83\text{--}84\%$ of rainy area below $1$\,mm/h (\Cref{fig:supp_distribution_consistency}c), reflecting the well-known NWP drizzle bias. In the upper tail (\Cref{fig:supp_distribution_consistency}a,b), \pardig retains the heaviest extremes ($p_{99.999} \approx 57\text{--}59$\,mm/h), followed by MSWEP-GC ($46$\,mm/h), IMERG ($34\text{--}37$\,mm/h), and the ECMWF models ($23\text{--}35$\,mm/h).

\subsection{Multi-Decadal Homogeneity and Observing-System Evolution (2001--2025)}
\label{sec:supp_long_term_trends}

\begin{figure}[htbp]
    \centering
    \includegraphics[width=\linewidth]{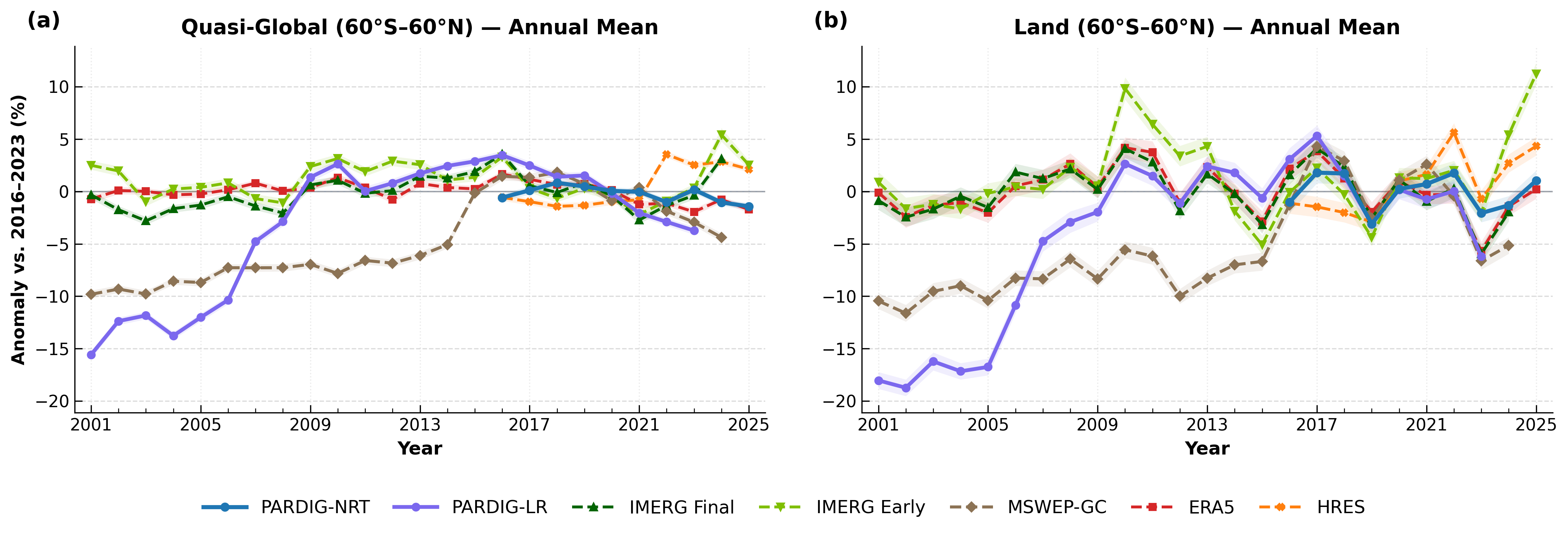}
    \caption{\textbf{Multi-decadal precipitation stability across satellite eras (2001--2025).} Annual mean precipitation anomalies ($\pm 1$ standard error of the mean across $400$ shared seasonally and diurnally stratified hourly slots per complete year, $\cos(\text{lat})$ area-weighted) relative to each product's 2016--2023 overlap mean: (\textbf{a})~Quasi-global ($\ang{60}\text{S}$--$\ang{60}\text{N}$) and (\textbf{b})~Land ($\ang{60}\text{S}$--$\ang{60}\text{N}$). While ERA5, IMERG~V07, \pardignrt, and HRES remain broadly stable across their records, \pardiglr (prior to 2008) and MSWEP-GC (prior to 2015) substantially underestimate early-period means relative to 2016--2023.}
    \label{fig:supp_long_term_trends}
\end{figure}

While \pardiglr and \pardignrt match closely across their 2016--2023 overlap (\Cref{tab:distribution_stats}), extending precipitation records across multiple decades spans major changes in the global observing system, including expanding passive microwave constellations, geostationary imager replacements, and the 2014 transition from TRMM to GPM. Maintaining long-term homogeneity across these transitions is a well-documented challenge across satellite and reanalysis products \citep{sun2018review, portier2025imerg_faq}. \Cref{fig:supp_long_term_trends} evaluates all seven products over complete calendar years from 2001 to 2025 quasi-globally ($\ang{60}\text{S}$--$\ang{60}\text{N}$) and over land, sampling $400$ shared seasonally and diurnally stratified hourly slots per year ($\cos(\text{lat})$ area-weighted) relative to each product's 2016--2023 mean.

ERA5, IMERG~V07 (both Final and Early), \pardignrt, and HRES maintain stable annual means throughout their examined records---spanning the full multi-decadal period back to 2001 for ERA5 and IMERG~V07, and 2016--2025 for \pardignrt and HRES. In IMERG~V07, this multi-decadal stability reflects explicit inter-sensor calibration across the TRMM and GPM eras alongside distribution-restoration of morphed fields via SHARPEN \citep{tan2021sharpen, huffman2023imerg, huffman2024imerg_notes}, with IMERG Final additionally applying monthly GPCC rain-gauge scaling over land.

By contrast, MSWEP-GC and \pardiglr exhibit clear shifts in the earlier part of the record. MSWEP-GC is $7\text{--}10\%$ lower across 2001--2013 both quasi-globally and over land before stepping upward across 2014--2015, a GPM-era input transition shift that is explicitly documented in the MSWEP~V3 release notes \citep{wang2026mswep, gloh2o2025mswep}. \pardiglr remains broadly stationary from 2008 through 2023, with its 2008--2013 average within $1\%$ of the 2016--2023 mean and no discontinuity at the 2014 start of its GPM training record. Prior to 2008, however, the \pardiglr mean is $13\%$ lower quasi-globally ($17\%$ over land) during 2001--2005 before ramping to its stationary level across 2006--2008. \pardiglr applies no explicit cross-era homogenization, and we hypothesize that this pre-2008 underestimation primarily reflects the sparser passive microwave coverage and older geostationary sensors of the early 2000s (alongside, as an additional factor, training and calibrating the model on the post-2014 GPM era). Multi-decadal trend analyses should therefore account for these observing-system transitions, particularly prior to 2008 in \pardiglr and 2015 in MSWEP-GC.

\FloatBarrier

\section{Physical Consistency: Diurnal and Seasonal Cycles}
\label{sec:diurnal_seasonal}

\begin{figure}[htbp]
    \centering
    \includegraphics[width=\linewidth, height=0.74\textheight, keepaspectratio]{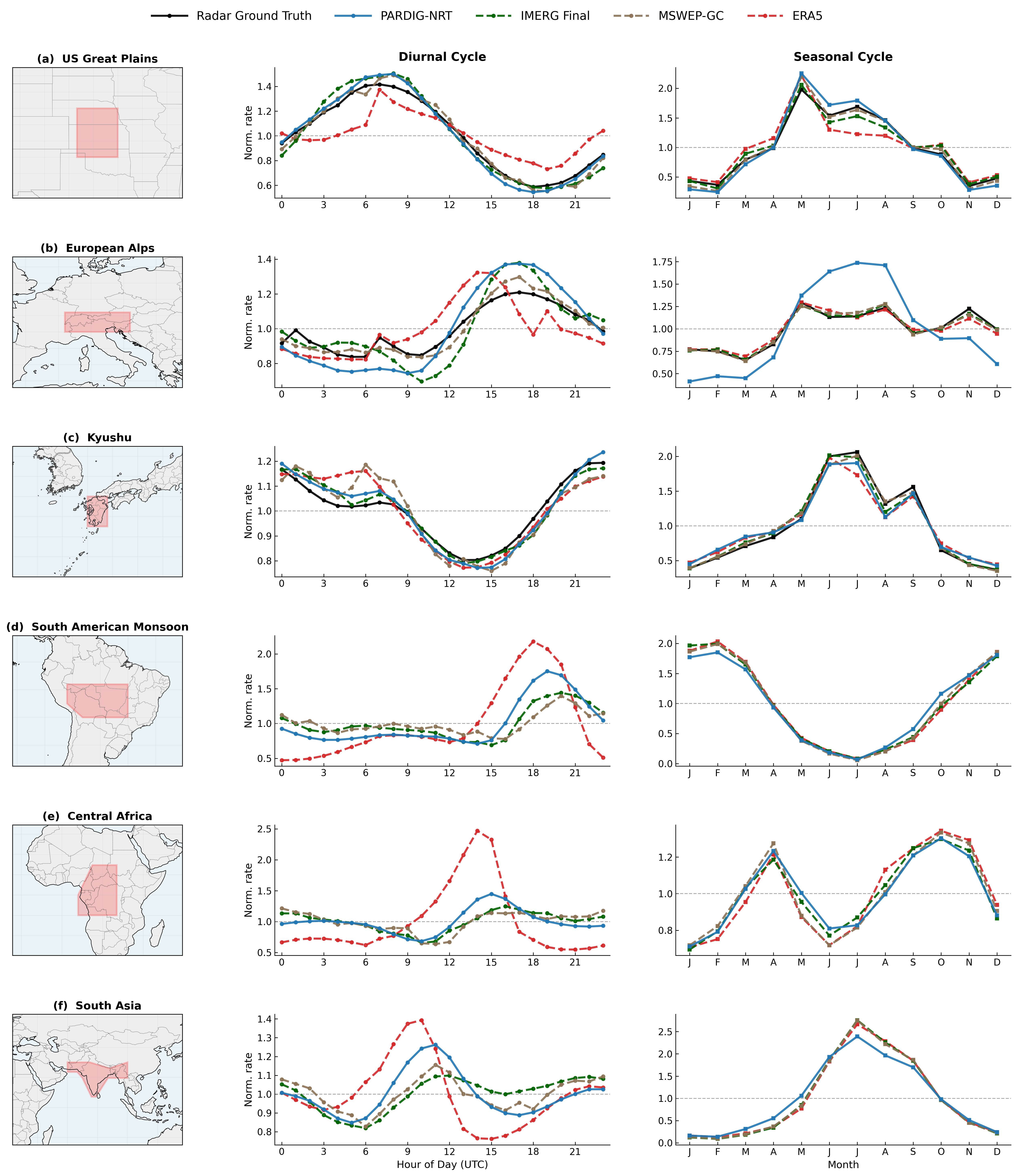}
    \caption{\textbf{Physical consistency across diurnal and seasonal precipitation cycles.} Evaluation of normalized diurnal cycles (middle column, hourly mean rate divided by regional mean) and seasonal cycles (right column, monthly mean rate divided by regional mean) across six climatologically diverse regions. Left column shows the locator map for each region (red boxes for custom regional domains \textbf{a}--\textbf{c}; red polygons for IPCC AR6 reference regions \textbf{d}--\textbf{f} \citep{iturbide2020update}): (\textbf{a}) US Great Plains (evaluated against MRMS radar ground truth), (\textbf{b}) European Alps (evaluated against EURADCLIM radar ground truth), (\textbf{c}) Kyushu, Japan (evaluated against JMA Radar-AMeDAS radar ground truth), and three AR6 regions without dense radar ground truth: (\textbf{d}) South American Monsoon (SAM), (\textbf{e}) Central Africa (CAF), and (\textbf{f}) South Asia (SAS). Curves show radar ground truth (solid black, where available), \pardignrt (solid blue), IMERG Final V07 (dashed green), MSWEP-GC (dashed brown), and ERA5 (dashed red). Dashed horizontal lines mark the normalized regional mean rate ($1.0$).}
    \label{fig:supp_diurnal_seasonal}
\end{figure}

\Cref{fig:supp_diurnal_seasonal} evaluates whether \pardig reproduces known physical patterns of precipitation
variability, namely the diurnal cycle (driven by solar heating and land-sea contrasts) and the seasonal
cycle (driven by large-scale circulation and monsoon dynamics), across six
climatologically distinct regions.

\textbf{Regions with radar ground truth.} Over the US Great Plains, \pardig correctly reproduces the well-documented nocturnal convective peak (${\sim}$3--9 UTC,
corresponding to late evening through early morning local time), driven by the nocturnal
low-level jet and eastward-propagating mesoscale convective systems \citep{higgins1997influence, carbone2008rainfall}, closely tracking
the MRMS radar observations. IMERG Final and MSWEP-GC show a similar phase and amplitude,
whereas ERA5 misses the early-night buildup (2--6~UTC) and overpredicts daytime rates, reflecting known diurnal-cycle biases in models with parameterized convection \citep{bechtold2014representing, covey2016metrics}.
Over the European Alps, MSWEP-GC (supervised on EURADCLIM) closely tracks EURADCLIM in diurnal phase and amplitude; \pardig and IMERG Final capture the timing of the late-afternoon peak (whereas ERA5 peaks 2--3 hours early) but overestimate its amplitude, with \pardig underpredicting nighttime and early-morning rates.
In the normalized seasonal cycle, IMERG Final, MSWEP-GC, and ERA5 closely match EURADCLIM across all months, whereas \pardig exhibits an exaggerated seasonal amplitude, underestimating wintertime precipitation relative to summertime precipitation.
While the cause of this winter-to-summer imbalance remains to be verified, one hypothesis is that GPM DPR's reduced sensitivity to snowfall \citep{chase2022direct} in the GPM CORRA training targets may cause \pardig to underpredict alpine winter precipitation.
Over Kyushu, all products reproduce the East Asian monsoon seasonality (June--July \emph{Baiu} peak, August dip, and September typhoon peak), and \pardig and IMERG Final closely match JMA Radar-AMeDAS's diurnal cycle (peaking near 23 UTC), whereas ERA5 and MSWEP-GC exhibit an exaggerated afternoon peak around 06 UTC.

\textbf{AR6 regions without radar ground truth.} We complement the radar-evaluated
regions with three tropical and subtropical IPCC AR6 regions \citep{iturbide2020update} that exhibit well-documented
precipitation regimes but lack the dense radar networks needed for direct
verification: the South American Monsoon (SAM), Central Africa (CAF), and South
Asia (SAS). These regions are important because they are home to billions of people, represent a large fraction
of the global precipitation budget and are precisely where satellite-based estimates
are most relied upon.

Across all three regions, the seasonal cycle is consistently well captured by all
products: the SAM wet season (October--March), the bimodal Central African seasonality \citep{nicholson2018itcz}, and the South Asian monsoon peak (June--September)
are all reproduced with close agreement in both phase and amplitude.

The diurnal cycle shows a consistent ordering across all three regions. ERA5 produces an early and strongly amplified afternoon peak, consistent with its known tendency to trigger parameterized convection prematurely over land. \pardignrt peaks roughly 1\,hour later than ERA5 and 1--2\,hours earlier than IMERG Final, with an intermediate amplitude between the sharper ERA5 signal and the broader, more muted IMERG Final and MSWEP-GC curves.

\FloatBarrier

\section{Zonal-Mean Precipitation Climatology}
\label{sec:supp_zonal}

\begin{figure}[htbp]
    \centering
    \includegraphics[width=\linewidth]{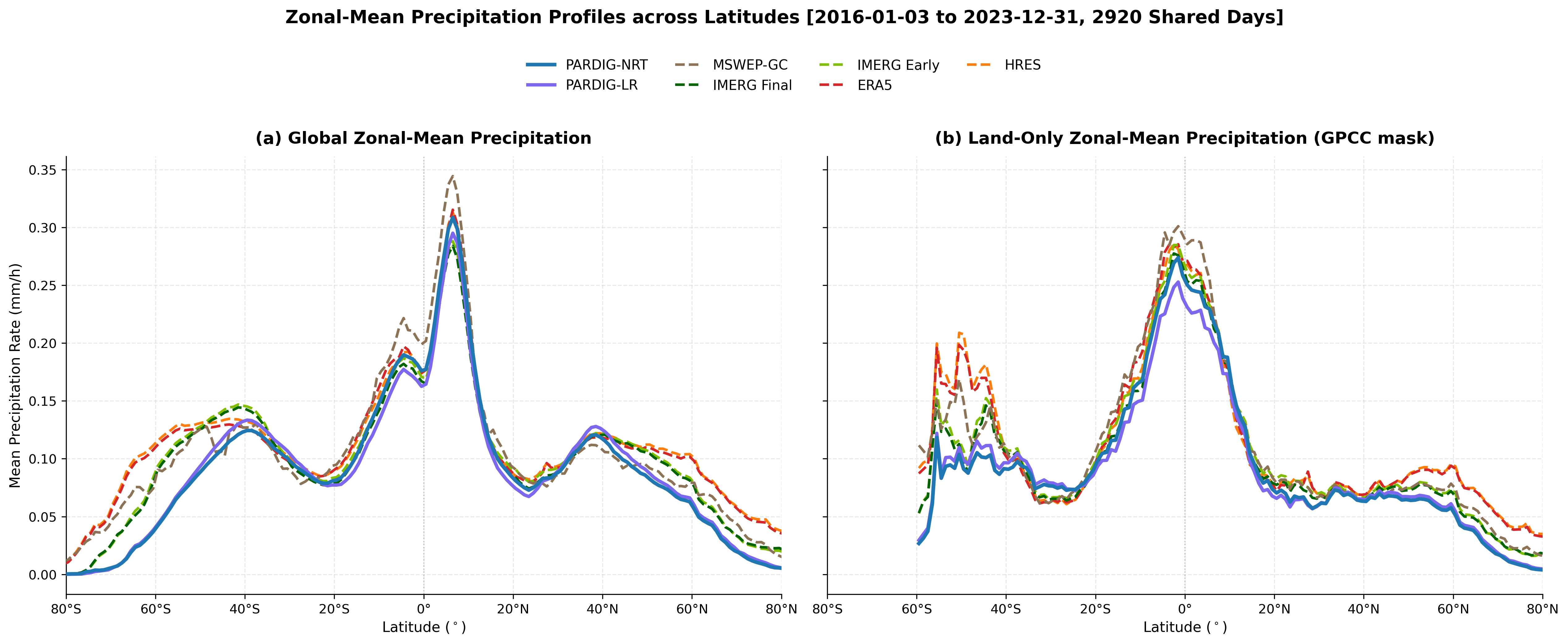}
    \caption{\textbf{Zonal-mean precipitation profiles across latitudes} for \pardignrt, \pardiglr, MSWEP-GC, IMERG Final, IMERG Early, ERA5, and HRES, computed over 2{,}920 shared days (2016--2023). (a)~Global zonal mean across all longitudes. (b)~Land-only zonal mean, masked using GPCC gauge coverage at 1\textdegree{} resolution. All products are regridded to the common 1\textdegree{} GPCC grid before zonal averaging.}
    \label{fig:supp_zonal_latitude}
\end{figure}

To assess whether \pardig reproduces the expected meridional structure of global precipitation, \Cref{fig:supp_zonal_latitude} presents zonal-mean precipitation profiles as a continuous function of latitude for \pardignrt, \pardiglr, MSWEP-GC, IMERG Final, IMERG Early, ERA5, and HRES over 2{,}920 shared days (2016--2023). All products are regridded to a common 1\textdegree{} grid and averaged zonally across all longitudes (panel~a) and across land-only grid cells (panel~b, using GPCC gauge coverage to define the land mask).

All evaluated products agree closely on the latitudinal positions of the major global circulation features: the Intertropical Convergence Zone (ITCZ) maximum near 6--8\textdegree N, the secondary Southern Hemisphere tropical peak near 5\textdegree S, the subtropical dry zones near $\pm$20--25\textdegree, and the secondary mid-latitude storm-track peaks near $\pm$40\textdegree. Across the tropics and subtropics (40\textdegree S--40\textdegree N), \pardignrt tracks IMERG, ERA5, and HRES closely in both the global and land-only profiles, while MSWEP-GC estimates higher rates at both the ITCZ and Southern Hemisphere tropical peaks globally (panel~a). Over equatorial land (panel~b), \pardignrt, IMERG, ERA5, and HRES peak closely together around 0.27--0.29\,mm/h, with MSWEP-GC slightly higher and \pardiglr somewhat lower (${\sim}$0.25\,mm/h), a difference that is much smaller in the ocean-dominated global profile.

Poleward of the mid-latitude storm tracks (${\sim}$50\textdegree), the products stratify into three distinct groups across both panels: ERA5 and HRES maintain the highest high-latitude precipitation rates, MSWEP-GC, IMERG Final, and IMERG Early occupy an intermediate range, and \pardignrt and \pardiglr estimate the lowest mean rates toward the poles. In the land-only profile between 40\textdegree S and 60\textdegree S (panel~b), the curves are noticeably noisier and show sharp spikes (particularly in ERA5 and HRES) because the vast majority of this latitude belt is ocean, leaving only a small number of land grid cells per latitude band across southern South America, Tasmania, and New Zealand.

While the exact causes of the lower \pardig values poleward of ${\sim}$50\textdegree{} are uncertain, we hypothesize several plausible contributing factors across two distinct latitude bands. Within the high-latitude portion of the GPM training domain (${\sim}$50--67\textdegree), geostationary satellites observe at increasingly oblique viewing angles, reducing the information content of the infrared imagery, and, as discussed in \Cref{sec:diurnal_seasonal}, GPM DPR's reduced sensitivity to light rain and frozen hydrometeors may cause \pardig to underestimate high-latitude precipitation and snowfall. Poleward of ${\sim}$67\textdegree, the GPM DPR training target is absent, meaning \pardig must extrapolate into polar regions without direct supervision while relying on degraded inputs, and any underestimation learned within ${\sim}$50--67\textdegree{} would carry over into this extrapolation regime. At the same time, the substantial gap between the ECMWF models (ERA5/HRES) and both IMERG and MSWEP-GC across most high latitudes underscores that existing reference and operational baselines also disagree in these regions, where rigorous evaluation is challenging.

\FloatBarrier

\section{Additional Evaluation Metrics}
\label{sec:supp_metrics}

\begin{figure}[!htbp]
    \centering
    \includegraphics[width=\linewidth]{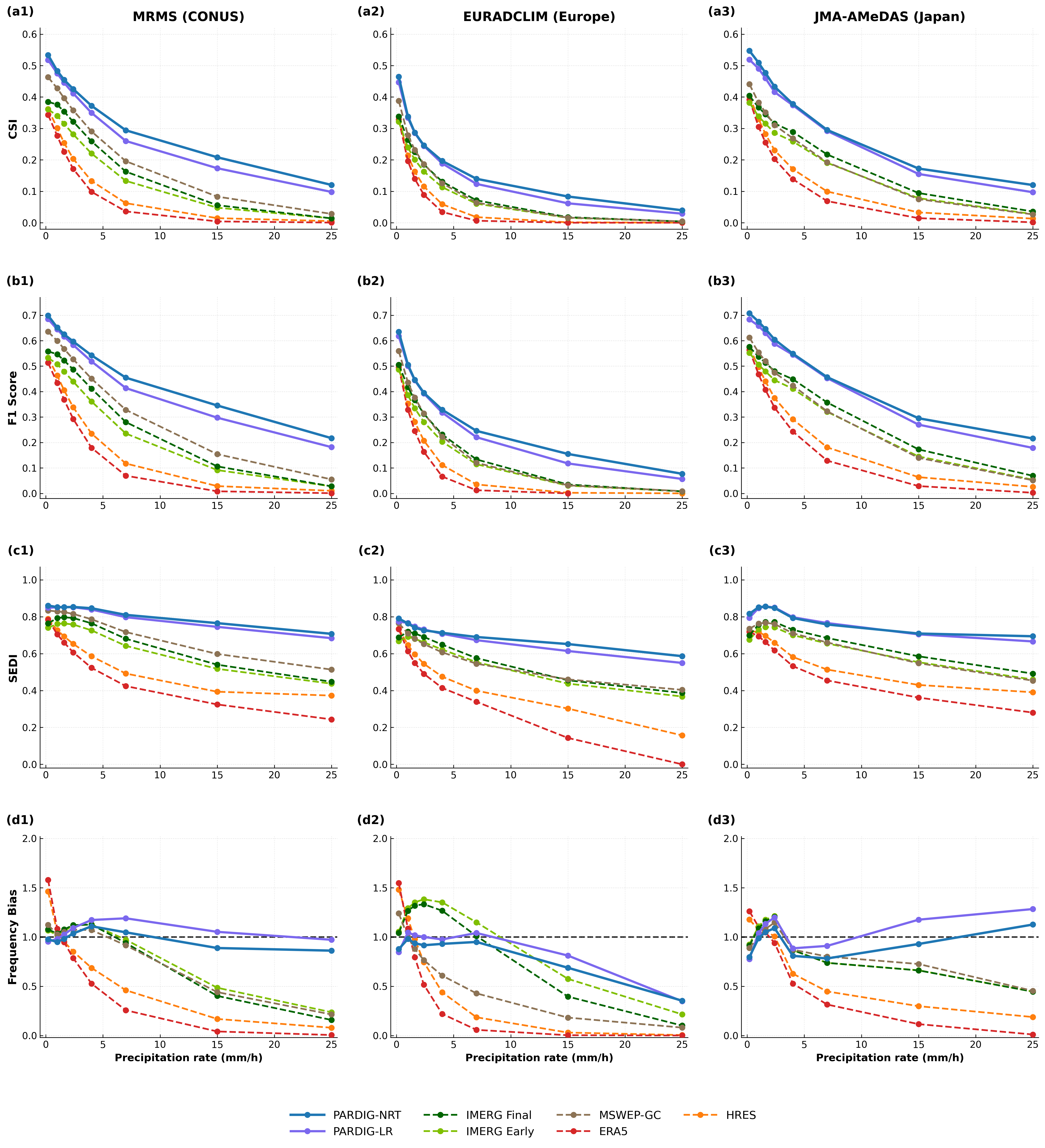}
    \caption{\textbf{Additional categorical verification metrics vs precipitation threshold} across independent radar networks: MRMS (CONUS, column 1), EURADCLIM (Europe, column 2), and JMA Radar-AMeDAS (Japan, column 3).
    (a1--a3) Critical Success Index (CSI), including the \pardiglr companion dataset (numerical values in \Cref{tab:csi_all}).
    (b1--b3) $\text{F}_1$ Score (harmonic mean of precision and recall; paired precision and recall values in \Cref{tab:pr_all}).
    (c1--c3) Symmetric Extremal Dependence Index (SEDI).
    (d1--d3) Frequency Bias ($\text{FBIAS} = 1$ dashed line indicates unbiased frequency).}
    \label{fig:supp_all_metrics}
\end{figure}

This section extends the main-text verification (\Cref{sec:eval_setup}) with additional categorical and spatial metrics across the three radar networks (MRMS over CONUS, EURADCLIM over Europe, and JMA Radar-AMeDAS over Japan), extreme-event case studies, and global surface stations. For the multi-year hourly radar evaluations, \Cref{fig:supp_all_metrics} summarizes CSI, $\text{F}_1$ score, SEDI, and frequency bias across the three radar domains, with \pardiglr included alongside \pardignrt and the external baselines.

\subsection{CSI with \pardiglr}
\suppressfloats[t]

Figure~\ref{fig:supp_all_metrics}(a1--a3) presents the Critical Success Index across precipitation thresholds for all seven products, replicating the main-text evaluation (Figure~2) with the addition of the \pardiglr companion dataset, and \Cref{tab:csi_all} reports numerical CSI values at selected thresholds. Across all three radar domains, \pardiglr exhibits high skill that closely tracks \pardignrt, outperforming all external baselines (MSWEP-GC, IMERG Final, IMERG Early, ERA5, and HRES) at every hourly threshold.

\begin{table}[!htb]
\centering
\small
\setlength{\tabcolsep}{4.5pt}
\caption{\textbf{CSI at selected precipitation thresholds across independent radar networks}. Evaluations for hourly accumulations over the respective multi-year records against MRMS (CONUS), EURADCLIM (Europe), and JMA Radar-AMeDAS (Japan). Best values per column in bold.}
\label{tab:csi_all}
\begin{tabular}{@{}lccccccc@{}}
\toprule
Product & 0.2\,mm/h & 1\,mm/h & 4\,mm/h & 7\,mm/h & 15\,mm/h & 25\,mm/h & 50\,mm/h \\
\midrule
\multicolumn{8}{@{}l}{\textbf{(a) MRMS (CONUS)}} \\
\midrule
\pardignrt & \textbf{0.53} & \textbf{0.48} & \textbf{0.37} & \textbf{0.29} & \textbf{0.21} & \textbf{0.12} & \textbf{0.03} \\
\pardiglr & 0.52 & \textbf{0.48} & 0.35 & 0.26 & 0.17 & 0.10 & 0.02 \\
IMERG Final & 0.38 & 0.38 & 0.26 & 0.16 & 0.06 & 0.01 & 0.00 \\
IMERG Early & 0.36 & 0.34 & 0.22 & 0.13 & 0.05 & 0.01 & 0.00 \\
MSWEP-GC & 0.46 & 0.43 & 0.29 & 0.20 & 0.08 & 0.03 & 0.00 \\
ERA5 & 0.34 & 0.28 & 0.10 & 0.04 & 0.00 & 0.00 & 0.00 \\
HRES & 0.36 & 0.30 & 0.13 & 0.06 & 0.01 & 0.01 & 0.00 \\
\midrule
\multicolumn{8}{@{}l}{\textbf{(b) EURADCLIM (Europe)}} \\
\midrule
\pardignrt & \textbf{0.46} & \textbf{0.34} & \textbf{0.20} & \textbf{0.14} & \textbf{0.08} & \textbf{0.04} & \textbf{0.01} \\
\pardiglr & 0.45 & 0.33 & 0.19 & 0.12 & 0.06 & 0.03 & \textbf{0.01} \\
IMERG Final & 0.34 & 0.26 & 0.13 & 0.07 & 0.02 & 0.00 & 0.00 \\
IMERG Early & 0.32 & 0.24 & 0.11 & 0.06 & 0.02 & 0.00 & 0.00 \\
MSWEP-GC & 0.39 & 0.28 & 0.13 & 0.06 & 0.02 & 0.00 & 0.00 \\
ERA5 & 0.33 & 0.20 & 0.03 & 0.01 & 0.00 & 0.00 & 0.00 \\
HRES & 0.34 & 0.21 & 0.06 & 0.02 & 0.00 & 0.00 & 0.00 \\
\midrule
\multicolumn{8}{@{}l}{\textbf{(c) JMA Radar-AMeDAS (Japan)}} \\
\midrule
\pardignrt & \textbf{0.55} & \textbf{0.51} & \textbf{0.38} & \textbf{0.30} & \textbf{0.17} & \textbf{0.12} & \textbf{0.04} \\
\pardiglr & 0.52 & 0.49 & 0.37 & 0.29 & 0.16 & 0.10 & 0.03 \\
IMERG Final & 0.40 & 0.37 & 0.29 & 0.22 & 0.09 & 0.04 & 0.00 \\
IMERG Early & 0.38 & 0.34 & 0.26 & 0.19 & 0.08 & 0.03 & 0.00 \\
MSWEP-GC & 0.44 & 0.38 & 0.27 & 0.19 & 0.07 & 0.03 & 0.00 \\
ERA5 & 0.39 & 0.31 & 0.14 & 0.07 & 0.01 & 0.00 & 0.00 \\
HRES & 0.40 & 0.33 & 0.17 & 0.10 & 0.03 & 0.01 & 0.00 \\
\bottomrule
\end{tabular}
\end{table}

\subsection{F1 Score, Precision, and Recall}
\suppressfloats[t]

The $\text{F}_1$ score is the harmonic mean of precision ($\text{TP}/(\text{TP}+\text{FP})$) and recall ($\text{TP}/(\text{TP}+\text{FN})$):
\begin{equation}
\text{F}_1 = \frac{2\,\text{TP}}{2\,\text{TP} + \text{FP} + \text{FN}}.
\end{equation}
Because achieving a high $\text{F}_1$ requires both high precision and high recall, it penalizes products that achieve high precision by underforecasting, or high recall by overforecasting.

Figure~\ref{fig:supp_all_metrics}(b1--b3) displays $\text{F}_1$ across thresholds. \pardignrt achieves the highest $\text{F}_1$ at every threshold across all three radar networks, with substantial margins over IMERG Final (e.g., $+0.13$ at $4$\,mm/h and $+0.19$ at $25$\,mm/h against MRMS). \Cref{tab:pr_all} decomposes performance at selected thresholds into paired precision/recall values: for instance, against EURADCLIM at $4$\,mm/h, IMERG Final achieves $0.21\,/\,0.26$ (precision/recall) while \pardignrt reaches $0.34\,/\,0.32$, improving both detection completeness and prediction accuracy simultaneously.

\begin{table}[!htb]
\centering
\footnotesize
\setlength{\tabcolsep}{2.5pt}
\caption{\textbf{Precision / Recall at selected precipitation thresholds across independent radar networks}. Values formatted as Precision / Recall for MRMS (CONUS), EURADCLIM (Europe), and JMA Radar-AMeDAS (Japan). Best $\text{F}_1$ per column in bold.}
\label{tab:pr_all}
\begin{tabular}{@{}lccccccc@{}}
\toprule
Product & 0.2\,mm/h & 1\,mm/h & 4\,mm/h & 7\,mm/h & 15\,mm/h & 25\,mm/h & 50\,mm/h \\
\midrule
\multicolumn{8}{@{}l}{\textbf{(a) MRMS (CONUS)}} \\
\midrule
\pardignrt & \textbf{0.71\,/\,0.69} & \textbf{0.67\,/\,0.63} & \textbf{0.52\,/\,0.57} & \textbf{0.44\,/\,0.47} & \textbf{0.37\,/\,0.33} & \textbf{0.24\,/\,0.20} & \textbf{0.06\,/\,0.06} \\
\pardiglr & 0.71\,/\,0.67 & 0.65\,/\,0.64 & 0.48\,/\,0.56 & 0.38\,/\,0.45 & 0.29\,/\,0.30 & 0.19\,/\,0.18 & 0.05\,/\,0.06 \\
IMERG Final & 0.54\,/\,0.58 & 0.54\,/\,0.56 & 0.39\,/\,0.44 & 0.29\,/\,0.27 & 0.19\,/\,0.07 & 0.11\,/\,0.02 & 0.03\,/\,0.00 \\
IMERG Early & 0.52\,/\,0.55 & 0.50\,/\,0.51 & 0.34\,/\,0.38 & 0.24\,/\,0.23 & 0.14\,/\,0.07 & 0.08\,/\,0.02 & 0.01\,/\,0.00 \\
MSWEP-GC & 0.60\,/\,0.67 & 0.59\,/\,0.61 & 0.44\,/\,0.47 & 0.34\,/\,0.31 & 0.25\,/\,0.11 & 0.16\,/\,0.03 & 0.03\,/\,0.00 \\
ERA5 & 0.42\,/\,0.66 & 0.42\,/\,0.45 & 0.26\,/\,0.14 & 0.17\,/\,0.04 & 0.11\,/\,0.00 & 0.08\,/\,0.00 & --- \\
HRES & 0.45\,/\,0.65 & 0.45\,/\,0.47 & 0.29\,/\,0.20 & 0.19\,/\,0.09 & 0.10\,/\,0.02 & 0.08\,/\,0.01 & 0.03\,/\,0.00 \\
\midrule
\multicolumn{8}{@{}l}{\textbf{(b) EURADCLIM (Europe)}} \\
\midrule
\pardignrt & \textbf{0.68\,/\,0.60} & \textbf{0.51\,/\,0.50} & \textbf{0.34\,/\,0.32} & \textbf{0.25\,/\,0.24} & \textbf{0.19\,/\,0.13} & \textbf{0.15\,/\,0.05} & \textbf{0.08\,/\,0.01} \\
\pardiglr & 0.67\,/\,0.57 & 0.49\,/\,0.51 & 0.32\,/\,0.31 & 0.22\,/\,0.23 & 0.13\,/\,0.11 & 0.11\,/\,0.04 & 0.07\,/\,0.01 \\
IMERG Final & 0.50\,/\,0.52 & 0.37\,/\,0.47 & 0.21\,/\,0.26 & 0.13\,/\,0.13 & 0.06\,/\,0.02 & 0.04\,/\,0.00 & 0.01\,/\,0.00 \\
IMERG Early & 0.47\,/\,0.50 & 0.34\,/\,0.44 & 0.18\,/\,0.24 & 0.11\,/\,0.12 & 0.04\,/\,0.02 & 0.02\,/\,0.00 & 0.00\,/\,0.00 \\
MSWEP-GC & 0.51\,/\,0.63 & 0.44\,/\,0.44 & 0.29\,/\,0.18 & 0.20\,/\,0.09 & 0.10\,/\,0.02 & 0.06\,/\,0.00 & 0.02\,/\,0.00 \\
ERA5 & 0.41\,/\,0.63 & 0.32\,/\,0.34 & 0.19\,/\,0.04 & 0.12\,/\,0.01 & 0.02\,/\,0.00 & 0.00\,/\,0.00 & --- \\
HRES & 0.42\,/\,0.62 & 0.33\,/\,0.39 & 0.19\,/\,0.08 & 0.12\,/\,0.02 & 0.04\,/\,0.00 & 0.02\,/\,0.00 & 0.00\,/\,0.00 \\
\midrule
\multicolumn{8}{@{}l}{\textbf{(c) JMA Radar-AMeDAS (Japan)}} \\
\midrule
\pardignrt & \textbf{0.80\,/\,0.64} & \textbf{0.68\,/\,0.67} & \textbf{0.61\,/\,0.50} & \textbf{0.52\,/\,0.41} & \textbf{0.31\,/\,0.28} & \textbf{0.21\,/\,0.23} & \textbf{0.06\,/\,0.13} \\
\pardiglr & 0.78\,/\,0.61 & 0.65\,/\,0.67 & 0.58\,/\,0.51 & 0.48\,/\,0.43 & 0.25\,/\,0.29 & 0.16\,/\,0.20 & 0.04\,/\,0.10 \\
IMERG Final & 0.60\,/\,0.55 & 0.52\,/\,0.56 & 0.48\,/\,0.42 & 0.42\,/\,0.31 & 0.22\,/\,0.14 & 0.11\,/\,0.05 & 0.06\,/\,0.00 \\
IMERG Early & 0.58\,/\,0.53 & 0.48\,/\,0.53 & 0.44\,/\,0.38 & 0.38\,/\,0.28 & 0.18\,/\,0.12 & 0.09\,/\,0.04 & 0.02\,/\,0.00 \\
MSWEP-GC & 0.65\,/\,0.58 & 0.55\,/\,0.55 & 0.46\,/\,0.40 & 0.36\,/\,0.29 & 0.17\,/\,0.12 & 0.08\,/\,0.04 & 0.01\,/\,0.00 \\
ERA5 & 0.51\,/\,0.64 & 0.45\,/\,0.49 & 0.35\,/\,0.19 & 0.27\,/\,0.08 & 0.14\,/\,0.02 & 0.13\,/\,0.00 & --- \\
HRES & 0.53\,/\,0.62 & 0.48\,/\,0.51 & 0.38\,/\,0.24 & 0.29\,/\,0.13 & 0.14\,/\,0.04 & 0.08\,/\,0.02 & 0.01\,/\,0.00 \\
\bottomrule
\end{tabular}
\end{table}

\subsection{SEDI}

The Symmetric Extremal Dependence Index (SEDI) \citep{ferro2011extremal} assesses performance for rare and extreme events without the base-rate degeneracy that causes standard scores (such as CSI or Heidke Skill Score) to approach zero as the event frequency tends to zero:
\begin{equation}
\text{SEDI} = \frac{\ln F - \ln H - \ln(1-F) + \ln(1-H)}{\ln F + \ln H + \ln(1-F) + \ln(1-H)},
\end{equation}
where $H = \text{TP}/(\text{TP} + \text{FN})$ is the hit rate and $F = \text{FP}/(\text{FP} + \text{TN})$ is the false alarm rate. A perfect prediction yields $\text{SEDI} = 1$, random forecasting yields $\text{SEDI} = 0$, and negative values indicate performance worse than random chance.

Figure~\ref{fig:supp_all_metrics}(c1--c3) shows SEDI as a function of threshold. Both \pardignrt and \pardiglr maintain high extremal skill ($\text{SEDI} > 0.55\text{--}0.8$) across the full threshold spectrum up to $25$\,mm/h, demonstrating robust capability in capturing intense precipitation events where baseline products degrade significantly.

\subsection{Frequency Bias}

The Frequency Bias Index measures the ratio of estimated to observed event frequency:
\begin{equation}
\text{FBIAS} = \frac{\text{TP} + \text{FP}}{\text{TP} + \text{FN}},
\end{equation}
where $\text{FBIAS} = 1$ indicates an unbiased estimate of event frequency, values above 1 indicate overestimation (overforecasting precipitation occurrences), and values below 1 indicate underestimation (missed precipitation).

Figure~\ref{fig:supp_all_metrics}(d1--d3) shows frequency bias as a function of precipitation threshold. Both \pardignrt and \pardiglr maintain near-unity frequency bias ($\approx 0.8\text{--}1.3$) across the full threshold range up to $25$\,mm/h over CONUS and Japan, and up to $7$\,mm/h over Europe (where they decline at heavier rates but remain closer to unity than all baselines). In contrast, the numerical weather prediction models (ERA5 and HRES) overestimate frequency at the lowest threshold ($0.2$\,mm/h) and exhibit severe underestimation at higher rain rates, with frequency bias dropping rapidly toward zero beyond 5\,mm/h. MSWEP-GC and both IMERG variants (Early and Final) are less biased at light-to-moderate rain rates before progressively underforecasting at heavy intensities ($\ge 15$\,mm/h).

\subsection{Case Study Verification and Additional Events with \pardiglr and MSWEP-GC}
\label{sec:supp_case_study}

\begin{figure}[htbp]
    \centering
    \makebox[\linewidth][c]{\includegraphics[width=1.08\linewidth, height=0.80\textheight, keepaspectratio]{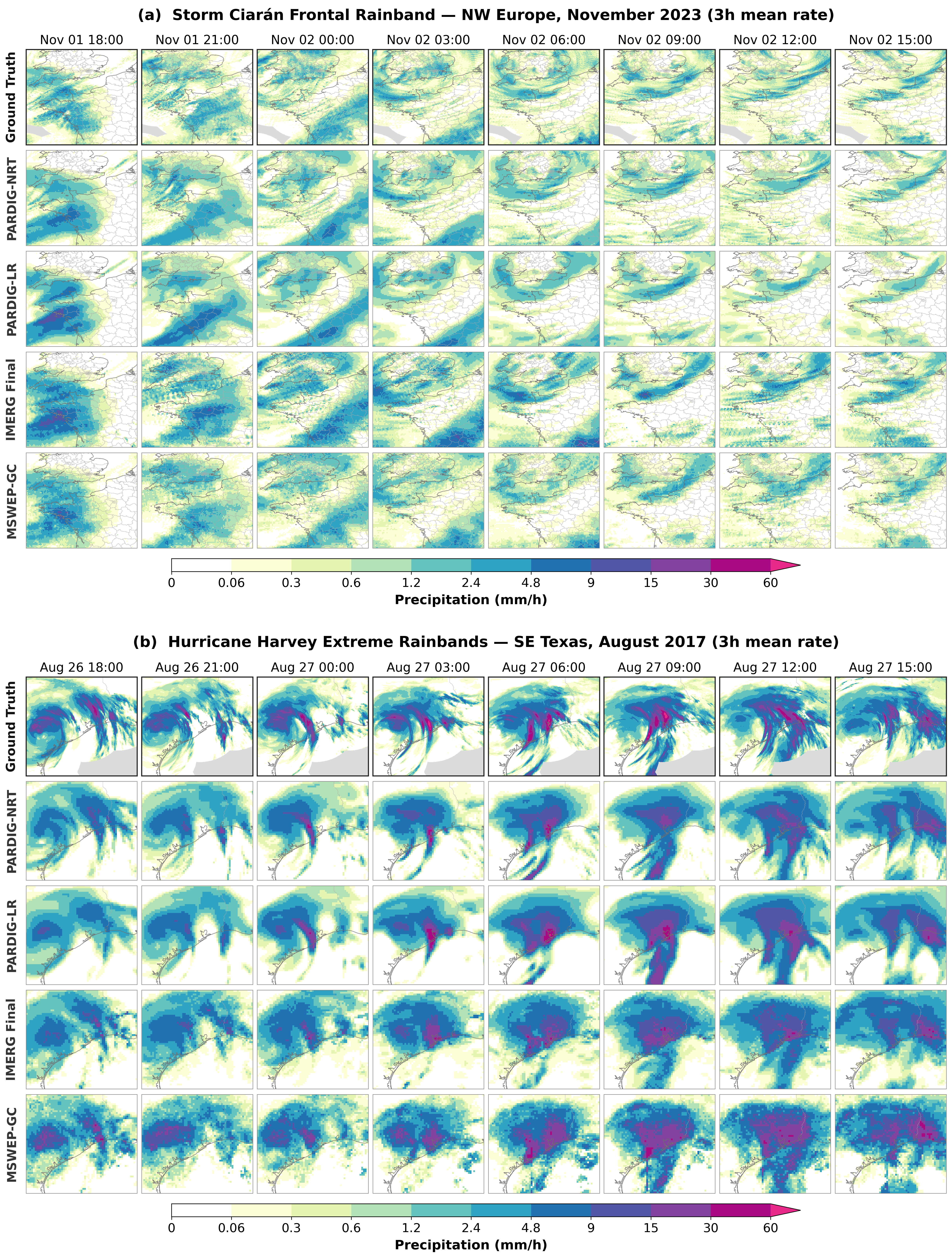}}
    \caption{\textbf{Additional case studies comparing \pardignrt, \pardiglr, IMERG Final, and MSWEP-GC.}
    Sequential 8-step progressions (3h mean rate) for (\textbf{a}) Storm Ciar\'an frontal rainband over northwestern Europe (1--2 November 2023, EURADCLIM) and (\textbf{b}) Hurricane Harvey extreme rainbands over southeast Texas (26--27 August 2017, MRMS). Rows display gauge-adjusted radar ground truth (top, bold frame), \pardignrt (second row), \pardiglr (third row), IMERG Final V07 (fourth row), and MSWEP-GC (bottom row) in mm/h at their native spatial resolutions. Column headers denote accumulation interval start times in UTC ($[t, t + \Delta t)$); light grey shading in the ground-truth panels indicates regions outside radar coverage. Numerical CSI verification across thresholds is detailed in \Cref{tab:case_study_csi}.}
    \label{fig:supp_case_study}
\end{figure}

\Cref{fig:supp_case_study} presents two additional extreme precipitation case studies from the 2016--2023 overlap period where \pardignrt, \pardiglr, IMERG Final, and MSWEP-GC are all available: (\textbf{a}) the Storm Ciar\'an extratropical frontal rainband over northwestern Europe (November 2023, 3h mean rate, evaluated against EURADCLIM) and (\textbf{b}) Hurricane Harvey's stalling rainbands over southeast Texas (August 2017, 3h mean rate, evaluated against MRMS). Visually, both \pardig variants capture similar synoptic and mesoscale rainband geometry, with \pardignrt exhibiting finer small-scale spatial detail. During Storm Ciar\'an, both reconstruct a coherent frontal rainband and wrap-around occlusion without the cross-track swath striping visible in IMERG Final, and during Hurricane Harvey both resolve sharp coastal feeder bands; MSWEP-GC also captures the broad rainband structure in both events, though with noticeable pixel-level speckling across the precipitation shield.

\Cref{tab:case_study_csi} reports the event-level CSI across all 8 sequential time steps for the three main-text case studies (\Cref{fig:figure_4}) alongside the two events in \Cref{fig:supp_case_study}. In Storm Ciar\'an, \pardignrt leads at $0.2$--$1$\,mm/h while MSWEP-GC is highest at $4$--$7$\,mm/h; in Hurricane Harvey, both \pardignrt and \pardiglr lead across $0.2$--$4$\,mm/h (and \pardiglr at $7$\,mm/h), while MSWEP-GC achieves the highest CSI in the heaviest core ($15$--$25$\,mm/h).

\begin{table}[htbp]
\centering
\footnotesize
\setlength{\tabcolsep}{2.2pt}
\caption{\textbf{Critical Success Index (CSI) for extreme precipitation case studies.} Evaluated across all 8 sequential time steps for each event against independent radar composites (\Cref{sec:gt_data}); ``--'' indicates thresholds not exceeded in either the radar observations or the product estimates during the event.}
\label{tab:case_study_csi}
\begin{tabular}{@{}llccccccc@{}}
\toprule
Event & Product & Accum & 0.2\,mm/h & 1\,mm/h & 4\,mm/h & 7\,mm/h & 15\,mm/h & 25\,mm/h \\
\midrule
Texas July 2025 (MRMS) & \pardignrt & 1h & \textbf{0.68} & \textbf{0.65} & \textbf{0.50} & \textbf{0.37} & \textbf{0.29} & \textbf{0.10} \\
 & IMERG Final & 1h & 0.51 & 0.50 & 0.38 & 0.25 & 0.08 & 0.02 \\
\midrule
Slovenia Aug 2023 (EURADCLIM) & \pardignrt & 3h & \textbf{0.86} & \textbf{0.67} & \textbf{0.38} & \textbf{0.22} & \textbf{0.08} & \textbf{0.06} \\
 & IMERG Final & 3h & 0.78 & 0.56 & 0.32 & 0.16 & 0.03 & 0.00 \\
\midrule
Japan Aug 2024 (JMA Radar-AMeDAS) & \pardignrt & 6h & \textbf{0.79} & \textbf{0.64} & \textbf{0.46} & \textbf{0.38} & \textbf{0.12} & 0.00 \\
 & IMERG Final & 6h & 0.73 & 0.56 & 0.43 & 0.36 & 0.09 & 0.00 \\
\midrule
Storm Ciar\'an Nov 2023 (EURADCLIM) & \pardignrt & 3h & \textbf{0.74} & \textbf{0.52} & 0.06 & 0.00 & -- & -- \\
 & \pardiglr & 3h & 0.69 & 0.48 & 0.07 & 0.01 & -- & -- \\
 & IMERG Final & 3h & 0.68 & 0.43 & 0.07 & 0.01 & -- & -- \\
 & MSWEP-GC & 3h & 0.71 & 0.49 & \textbf{0.10} & \textbf{0.02} & -- & -- \\
\midrule
Hurricane Harvey Aug 2017 (MRMS) & \pardignrt & 3h & \textbf{0.80} & \textbf{0.73} & \textbf{0.59} & 0.41 & 0.21 & 0.14 \\
 & \pardiglr & 3h & \textbf{0.80} & \textbf{0.73} & \textbf{0.60} & \textbf{0.45} & 0.25 & 0.15 \\
 & IMERG Final & 3h & 0.76 & 0.66 & 0.50 & 0.39 & 0.21 & 0.05 \\
 & MSWEP-GC & 3h & 0.77 & 0.67 & 0.51 & 0.42 & \textbf{0.26} & \textbf{0.19} \\
\bottomrule
\end{tabular}
\end{table}

\noindent\textit{Note:} CSI values across events with different accumulation windows are not directly comparable, because the same mean-rate threshold corresponds to different total accumulations and therefore different climatological rarities. For example, a 6-hour mean rate of 25\,mm/h implies a 150\,mm total over the window, an event that is rarer than a 1-hour total of 25\,mm. Consequently, the rapid decline of CSI at high thresholds for the 3h and 6h events partly reflects the increasing rarity of extreme mean rates at longer accumulation windows.

\subsection{Fractions Skill Score (FSS)}
\label{sec:supp_fss}

Grid-point categorical metrics such as CSI, $\text{F}_1$, and SEDI evaluate exact cell-by-cell correspondence and can penalize high-resolution precipitation fields twice when small spatial displacements occur (once as a false alarm and once as a missed event). To evaluate spatial skill across relaxed neighborhood tolerances, we compute the Fractions Skill Score (FSS) \citep{roberts2008scale}. For a given precipitation threshold and square spatial neighborhood of width $s$, let $f_p(x, y)$ and $f_o(x, y)$ denote the fraction of grid cells within the $s \times s$ window centered at $(x, y)$ where predicted and observed precipitation exceed the threshold. FSS compares the mean squared error of these neighborhood fractions against a worst-case reference:
\begin{equation}
\text{FSS}(s) = 1 - \frac{\frac{1}{N}\sum_{x,y}\bigl(f_p(x, y) - f_o(x, y)\bigr)^2}{\frac{1}{N}\sum_{x,y}\bigl(f_p(x, y)^2 + f_o(x, y)^2\bigr)},
\end{equation}
ranging from $0$ (complete mismatch) to $1$ (perfect neighborhood agreement), with $\text{FSS} \approx 0.5 + \bar{f}_o/2 \approx 0.5$ (where $\bar{f}_o$ is the domain-wide wet-pixel fraction) commonly used as a rough reference for useful spatial skill \citep{roberts2008scale}.

\begin{figure}[htbp]
    \centering
    \includegraphics[width=\linewidth]{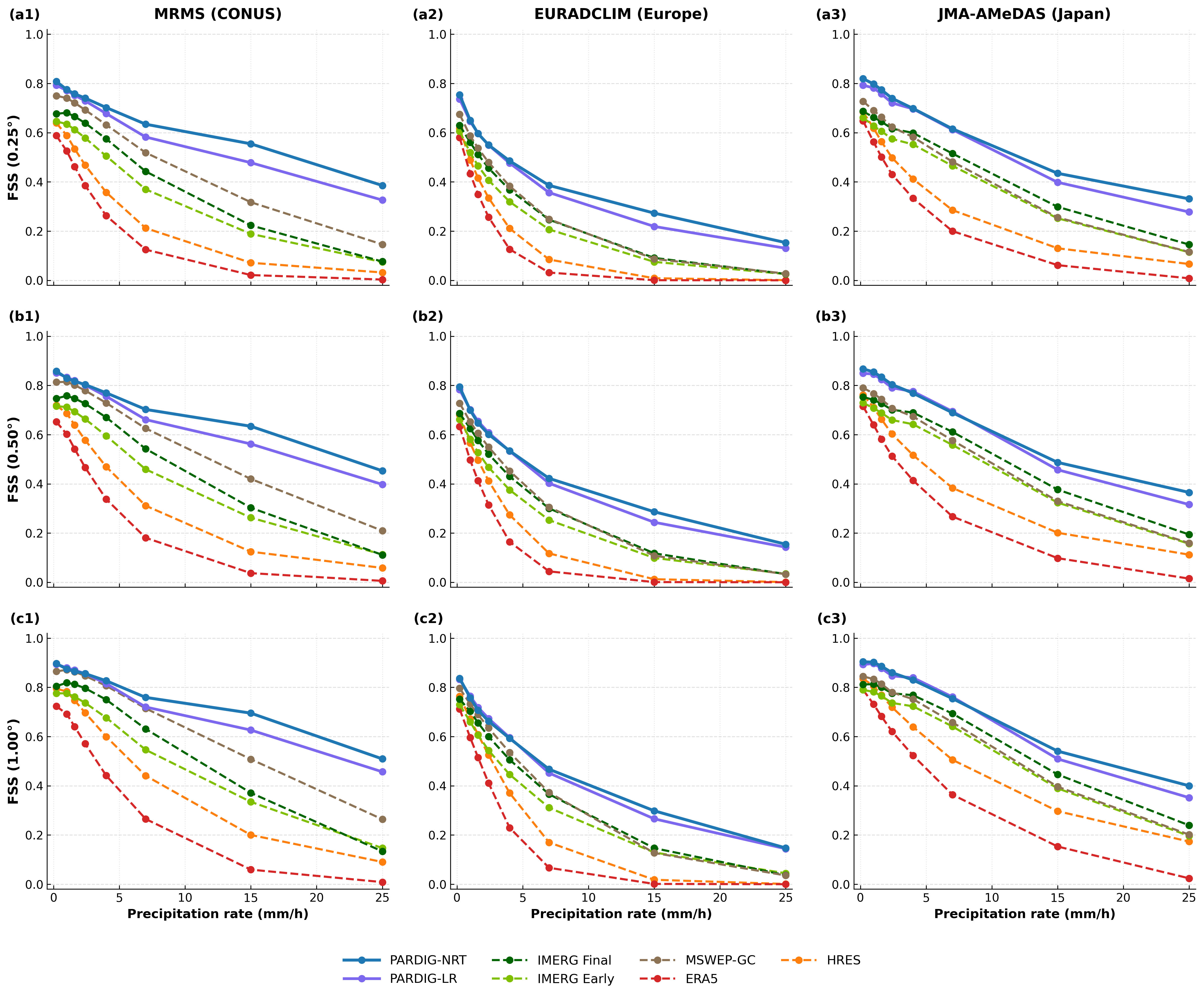}
    \caption{\textbf{Fractions Skill Score (FSS) vs precipitation threshold across spatial neighborhood scales} evaluated over 2016--2023 against MRMS (CONUS, column 1), EURADCLIM (Europe, column 2), and JMA Radar-AMeDAS (Japan, column 3).
    (a1--a3) $\ang{0.25}$ neighborhood.
    (b1--b3) $\ang{0.5}$ neighborhood.
    (c1--c3) $\ang{1}$ neighborhood.}
    \label{fig:supp_fss}
\end{figure}

\Cref{fig:supp_fss} presents FSS across precipitation thresholds ($0.2$--$25$\,mm/h) for three neighborhood scales: $\ang{0.25}$ (${\sim}25$\,km), $\ang{0.5}$ (${\sim}50$\,km), and $\ang{1}$ (${\sim}100$\,km). Across all three radar networks and spatial scales, \pardignrt and \pardiglr outperform every baseline across nearly all thresholds (with only isolated ties at light-to-moderate rates at coarser scales). Relaxing the spatial tolerance up to $\ang{1}$ (${\sim}100$\,km) also does not close the heavy-rain gap across CONUS, Europe, or Japan: over CONUS at $15$\,mm/h, for example, \pardignrt reaches $0.55$ at $\ang{0.25}$ and $0.70$ at $\ang{1}$, whereas MSWEP-GC ($0.51$) and IMERG Final ($0.37$) remain substantially lower even at $\ang{1}$ ($0.26$ and $0.13$ vs.\ $0.51$ at $25$\,mm/h). Thus, the baselines' lower grid-point CSI at heavy rain rates is not merely a double-penalty artifact of small spatial displacements.

\subsection{Global Station-Level CSI Across Thresholds}
\label{sec:supp_station_csi}

\begin{figure}[htbp]
    \centering
    \includegraphics[width=\linewidth]{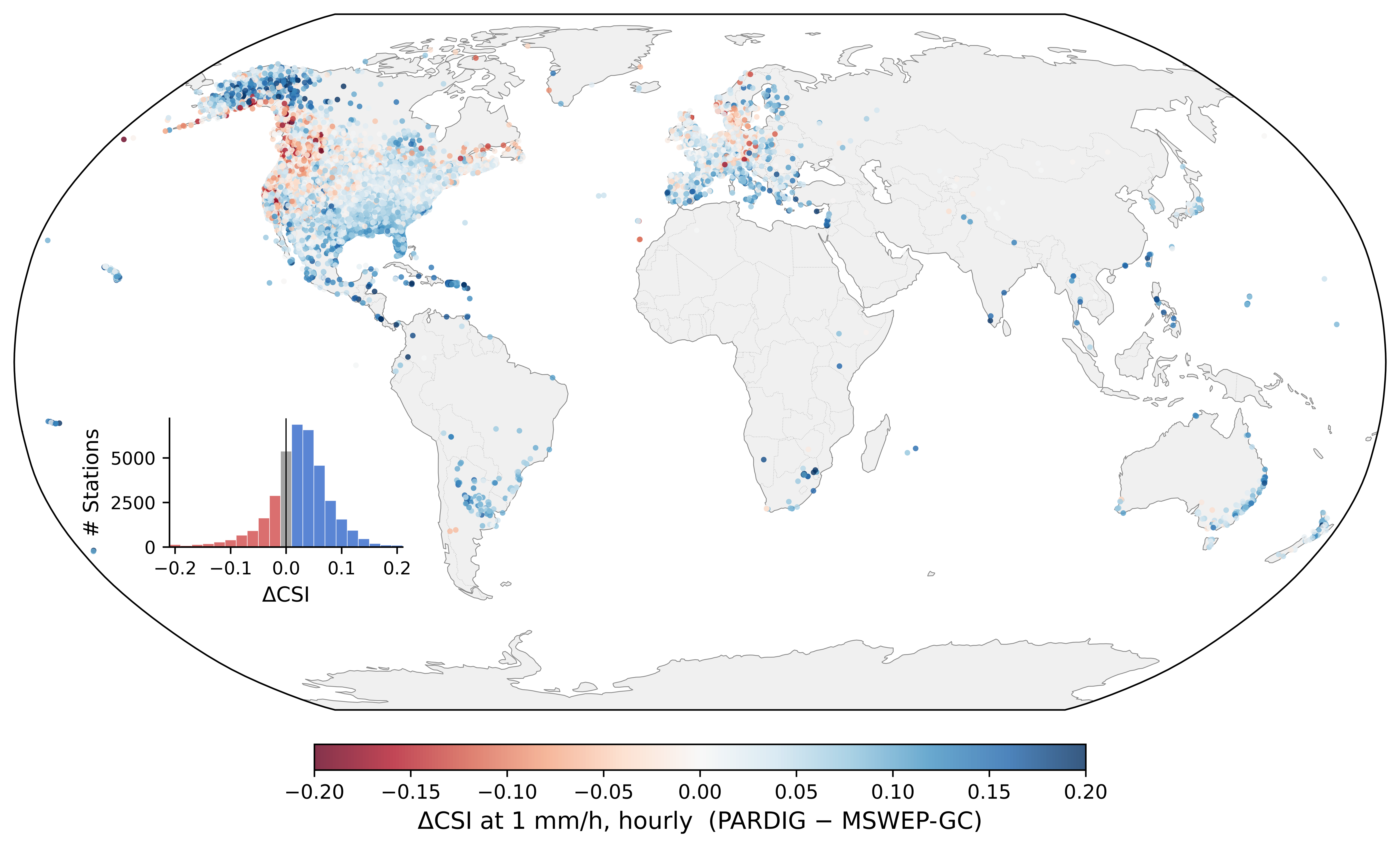}
    \caption{\textbf{Global station-level comparison between \pardignrt and MSWEP-GC.} Per-station CSI difference ($\Delta\text{CSI}$: \pardignrt minus MSWEP-GC) across the same $36{,}678$ quality-controlled Synoptic surface rain-gauge stations worldwide (2018--2023) as \Cref{fig:figure_3}, evaluated at the 1\,mm/h hourly accumulation threshold. Blue circles indicate stations where \pardignrt achieves higher CSI; red circles indicate stations where MSWEP-GC achieves higher CSI (color scale clipped to $[-0.20, +0.20]$). The inset histogram shows the distribution of per-station $\Delta\text{CSI}$ values across all $36{,}678$ stations (with the neutral center bin $|\Delta\text{CSI}| < 0.01$ shaded grey).}\label{fig:supp_station_map_mswep}
\end{figure}

\Cref{fig:supp_station_map_mswep} shows the corresponding per-station $\Delta\text{CSI}$ map between \pardignrt and MSWEP-GC at $1$\,mm/h across the same $36{,}678$ stations as \Cref{fig:figure_3}. Importantly, station-based verification is inherently biased against \pardig in this comparison: because MSWEP~V3 trains on and directly assimilates observations from a large fraction of the global gauge network, evaluating at gauge coordinates samples MSWEP-GC precisely where its local gauge corrections peak, whereas gauge-free \pardig treats station cells no differently from surrounding ungauged grid cells. Even on this unfavorable footing, \pardignrt outperforms MSWEP-GC at $65.4\%$ of stations worldwide ($14.7\%$ tied and $19.9\%$ favoring MSWEP-GC; mean CSI $0.36$ vs.\ $0.33$). Consistent with \Cref{fig:figure_3}, the advantage is largest across the $2{,}841$ stations in the tropical and subtropical belt ($\ang{30}$S--$\ang{30}$N), where \pardignrt outperforms MSWEP-GC at $92.5\%$ of stations (vs.\ $2.1\%$ for MSWEP-GC and $5.5\%$ tied) with a mean gain of $+0.08$ CSI points ($0.36$ vs.\ $0.27$, $+30\%$ relative), compared with $63.2\%$ of stations and a mean gain of $+0.02$ CSI points ($0.36$ vs.\ $0.34$, $+6\%$ relative) poleward of $\ang{30}$.

\begin{figure}[htbp]
    \centering
    \includegraphics[width=\linewidth]{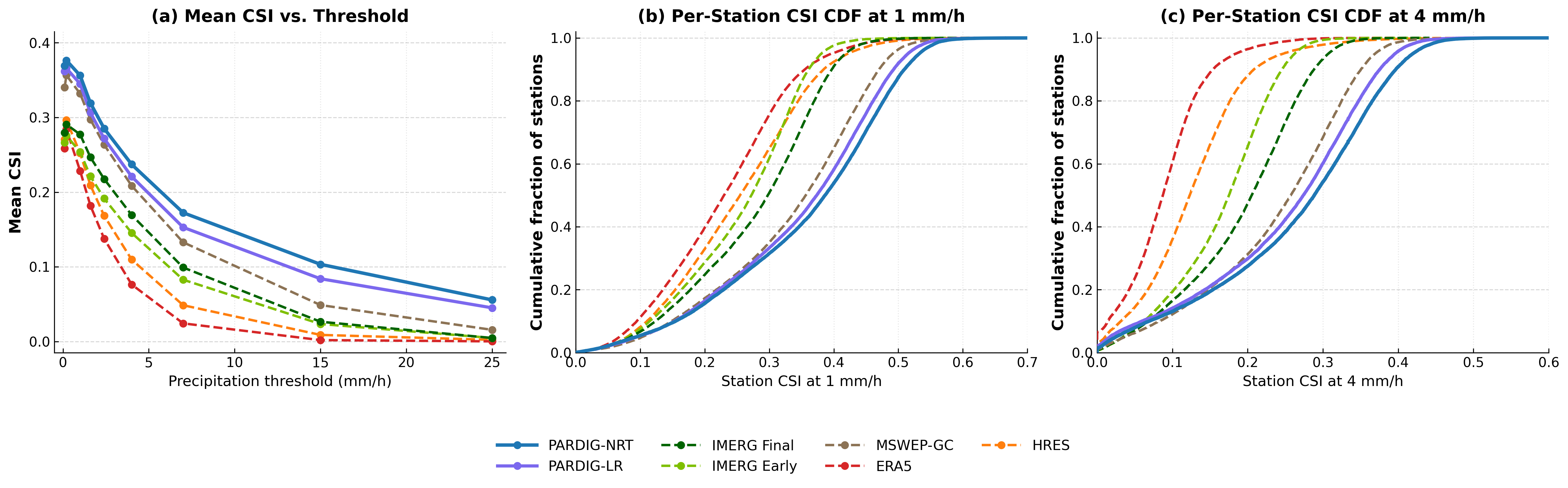}
    \caption{\textbf{Global Synoptic station-level CSI across hourly precipitation thresholds (2018--2023)} for \pardignrt, \pardiglr, IMERG Final, IMERG Early, MSWEP-GC, ERA5, and HRES.
    (a) Station-averaged CSI vs.\ hourly precipitation threshold ($0.2$--$25$\,mm/h) across $36{,}678$ quality-controlled stations (\Cref{sec:ground_truth}; numerical values in \Cref{tab:csi_stations_1h}).
    (b) Per-station CSI cumulative distribution function (CDF) at $1$\,mm/h ($36{,}678$ stations).
    (c) Per-station CSI CDF at $4$\,mm/h ($26{,}957$ stations meeting the same quality-control criteria at the $4$\,mm/h threshold).}
    \label{fig:supp_station_csi}
\end{figure}

Across the full hourly threshold range ($0.2$--$25$\,mm/h), \Cref{fig:supp_station_csi} and \Cref{tab:csi_stations_1h} compare station-averaged CSI (\Cref{fig:supp_station_csi}a) and per-station CSI cumulative distribution functions at $1$\,mm/h and $4$\,mm/h (\Cref{fig:supp_station_csi}b--c) across all seven products. Both \pardignrt and \pardiglr outperform every external baseline at every threshold, from moderate rain (mean CSI of $0.24$ and $0.22$ at $4$\,mm/h vs.\ $0.21$ for MSWEP-GC and $0.17$ for IMERG Final) out to heavy rates ($0.06$ and $0.05$ at $25$\,mm/h vs.\ $\le 0.02$ for all baselines), even though MSWEP-GC directly uses gauge observations overlapping this station network—giving it an inherent advantage over the other baselines. In the per-station CDFs (\Cref{fig:supp_station_csi}b--c), the \pardignrt and \pardiglr curves are shifted toward higher CSI across nearly the entire station distribution, showing that the higher mean skill reflects broad gains across stations rather than isolated outliers.

\begin{table}[htbp]
\centering
\small
\setlength{\tabcolsep}{4.5pt}
\caption{\textbf{Critical Success Index (CSI) across hourly precipitation thresholds at global Synoptic surface stations}. Evaluated across $36{,}678$ quality-controlled stations worldwide (2018--2023, 1-hour accumulation; \Cref{sec:ground_truth}). Best values per column in bold.}
\label{tab:csi_stations_1h}
\begin{tabular}{@{}lcccccc@{}}
\toprule
Product & 0.2\,mm/h & 1\,mm/h & 4\,mm/h & 7\,mm/h & 15\,mm/h & 25\,mm/h \\
\midrule
\pardignrt & \textbf{0.38} & \textbf{0.36} & \textbf{0.24} & \textbf{0.17} & \textbf{0.10} & \textbf{0.06} \\
\pardiglr & 0.37 & 0.35 & 0.22 & 0.15 & 0.08 & 0.05 \\
IMERG Final & 0.29 & 0.28 & 0.17 & 0.10 & 0.03 & 0.00 \\
IMERG Early & 0.28 & 0.25 & 0.15 & 0.08 & 0.02 & 0.00 \\
MSWEP-GC & 0.36 & 0.33 & 0.21 & 0.13 & 0.05 & 0.02 \\
ERA5 & 0.28 & 0.23 & 0.08 & 0.02 & 0.00 & 0.00 \\
HRES & 0.30 & 0.25 & 0.11 & 0.05 & 0.01 & 0.00 \\
\bottomrule
\end{tabular}
\end{table}

\FloatBarrier

\section{Spatial Structure and Multi-Scale Variability}
\label{sec:supp_spatial_structure}

While standard categorical verification metrics (\Cref{sec:supp_metrics}) evaluate detection skill at individual grid points and across intensity thresholds, they do not directly evaluate whether precipitation fields exhibit physically realistic spatial coherence, multi-scale variance, and spatial gradients. Capturing fine-scale structure is critical for hydrological applications, where localized runoff and streamflow responses depend strongly on the spatial organization and peak sharpness of precipitation fields \citep{cristiano2017spatial}.

As an initial exploration of the spatial characteristics of \pardig (which is also partially complemented by the event-level case studies in \Cref{fig:figure_4} and \Cref{fig:supp_case_study}), we present two complementary analyses:
\begin{enumerate}[leftmargin=*]
    \item \textbf{Global Spherical Harmonic Spectral Analysis} (\Cref{sec:supp_spectral}): Evaluates the variance power distribution across spatial scales globally, from planetary wavenumbers down to the grid-scale Nyquist limit ($\lambda \approx 22$\,km), across all 2023 hourly fields.
    \item \textbf{Spatial Variogram Analysis over CONUS} (\Cref{sec:supp_variogram}): Evaluates the spatial correlation structure and decorrelation scale directly against independent, high-resolution radar ground truth (MRMS) on a common grid, verifying whether fine-scale spatial variations reflect genuine physical structure.
\end{enumerate}

\subsection{Global Spectral Analysis}
\label{sec:supp_spectral}

To evaluate the effective spatial resolution and multi-scale variance distribution of precipitation products, we compute the global spherical harmonic power spectrum (SHPS) across all hourly timesteps for 2023 on the common \tenthdegree global grid. Missing values at polar latitudes (where some satellite products lack coverage) are filled with zeros prior to the spherical harmonic transform; this has only a small effect on the spectrum given the low precipitation rates and small spherical area of polar regions. Using standard spherical harmonic decomposition on the sphere~\citep[as used in, e.g.,][]{price2025probabilistic}, any spherical scalar precipitation field $P(\theta, \phi)$ is expanded into real orthonormal spherical harmonics:
\begin{equation}
    P(\theta, \phi) = \sum_{l=0}^{l_{\max}} \sum_{m=-l}^{l} a_{l,m} Y_{l,m}(\theta, \phi),
\end{equation}
where $l$ is the total spherical harmonic degree (wavenumber), $m$ is the zonal wavenumber, and $l_{\max} = 1799$ corresponds to the Nyquist truncation limit on the \tenthdegree grid ($\Delta x \approx 11.1$\,km at the equator, $\lambda_{\text{Nyquist}} \approx 22.2$\,km). The total power $E(l)$ at degree $l$ is given by:
\begin{equation}
    E(l) = \frac{1}{4\pi} \sum_{m=-l}^{l} |a_{l,m}|^2,
\end{equation}
where the approximate physical spatial wavelength corresponds to $\lambda \approx 2\pi R_{\text{Earth}} / l \approx 40{,}000\,\text{km} / l$.

\begin{figure}[!htbp]
    \centering
    \includegraphics[width=0.88\linewidth]{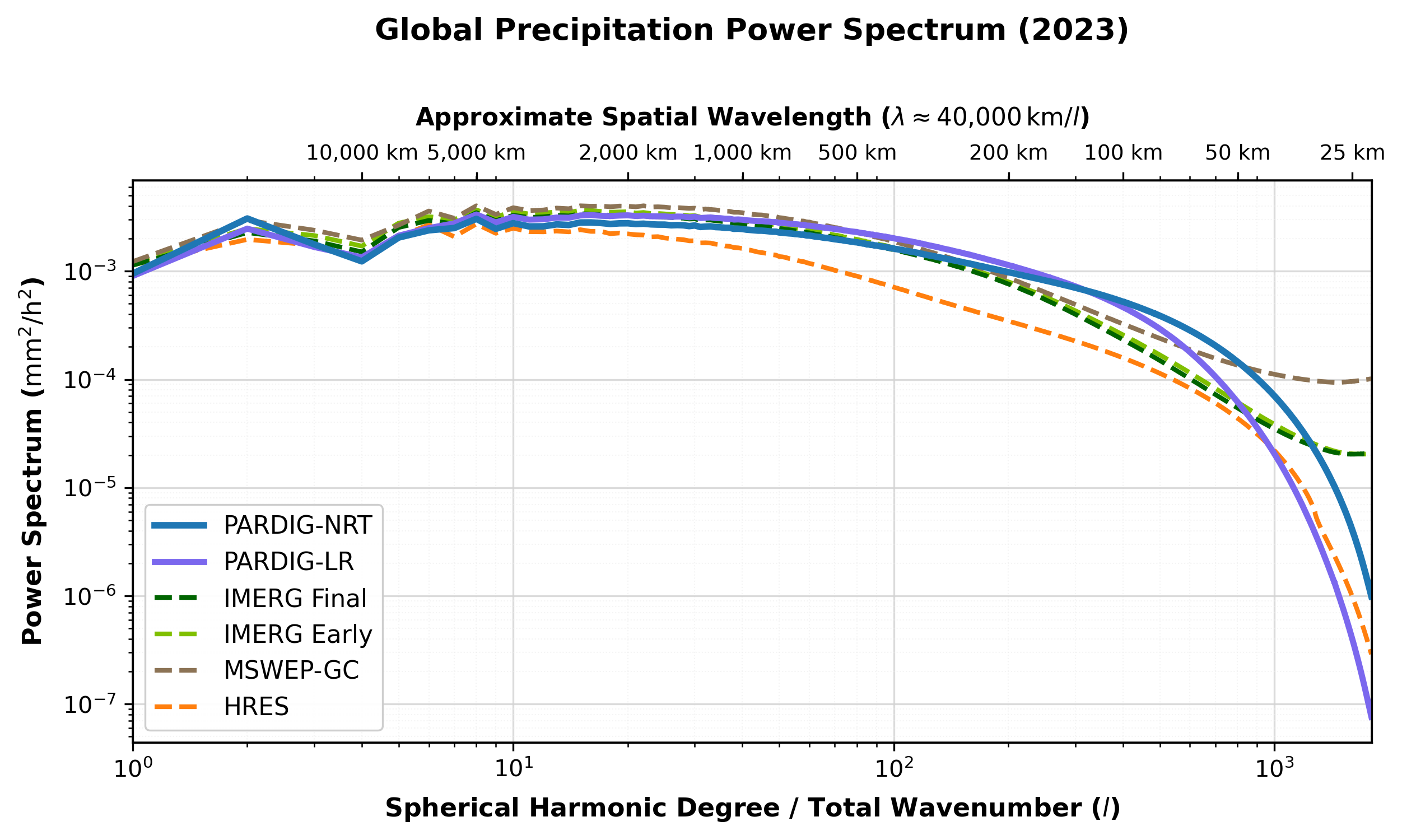}
    \caption{\textbf{Global precipitation spherical harmonic power spectrum for 2023.} Spherical harmonic total power $E(l)$ in $\text{mm}^2/\text{h}^2$ as a function of spherical harmonic degree $l$ (bottom axis) and approximate physical spatial wavelength $\lambda \approx 40{,}000\,\text{km}/l$ (top axis) on the global \tenthdegree grid for \pardignrt, \pardiglr, IMERG Final, IMERG Early, MSWEP-GC, and ECMWF HRES (ERA5 is omitted because its native \quarterdegree grid truncates at $l = 719$).}
    \label{fig:supp_spectral_analysis}
\end{figure}

\Cref{fig:supp_spectral_analysis} displays the time-averaged power spectrum $E(l)$ from planetary wavelengths ($\lambda \approx 40{,}000$\,km, $l=1$) down to the Nyquist limit ($\lambda \approx 22$\,km, $l=1799$).

At synoptic scales (1000--5000\,km), \pardig, IMERG, and MSWEP-GC exhibit comparable spectral power while HRES is consistently lower, whereas at mesoscale wavelengths (50--200\,km), \pardignrt retains substantially more power than all external baselines (roughly $1.6\text{--}3\times$ higher at 100\,km). \pardiglr tracks or slightly exceeds \pardignrt down to ${\sim}150\text{--}200$\,km before rolling off more steeply below 100\,km. Toward the shortest wavelengths (${<}\,50$\,km), MSWEP-GC and IMERG flatten into high-wavenumber spectral shelves—possibly mirroring the pixel-scale speckling seen in \Cref{fig:supp_case_study}—whereas \pardignrt, \pardiglr, and HRES decay monotonically to the Nyquist limit. Because power spectra alone cannot distinguish physical convective variance from model-generated texture, \Cref{sec:supp_variogram} evaluates spatial correlation directly against ground radar.

\subsection{Spatial Variogram Analysis}
\label{sec:supp_variogram}

Because spherical harmonic decomposition requires globally continuous fields on the sphere, the spectral analysis in \Cref{sec:supp_spectral} cannot directly incorporate regional ground radar; we therefore pair it with a spatial semivariogram analysis over CONUS, providing two complementary perspectives (both methodologically and in the spatial scales they cover) that allow direct comparison of fine-scale correlation structure against MRMS ground truth. To compute the spatial semivariogram $\gamma(h) = \tfrac{1}{2}\,\mathbb{E}[(P(\mathbf{x}) - P(\mathbf{x}+\mathbf{h}))^2]$ over the contiguous United States (CONUS), we use high-density, gauge-adjusted MRMS radar data as a benchmark on the common \tenthdegree grid. We evaluate $\gamma(h)$ across 200 randomly sampled hourly timestamps uniformly distributed across 2023, averaging zonal and meridional semivariograms across lags of 1--50 grid points (${\sim}10$\,km per $\ang{0.1}$ step, up to $500$\,km). To ensure a fair comparison, all products are evaluated over the identical spatial domain defined by the MRMS spatial quality mask, which retains 94{,}216 valid land grid points (62.8\% of the $250 \times 600$ CONUS bounding box) by excluding ocean, coastal, and beam-blocked regions. We normalize each time-averaged semivariogram by the product's mean within-frame spatial variance $\mathrm{Var}(P)$ (computed across valid pixels within each hourly frame first and then averaged over the 200 frames), $\gamma(h)/\mathrm{Var}(P)$, to isolate spatial correlation structure from differences in total intensity. The decorrelation length $L_d$ is defined as the distance at which $\gamma(h)/\mathrm{Var}(P) = 0.9$. Because the native MRMS product is available at $0.01^\circ$ (${\sim}1$\,km), we coarsen it to \tenthdegree using two methods representing areal vs.\ point-scale references on the \tenthdegree grid: block-averaging ($10 \times 10$ pixel means, representing grid-cell averages) and subsampling (retaining every 10th pixel, preserving point-measurement variance).

\begin{figure}[!htbp]
    \centering
    \includegraphics[width=0.88\linewidth]{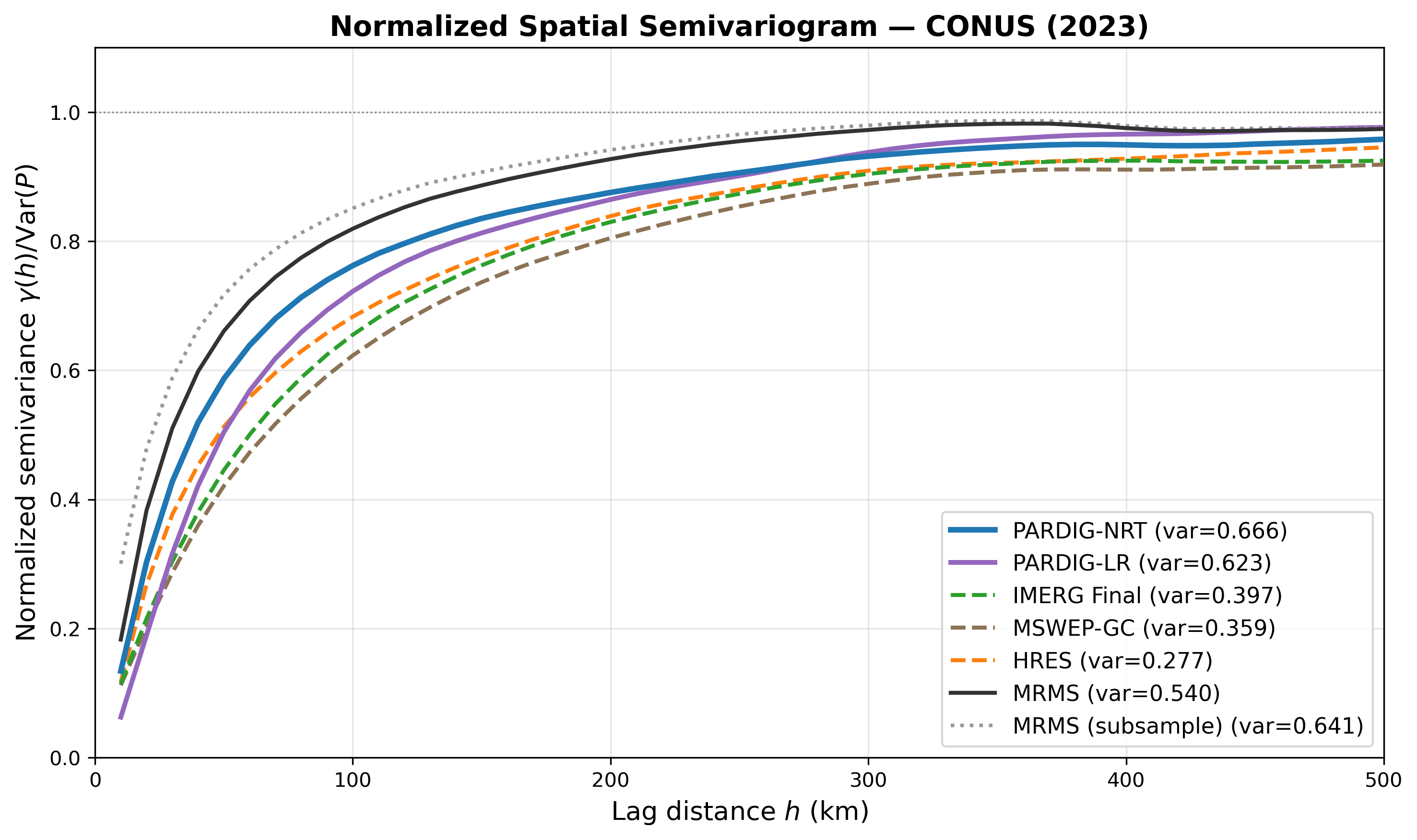}
    \caption{\textbf{Normalized spatial semivariogram over CONUS for 2023.} Normalized semivariance $\gamma(h)/\mathrm{Var}(P)$ as a function of spatial lag distance $h$ (in km) evaluated across 200 random hourly timestamps in 2023 on the common \tenthdegree grid. All products are evaluated over the identical spatial domain defined by the MRMS radar quality mask (94{,}216 valid land pixels, 62.8\% coverage). Mean within-frame spatial variance $\mathrm{Var}(P)$ ($\text{mm}^2/\text{h}^2$) for each product is indicated in the legend. Curves that rise more steeply indicate shorter spatial correlation lengths (sharper spatial features), while flatter curves indicate greater spatial smoothing.}
    \label{fig:supp_variogram_analysis}
\end{figure}

As shown in \Cref{fig:supp_variogram_analysis}, the normalized semivariogram of \pardignrt tracks block-averaged MRMS more closely than any other product across mesoscale distances. Its decorrelation length ($L_d = 240$\,km) is closer to the MRMS reference range (150--170\,km) than that of any external baseline ($290\text{--}300$\,km for HRES and IMERG Final, and $330$\,km for MSWEP-GC, which exhibits the slowest spatial decorrelation over CONUS despite its elevated sub-mesoscale spectral shelf), though it remains smoother than radar at the shortest lags. \pardiglr is smoother at fine scales, overtaking the external baselines beyond ${\sim}55$\,km and converging with \pardignrt around 200\,km ($L_d = 250$\,km).

A common concern is that deep learning architectures inherently produce over-smoothed fields. In atmospheric \emph{forecasting}, this typically arises when deterministic models trained with regression objectives (e.g., MSE) are forced toward the conditional mean of divergent trajectories at longer lead times. \pardig operates in a fundamentally different regime: as an \emph{estimation} model conditioned on concurrent observations, it does not face this trajectory-averaging pressure. More broadly, recent work has demonstrated that deep learning models need not produce blurred outputs when appropriate loss functions and architectures are employed \citep{ravuri2021skilful, kochkov2024neuralgcm, price2025probabilistic}. The spectral and variogram analyses above, partially complemented by the extreme-event case studies (\Cref{fig:figure_4} and \Cref{fig:supp_case_study}), provide initial evidence that \pardig preserves spatial gradients and fine-scale variance closer to radar ground truth than existing gridded precipitation products, while a more comprehensive characterization of spatial structure remains an important direction for future work.

\FloatBarrier
\section{Sub-Hourly Temporal Dynamics and Offset Consistency}
\label{sec:supp_subhourly}

While our primary evaluations focus on hourly accumulations to align with standard observational benchmarks and gridded products, \pardig natively estimates instantaneous precipitation rates at 15-minute intervals (HH:00, HH:15, HH:30, and HH:45 of each hour). A natural question for any high-cadence retrieval system is whether the sub-hourly output captures genuine high-frequency meteorological variability or whether successive frames are largely copies of one another. Here, we evaluate the temporal dynamics of \pardignrt against high-resolution ground radar (MRMS Instantaneous Precipitation Rate) and IMERG Final (30-minute) over the contiguous United States (CONUS), as the hourly baselines (MSWEP-GC, ERA5, and HRES) do not resolve sub-hourly intervals.

\subsection{Temporal Lag Autocorrelation}
\label{sec:supp_subhourly_autocorr}

\begin{figure}[htbp]
    \centering
    \includegraphics[width=0.88\linewidth]{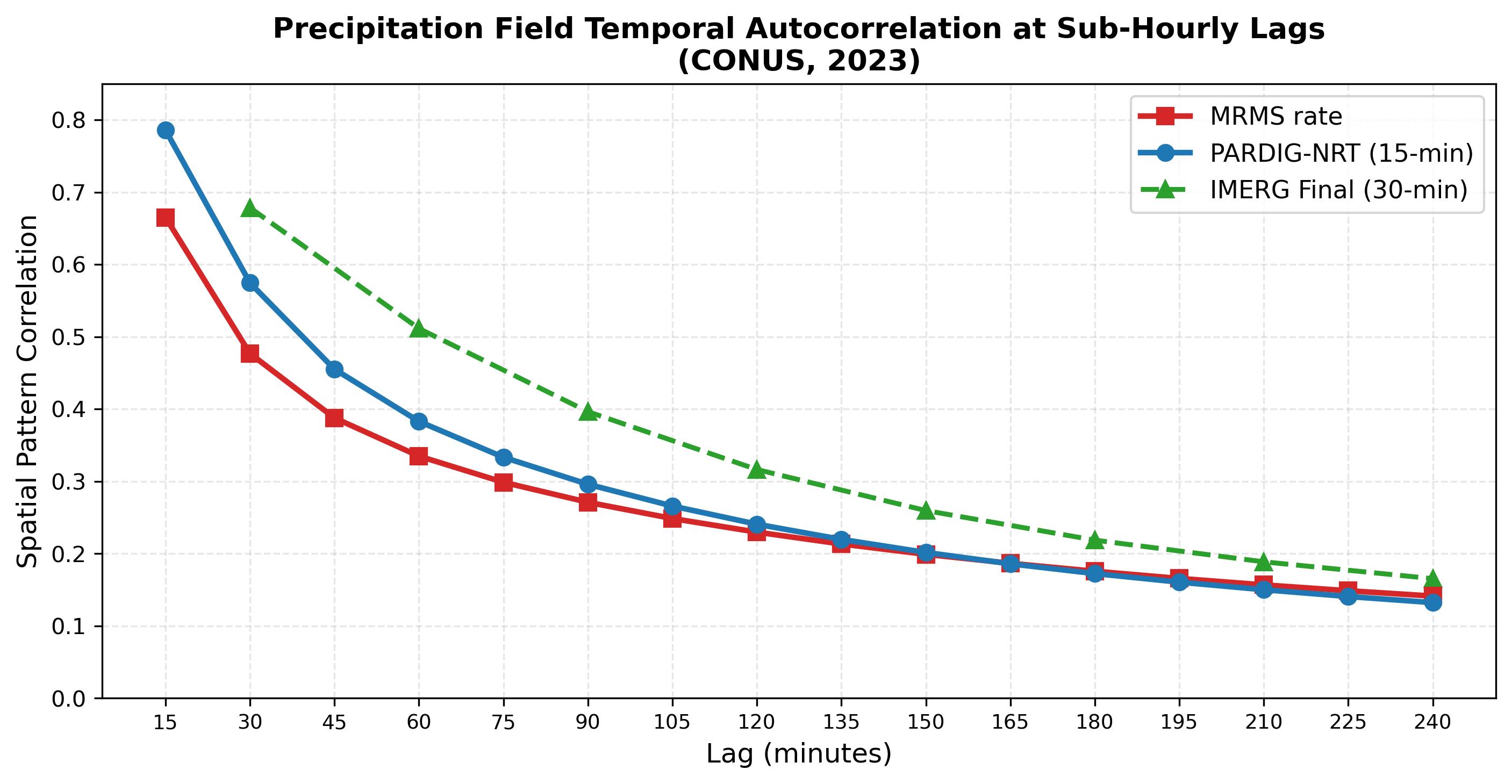}
    \caption{\textbf{Precipitation field temporal autocorrelation at sub-hourly lags} (CONUS, 2023). Spatial pattern correlation (Pearson $r$) on the common \tenthdegree grid across valid radar pixels at lags up to 4\,hours for \pardignrt, $10 \times 10$ block-averaged MRMS Instantaneous Precipitation Rate, and IMERG Final.}
    \label{fig:supp_subhourly_autocorr}
\end{figure}

To quantify temporal persistence and the rate of spatial pattern evolution, we compute the spatial pattern correlation (Pearson $r$) between precipitation fields at successive time steps on the common \tenthdegree grid as a function of temporal lag, up to four hours ($15$ to $240$\,minutes). \Cref{fig:supp_subhourly_autocorr} compares the resulting lag autocorrelation curves for \pardignrt, MRMS Rate ($10 \times 10$ block-averaged from $\ang{0.01}$ to \tenthdegree), and IMERG Final over CONUS across 2023.

The autocorrelation decay reveals two key characteristics. First, \pardignrt decorrelates rapidly from step to step (dropping from $r = 0.79$ at 15\,min to $0.38$ at 60\,min) and closely tracks the decay curve of ground radar (MRMS Rate: $0.67$ at 15\,min and $0.33$ at 60\,min, converging after ${\sim}2$\,hours), indicating genuine sub-hourly pattern evolution rather than near-duplicate frames. Second, IMERG Final exhibits substantially stronger temporal persistence across all lags ($r = 0.68$ at 30\,min and $0.51$ at 60\,min, vs.\ $0.57$ and $0.38$ for \pardignrt). While part of this difference reflects IMERG reporting 30-minute window averages rather than instantaneous snapshots, the slower decay even at 1--2\,hour lags likely also reflects temporal smoothing from Lagrangian morphing between passive microwave overpasses.

\subsection{Consistency Across Sub-Hourly Window Offsets}
\label{sec:supp_subhourly_offsets}

As described in Section~\ref{sec:methods}, \pardig operates on hourly inference windows, producing four instantaneous frames per window at nominal offsets of $00$:$45$ ($-15$\,min), $00$:$00$ (analysis time), $00$:$15$ ($+15$\,min), and $00$:$30$ ($+30$\,min). Because the nominal atmospheric state inputs are centered at $00$:$00$, it is important to verify that retrieval skill remains stable across all intra-window time steps as a sanity check on model consistency.

\Cref{tab:supp_subhourly_csi} presents Critical Success Index (CSI) across six precipitation thresholds evaluated against MRMS Instantaneous Precipitation Rate for the four intra-hour offsets over CONUS (2023).

\begin{table}[htbp]
\centering
\small
\setlength{\tabcolsep}{5pt}
\caption{\textbf{\pardignrt CSI by sub-hourly offset across precipitation thresholds}. Evaluated against MRMS Instantaneous Precipitation Rate (CONUS, 2023) for each sub-hourly offset within the inference window. Values are reported to three decimal places to examine intra-hour stability.}
\label{tab:supp_subhourly_csi}
\begin{tabular}{@{}lcccc@{}}
\toprule
Threshold & 00:45 ($-15$\,min) & 00:00 & 00:15 ($+15$\,min) & 00:30 ($+30$\,min) \\
\midrule
0.2\,mm/h & 0.484 & 0.487 & 0.486 & 0.483 \\
1.0\,mm/h & 0.426 & 0.428 & 0.428 & 0.424 \\
4.0\,mm/h & 0.277 & 0.278 & 0.278 & 0.275 \\
7.0\,mm/h & 0.215 & 0.215 & 0.215 & 0.213 \\
15.0\,mm/h & 0.194 & 0.194 & 0.192 & 0.190 \\
25.0\,mm/h & 0.146 & 0.146 & 0.142 & 0.140 \\
\bottomrule
\end{tabular}
\end{table}

Across all thresholds, the maximum CSI variation between offsets is $\le 0.006$ (e.g., $0.424$--$0.428$ at $1$\,mm/h). The slight dip at the $00$:$30$ offset (furthest from the centers of the input and target windows) is consistent with expectations but negligible in practice. This stability confirms that \pardignrt delivers reliable, consistent precipitation estimates across all sub-hourly output steps, without phase-dependent artifacts or substantial skill degradation away from the nominal analysis hour.

\FloatBarrier

\section{Accumulation Timescale Evaluation}
\label{sec:supp_accumulation}
\suppressfloats[t]

While our primary evaluations focus on hourly resolution (\Cref{fig:figure_2} and \Cref{sec:supp_metrics}), many operational hydrology, flood prediction, and agricultural applications require sub-daily or daily precipitation accumulations. Evaluating longer accumulation windows also tests whether temporal integration, which smooths out short-term phase and timing errors in numerical weather prediction and satellite products, closes the skill gap with \pardig.

\begin{figure}[!htb]
    \centering
    \includegraphics[width=\linewidth]{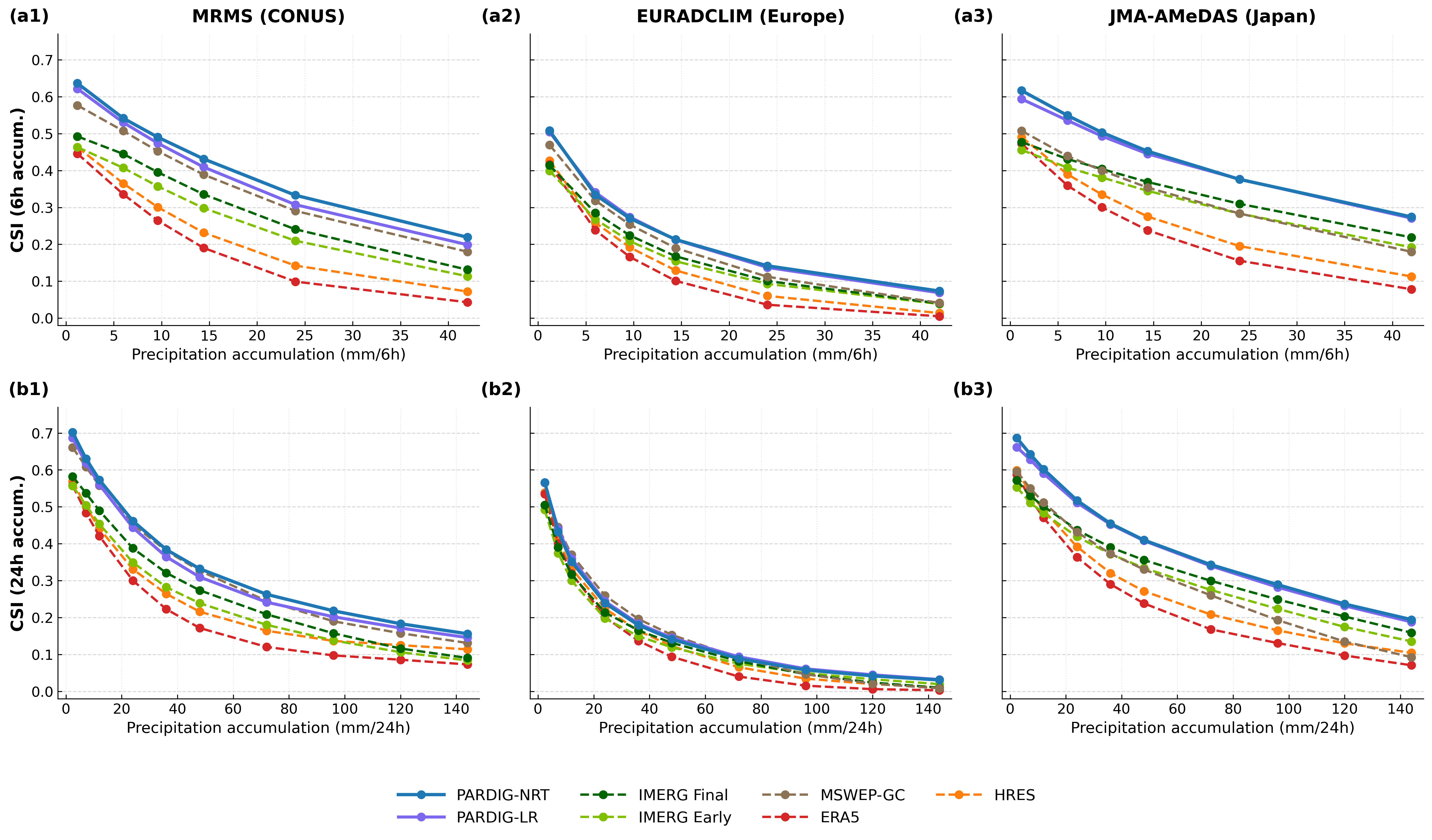}
    \caption{\textbf{CSI vs precipitation accumulation threshold for 6-hour and 24-hour windows across regional radar networks (2016--2023)} for MRMS (CONUS, column 1), EURADCLIM (Europe, column 2), and JMA Radar-AMeDAS (Japan, column 3).
    (a1--a3) 6-hour accumulation ($1.2$--$42$\,mm/6h).
    (b1--b3) 24-hour accumulation ($2.4$--$144$\,mm/24h).}
    \label{fig:supp_accum_csi}
\end{figure}

\Cref{fig:supp_accum_csi} presents the multi-year average Critical Success Index (CSI, 2016--2023) as a function of precipitation accumulation for 6-hour ($1.2$--$42$\,mm/6h) and 24-hour ($2.4$--$144$\,mm/24h) windows across the three regional gauge-radar networks: MRMS (CONUS), EURADCLIM (Europe), and JMA Radar-AMeDAS (Japan).

At the 6-hour timescale, \pardignrt and \pardiglr outperform all external baselines, including MSWEP-GC, at every threshold across all three radar networks (\Cref{fig:supp_accum_csi}a1--a3), with the two variants achieving comparable CSI values. At 24-hour accumulations (\Cref{fig:supp_accum_csi}b1--b3), \pardignrt leads every baseline across all thresholds over CONUS and Japan (with \pardiglr trailing only MSWEP-GC slightly around $24\text{--}48$\,mm/24h over CONUS), whereas over Europe (EURADCLIM), daily gauge-corrected MSWEP-GC matches \pardig at $2.4$\,mm/24h and slightly exceeds both \pardig variants at moderate daily totals ($12\text{--}48$\,mm/24h; e.g., $0.26$ vs.\ $0.24$ CSI at $24$\,mm/24h) before falling below both \pardignrt and \pardiglr at heavy daily extremes ($96\text{--}144$\,mm/24h).

\FloatBarrier
\section{Temporal Generalization}
\label{sec:supp_temporal}

All \pardig models are trained on GPM CORRA observations through December 2023. To evaluate temporal stability and verify whether the model suffers from overfitting to training periods or specific sensor characteristics, we analyze multi-year performance along two complementary axes:
\begin{enumerate}[label=(\roman*),leftmargin=*]
    \item \textbf{Independent Ground Truth (Radar):} Year-by-year evaluation against gauge-adjusted radar networks across three continents: MRMS (CONUS, 2016--2025, including the held-out 2024--2025 years), EURADCLIM (Europe, 2016--2023), and JMA Radar-AMeDAS (Japan, 2016--2025, including the held-out 2024--2025 years).
    \item \textbf{Training Sensor Evaluation (GPM):} Year-by-year evaluation against GPM CORRA satellite observations across 2016--2025 (with 2016--2023 representing in-training observations and 2024--2025 as the held-out test years).
\end{enumerate}

\subsection{Multi-Year Radar Stability}
\label{sec:supp_yearly_radar}

\begin{figure}[htbp]
    \centering
    \includegraphics[width=\linewidth]{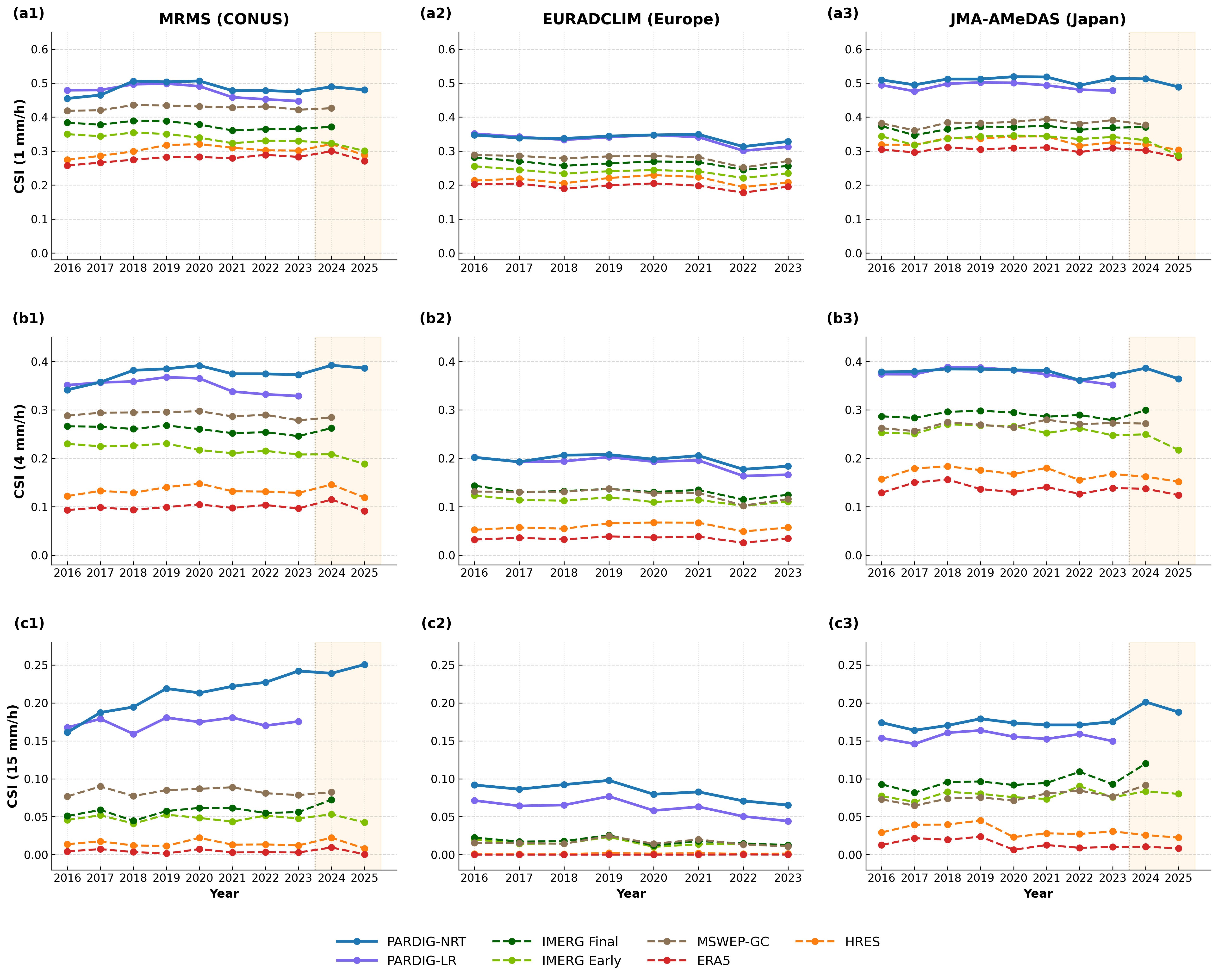}
    \caption{\textbf{Yearly CSI stability across radar networks} for MRMS (CONUS, 2016--2025), EURADCLIM (Europe, 2016--2023), and JMA Radar-AMeDAS (Japan, 2016--2025) at 1\,mm/h (top row), 4\,mm/h (middle row), and 15\,mm/h (bottom row). The \pardignrt held-out years 2024--2025 are shaded on the MRMS and JMA Radar-AMeDAS panels (\pardiglr and EURADCLIM end in 2023, and IMERG Final and MSWEP-GC end in 2024 as full-year 2025 records were not available). Numerical values at 1\,mm/h are provided in \Cref{tab:yearly_csi_radar_all} (with paired significance tests in \Cref{tab:significance}).}
    \label{fig:supp_yearly_csi}
\end{figure}

\Cref{fig:supp_yearly_csi} displays CSI across individual years for three precipitation thresholds: light (1\,mm/h), moderate (4\,mm/h), and heavy (15\,mm/h). Numerical yearly CSI values at 1\,mm/h are provided in \Cref{tab:yearly_csi_radar_all}.

Across all three radar networks and all evaluated thresholds, \pardig performance is consistent over time, with \pardignrt and \pardiglr achieving the highest CSI across all evaluated years. Over Europe (EURADCLIM), CONUS (MRMS), and Japan (JMA Radar-AMeDAS), yearly CSI curves show little interannual variation across the multi-year period. In the held-out 2024--2025 years, \pardignrt performance remains in line with in-training years across both CONUS ($\text{CSI} = 0.49$ and $0.48$ at 1\,mm/h in 2024 and 2025, compared to $0.48$ in 2023; and ${\sim}0.24\text{--}0.25$ across 2023--2025 at 15\,mm/h) and Japan ($\text{CSI} = 0.51$ and $0.49$ at 1\,mm/h in 2024 and 2025, compared to $0.49\text{--}0.52$ during 2016--2023). Notably, at 15\,mm/h over CONUS (\Cref{fig:supp_yearly_csi}(c1)), \pardignrt CSI rises steadily from ${\sim}0.16$ in 2016 to ${\sim}0.24\text{--}0.25$ in the held-out 2024--2025 years (whereas \pardiglr remains flat), a trend we hypothesize may be linked to progressive improvements in GOES geostationary observations over the Americas.

While categorical verification at high hourly thresholds ($\ge 10\text{--}15$\,mm/h) is often viewed as challenging due to event sparsity and spatial displacement penalties, \Cref{fig:supp_yearly_csi} indicates that annual radar-based evaluations can remain relatively stable even at $15$\,mm/h. Across the evaluated networks, interannual variations are modest, and the relative performance ordering among products is largely preserved from year to year. When aggregated over a full annual cycle of radar observations, the sample size of heavy precipitation pixels appears sufficient to support informative comparisons between products.

\begin{table}[htbp]
\centering
\small
\setlength{\tabcolsep}{2.5pt}
\caption{\textbf{Yearly CSI at 1\,mm/h across radar networks}. Years 2024--2025 are held-out test years for \pardignrt on MRMS (CONUS) and JMA Radar-AMeDAS (Japan); dashes indicate unavailable full-year records (\pardiglr and EURADCLIM end in 2023, and IMERG Final and MSWEP-GC were not available for all of 2025). Values are reported to three decimal places to examine interannual stability. Best product per year in bold.}
\label{tab:yearly_csi_radar_all}
\begin{tabular}{@{}lcccccccccc@{}}
\toprule
Product & 2016 & 2017 & 2018 & 2019 & 2020 & 2021 & 2022 & 2023 & \textbf{2024 (Held-out)} & \textbf{2025 (Held-out)} \\
\midrule
\multicolumn{11}{@{}l}{\textbf{(a) MRMS (CONUS)}} \\
\midrule
\pardignrt & 0.455 & 0.465 & \textbf{0.506} & \textbf{0.504} & \textbf{0.506} & \textbf{0.478} & \textbf{0.478} & \textbf{0.475} & \textbf{0.489} & \textbf{0.480} \\
\pardiglr & \textbf{0.479} & \textbf{0.480} & 0.497 & 0.499 & 0.491 & 0.458 & 0.452 & 0.447 & --- & --- \\
IMERG Final & 0.384 & 0.378 & 0.389 & 0.388 & 0.378 & 0.361 & 0.364 & 0.366 & 0.371 & --- \\
IMERG Early & 0.350 & 0.344 & 0.355 & 0.350 & 0.339 & 0.323 & 0.330 & 0.330 & 0.324 & 0.300 \\
MSWEP-GC & 0.419 & 0.420 & 0.436 & 0.434 & 0.431 & 0.428 & 0.431 & 0.422 & 0.426 & --- \\
ERA5 & 0.258 & 0.266 & 0.275 & 0.282 & 0.283 & 0.279 & 0.289 & 0.283 & 0.300 & 0.271 \\
HRES & 0.274 & 0.286 & 0.299 & 0.318 & 0.320 & 0.310 & 0.302 & 0.301 & 0.321 & 0.288 \\
\midrule
\multicolumn{11}{@{}l}{\textbf{(b) EURADCLIM (Europe)}} \\
\midrule
\pardignrt & 0.347 & 0.339 & \textbf{0.337} & \textbf{0.344} & \textbf{0.348} & \textbf{0.349} & \textbf{0.313} & \textbf{0.328} & --- & --- \\
\pardiglr & \textbf{0.351} & \textbf{0.342} & 0.333 & 0.341 & 0.347 & 0.341 & 0.301 & 0.313 & --- & --- \\
IMERG Final & 0.281 & 0.270 & 0.257 & 0.264 & 0.269 & 0.268 & 0.245 & 0.256 & --- & --- \\
IMERG Early & 0.255 & 0.245 & 0.234 & 0.241 & 0.244 & 0.240 & 0.221 & 0.235 & --- & --- \\
MSWEP-GC & 0.288 & 0.286 & 0.278 & 0.284 & 0.285 & 0.282 & 0.252 & 0.271 & --- & --- \\
ERA5 & 0.202 & 0.204 & 0.189 & 0.199 & 0.205 & 0.198 & 0.178 & 0.195 & --- & --- \\
HRES & 0.213 & 0.219 & 0.205 & 0.221 & 0.229 & 0.224 & 0.194 & 0.208 & --- & --- \\
\midrule
\multicolumn{11}{@{}l}{\textbf{(c) JMA Radar-AMeDAS (Japan)}} \\
\midrule
\pardignrt & \textbf{0.510} & \textbf{0.495} & \textbf{0.512} & \textbf{0.512} & \textbf{0.519} & \textbf{0.518} & \textbf{0.494} & \textbf{0.514} & \textbf{0.513} & \textbf{0.489} \\
\pardiglr & 0.494 & 0.476 & 0.498 & 0.502 & 0.501 & 0.494 & 0.481 & 0.478 & --- & --- \\
IMERG Final & 0.374 & 0.347 & 0.365 & 0.372 & 0.371 & 0.374 & 0.363 & 0.369 & 0.371 & --- \\
IMERG Early & 0.343 & 0.318 & 0.337 & 0.343 & 0.346 & 0.343 & 0.336 & 0.342 & 0.332 & 0.287 \\
MSWEP-GC & 0.382 & 0.361 & 0.384 & 0.382 & 0.386 & 0.394 & 0.380 & 0.391 & 0.377 & --- \\
ERA5 & 0.305 & 0.296 & 0.311 & 0.305 & 0.309 & 0.311 & 0.297 & 0.309 & 0.302 & 0.282 \\
HRES & 0.319 & 0.318 & 0.338 & 0.337 & 0.343 & 0.344 & 0.315 & 0.326 & 0.320 & 0.303 \\
\bottomrule
\end{tabular}
\end{table}

\FloatBarrier
\subsection{Statistical Significance of Inter-Product Differences}
\label{sec:supp_significance}

The year-by-year CSI results in \Cref{tab:yearly_csi_radar_all} enable formal testing of whether \pardig's improvement over baseline products is statistically significant across evaluation years. We apply paired $t$-tests, treating each year's CSI as a paired observation, across all shared evaluation years per radar network.

\Cref{tab:significance} reports, for each \pardig--baseline pair and radar network, the mean CSI difference ($\overline{\Delta\text{CSI}}$), its interannual standard deviation ($\sigma$), the paired $t$-statistic, and the $p$-value. All comparisons reach $p < 0.001$ over CONUS (9--10 shared years), Europe (8 years), and Japan (9--10 shared years), confirming statistically significant improvements over every external baseline.

\begin{table}[htbp]
\centering
\small
\setlength{\tabcolsep}{5pt}
\caption{\textbf{Statistical significance of CSI differences at 1\,mm/h} (paired $t$-tests across evaluation years). $n$: number of years; $\overline{\Delta}$: mean CSI difference (\pardignrt minus baseline); $\sigma$: interannual standard deviation of the difference.}
\label{tab:significance}
\begin{tabular}{@{}llccccc@{}}
\toprule
Region & vs Baseline & $n$ & $\overline{\Delta}$ & $\sigma$ & $t$ & $p$ \\
\midrule
\multirow{5}{*}{CONUS} & IMERG Final & 9 & +0.109 & 0.018 & 18.2 & $<$0.001 \\
 & IMERG Early & 10 & +0.149 & 0.022 & 21.5 & $<$0.001 \\
 & MSWEP-GC & 9 & +0.056 & 0.013 & 12.6 & $<$0.001 \\
 & ERA5 & 10 & +0.205 & 0.015 & 42.2 & $<$0.001 \\
 & HRES & 10 & +0.182 & 0.012 & 48.5 & $<$0.001 \\
\midrule
\multirow{5}{*}{Europe} & IMERG Final & 8 & +0.074 & 0.006 & 33.5 & $<$0.001 \\
 & IMERG Early & 8 & +0.099 & 0.007 & 42.3 & $<$0.001 \\
 & MSWEP-GC & 8 & +0.060 & 0.004 & 41.1 & $<$0.001 \\
 & ERA5 & 8 & +0.142 & 0.007 & 60.3 & $<$0.001 \\
 & HRES & 8 & +0.124 & 0.006 & 59.5 & $<$0.001 \\
\midrule
\multirow{5}{*}{Japan} & IMERG Final & 9 & +0.142 & 0.006 & 72.0 & $<$0.001 \\
 & IMERG Early & 10 & +0.175 & 0.011 & 48.1 & $<$0.001 \\
 & MSWEP-GC & 9 & +0.128 & 0.007 & 55.4 & $<$0.001 \\
 & ERA5 & 10 & +0.205 & 0.005 & 137.5 & $<$0.001 \\
 & HRES & 10 & +0.181 & 0.007 & 79.1 & $<$0.001 \\
\bottomrule
\end{tabular}
\end{table}

\FloatBarrier
\subsection{Training Set Memorization vs Out-of-Sample Generalization}

\begin{figure}[htbp]
    \centering
    \includegraphics[width=0.72\linewidth]{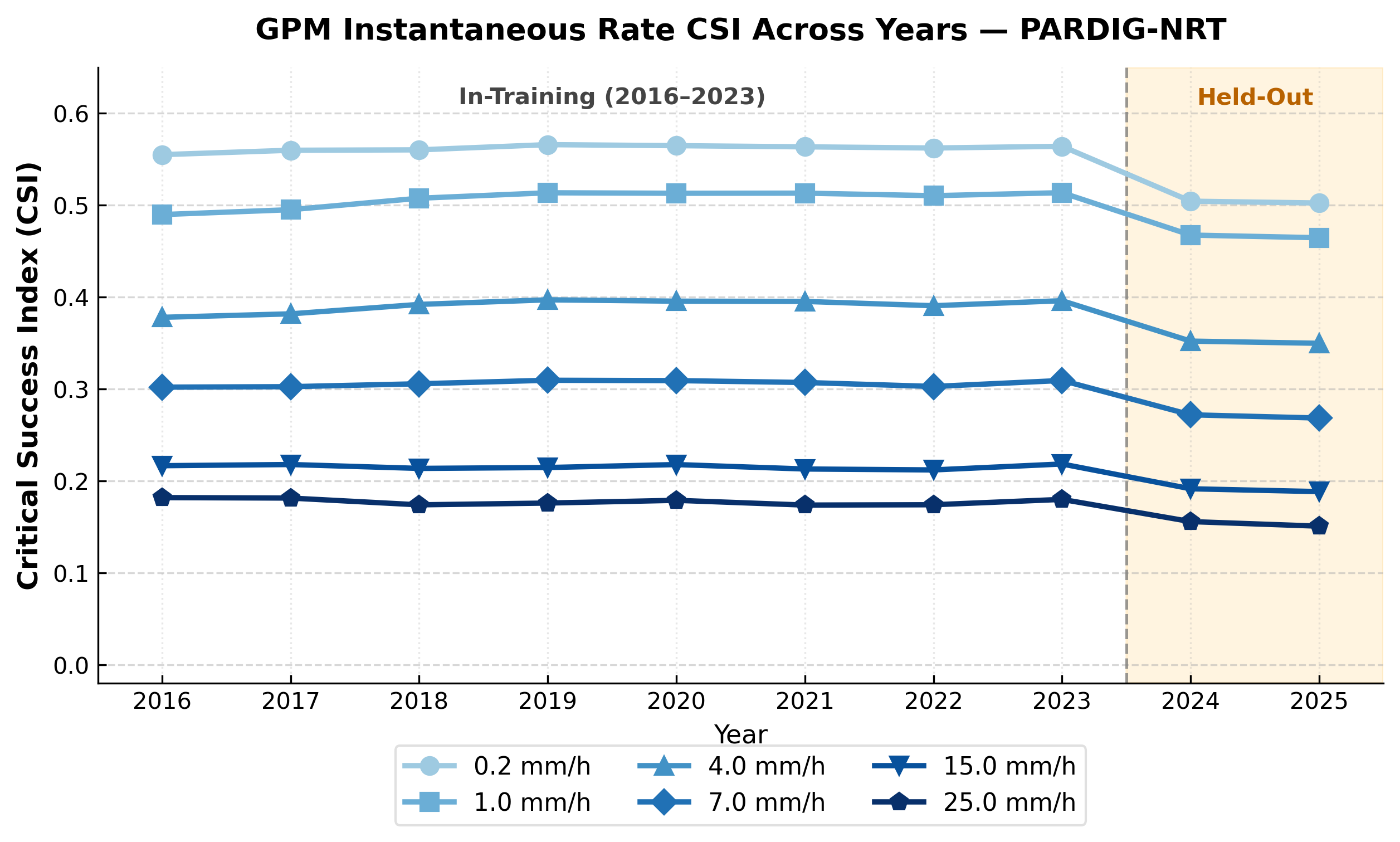}
    \caption{\textbf{GPM instantaneous rate yearly evaluation across thresholds} (0.2, 1, 4, 7, 15, 25\,mm/h) for \pardignrt spanning 2016--2025. The held-out test years (2024--2025) are highlighted with orange shading, illustrating the expected difference between in-sample evaluation on training swaths (2016--2023) and out-of-sample evaluation (2024--2025), which contrasts with the steady year-over-year performance observed on spatially independent radar networks (\Cref{fig:supp_yearly_csi}). Numerical values are provided in \Cref{tab:gpm_yearly_csi_hg}.}
    \label{fig:supp_gpm_yearly}
\end{figure}

To evaluate whether the model exhibits memorization of the training set, \Cref{fig:supp_gpm_yearly} examines \pardig instantaneous precipitation rate against GPM CORRA observations across individual years (2016--2025), and \Cref{tab:gpm_yearly_csi_hg} details numerical values across thresholds.

When evaluated against GPM directly (\Cref{fig:supp_gpm_yearly}), \pardig shows slightly higher performance on in-training years (2016--2023, where CSI averages ${\sim}0.51$ at 1\,mm/h), followed by a modest decrease in the held-out 2024--2025 years ($\text{CSI} = 0.47$ in both 2024 and 2025). This occurs because the 2016--2023 GPM evaluation is performed directly on the training swaths (the exact training examples), whereas 2024--2025 represent unseen satellite orbits. Notably, performance across the two held-out years is virtually identical across all thresholds, indicating no temporal drift beyond the training horizon.

In contrast, independent ground truth networks such as MRMS and JMA Radar-AMeDAS are evaluated over continuous geographic domains where GPM orbital swaths have negligible spatial-temporal overlap (${\sim}$245\,km swath width with revisit times of several days, intersecting only a few percent of the continuous domain). Consequently, radar evaluations even during 2016--2023 are essentially out-of-sample with respect to the training set.

Comparing \Cref{fig:supp_yearly_csi} (independent ground radar) and \Cref{fig:supp_gpm_yearly} (training target) demonstrates that while in-sample GPM scores reflect a degree of training example memorization, out-of-sample ground-radar performance remains steady in 2024--2025 relative to prior years. This strongly suggests that \pardig learns generalizable meteorological relationships rather than memorized sensor or time-period features, and that its operational skill does not meaningfully degrade between in-training years and held-out future periods.

\begin{table}[htbp]
\centering
\small
\setlength{\tabcolsep}{2.5pt}
\caption{\textbf{Yearly CSI across precipitation thresholds against GPM instant rate (\pardignrt)}. Years 2016--2023 represent in-training observations; 2024--2025 are held-out test years. Values are reported to three decimal places to demonstrate interannual stability.}
\label{tab:gpm_yearly_csi_hg}
\begin{tabular}{@{}lcccccccccc@{}}
\toprule
Threshold & 2016 & 2017 & 2018 & 2019 & 2020 & 2021 & 2022 & 2023 & \textbf{2024 (Held-out)} & \textbf{2025 (Held-out)} \\
\midrule
0.2\,mm/h & 0.555 & 0.560 & 0.560 & 0.566 & 0.565 & 0.563 & 0.562 & 0.564 & 0.504 & 0.502 \\
1\,mm/h & 0.490 & 0.495 & 0.508 & 0.514 & 0.513 & 0.513 & 0.510 & 0.514 & 0.468 & 0.465 \\
4\,mm/h & 0.378 & 0.382 & 0.392 & 0.397 & 0.396 & 0.395 & 0.391 & 0.396 & 0.352 & 0.350 \\
7\,mm/h & 0.302 & 0.303 & 0.306 & 0.310 & 0.309 & 0.307 & 0.303 & 0.309 & 0.272 & 0.269 \\
15\,mm/h & 0.217 & 0.218 & 0.214 & 0.215 & 0.218 & 0.213 & 0.212 & 0.219 & 0.192 & 0.189 \\
25\,mm/h & 0.182 & 0.182 & 0.174 & 0.176 & 0.179 & 0.174 & 0.174 & 0.180 & 0.156 & 0.151 \\
\bottomrule
\end{tabular}
\end{table}

\FloatBarrier

\section{GPM-Based Regional Evaluation}
\label{sec:supp_gpm_eval}

While the main text evaluates \pardig against independent ground truth, these sources are geographically restricted to well-instrumented land areas. To evaluate precipitation retrieval skill across broader portions of the globe, we evaluate \pardignrt against held-out GPM CORRA instantaneous precipitation rate observations (2024--2025). Because GPM CORRA observations provide quasi-global coverage between $67^\circ\text{S}$ and $67^\circ\text{N}$, they enable a systematic assessment of performance across distinct geographic and climatic regimes. At the same time, two caveats apply to this analysis: first, although evaluating on held-out years avoids direct training-set memorization and yields stable interannual scores (as shown in the preceding section when comparing 2024 and 2025), GPM CORRA remains the training target, so any systematic observational biases shared by GPM and \pardig (such as underestimating light drizzle or snowfall) will not be captured here; second, because CSI is sensitive to the underlying climatology, absolute scores across disparate regimes (e.g., tropics, arid regions, and oceans) should be compared with caution. Even so, this evaluation provides a valuable estimate of \pardig's relative performance across regions where independent ground-based networks are sparse or absent.

\begin{figure}[htbp]
    \centering
    \includegraphics[width=0.82\linewidth]{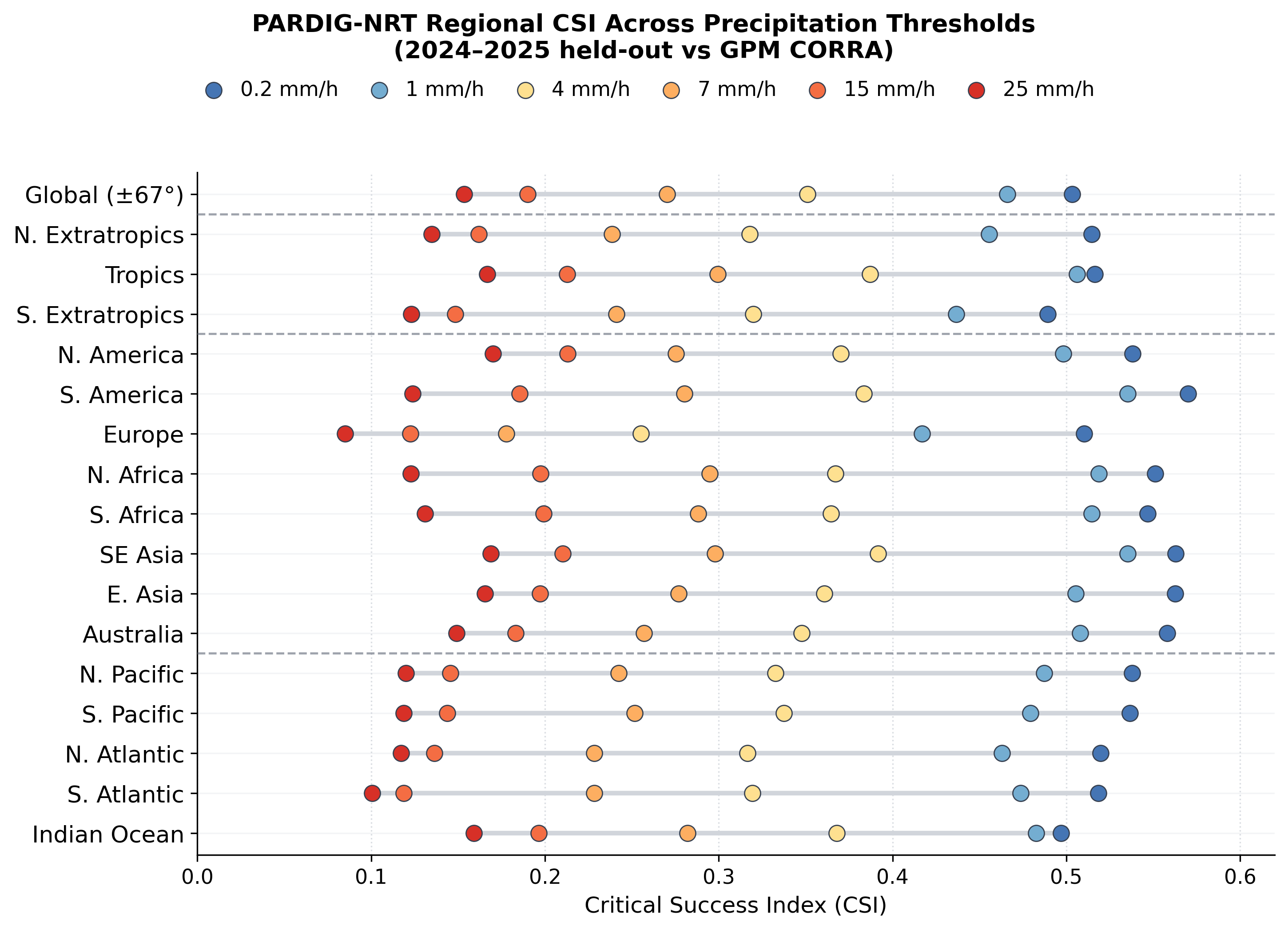}
    \caption{\textbf{PARDIG-NRT regional CSI across precipitation thresholds against held-out GPM CORRA} (2024--2025). Critical Success Index is evaluated across six precipitation thresholds (0.2, 1, 4, 7, 15, and 25\,mm/h) for the global domain and sixteen regional domains organized into three groups (Latitudinal Bands, Continental Regions, and Major Ocean Basins). Numerical values are provided in \Cref{tab:supp_gpm_regional_csi}.}
    \label{fig:supp_gpm_regional}
\end{figure}

\Cref{fig:supp_gpm_regional} and \Cref{tab:supp_gpm_regional_csi} report Critical Success Index (CSI) across six precipitation thresholds (0.2 to 25\,mm/h) for the global domain ($\pm 67^\circ$) and sixteen regional domains organized into three groups: latitudinal bands, continental regions, and major ocean basins:

\begin{enumerate}[leftmargin=*]
    \item \textbf{Latitudinal stratification:} Partitioning the globe into the Northern Extratropics, Tropics, and Southern Extratropics reveals a clear tropical-to-extratropical gradient around the global baseline: retrieval skill peaks in the Tropics and steps down into both extratropical bands. Within the extratropics, the Northern Hemisphere scores slightly higher than the Southern Hemisphere at most thresholds, consistent with its larger landmass fraction and the higher skill observed over continents than over oceans.

    \item \textbf{Continental regions:} Seven of the eight continental domains generally exceed the global benchmark and exhibit remarkably consistent skill across diverse climate zones: all seven fall within a narrow range of $\text{CSI} = 0.50\text{--}0.54$ at 1\,mm/h and $0.18\text{--}0.21$ at 15\,mm/h, spanning both well-instrumented mid-latitude landmasses (North America, East Asia, Australia) and tropical/subtropical continents (South America, Northern and Southern Africa, Southeast Asia). Strong and uniform performance across the latter is of particular practical importance, since applications in these regions rely most heavily on global products for lack of local observational alternatives.

    The sole exception is Europe, which scores below the global average ($\text{CSI} = 0.42$ at 1\,mm/h) and substantially lower than North America and East Asia despite comparable mid-latitude coverage. Crucially, this pattern is not unique to GPM: in the independent ground-radar evaluations (\Cref{tab:csi_all}), \emph{all evaluated products} exhibit a comparable drop over Europe against EURADCLIM relative to MRMS over CONUS and JMA over Japan. While reduced GPM DPR sensitivity to high-latitude light rain and snow, or geostationary sensor differences may contribute, the parallel drop across operational baselines might suggest that European precipitation regimes are intrinsically more challenging across both satellite and NWP systems.

    \item \textbf{Major ocean basins and land--ocean contrast:} Because oceans cover roughly 70\% of the $\pm 67^\circ$ domain, the global aggregate is primarily governed by marine precipitation. Across all five major ocean basins, detection skill at light-to-moderate thresholds clusters tightly near the global mean and sits systematically below the non-European continents. At moderate-to-heavy thresholds ($\ge 4$\,mm/h), the four subtropical-to-extratropical Pacific and Atlantic basins track one another closely, whereas the Indian Ocean, which extends into the deep tropics, retains substantially higher heavy-rain skill, mirroring the tropical-to-extratropical gradient seen in the latitudinal bands.
\end{enumerate}

\begin{table}[htbp]
\centering
\small
\setlength{\tabcolsep}{3.0pt}
\caption{\textbf{PARDIG-NRT regional CSI across precipitation thresholds}. Evaluated against held-out GPM CORRA instantaneous precipitation rate observations (2024--2025), stratified by geographic domain.}
\label{tab:supp_gpm_regional_csi}
\begin{tabular}{@{}llccccccc@{}}
\toprule
Region & Latitude & Longitude & 0.2\,mm/h & 1\,mm/h & 4\,mm/h & 7\,mm/h & 15\,mm/h & 25\,mm/h \\
\midrule
Global ($\pm$67$^\circ$) & 67$^\circ$S--67$^\circ$N & 180$^\circ$W--180$^\circ$E & 0.50 & 0.47 & 0.35 & 0.27 & 0.19 & 0.15 \\
\midrule
\multicolumn{9}{@{}l}{\textbf{(a) Latitudinal Bands}} \\
\midrule
N. Extratropics & 20$^\circ$N--65$^\circ$N & 180$^\circ$W--180$^\circ$E & 0.52 & 0.46 & 0.32 & 0.24 & 0.16 & 0.14 \\
Tropics & 20$^\circ$S--20$^\circ$N & 180$^\circ$W--180$^\circ$E & 0.52 & 0.51 & 0.39 & 0.30 & 0.21 & 0.17 \\
S. Extratropics & 65$^\circ$S--20$^\circ$S & 180$^\circ$W--180$^\circ$E & 0.49 & 0.44 & 0.32 & 0.24 & 0.15 & 0.12 \\
\midrule
\multicolumn{9}{@{}l}{\textbf{(b) Continental Regions}} \\
\midrule
N. America & 25$^\circ$N--60$^\circ$N & 125$^\circ$W--75$^\circ$W & 0.54 & 0.50 & 0.37 & 0.28 & 0.21 & 0.17 \\
S. America & 40$^\circ$S--5$^\circ$N & 75$^\circ$W--40$^\circ$W & 0.57 & 0.54 & 0.38 & 0.28 & 0.19 & 0.12 \\
Europe & 36$^\circ$N--65$^\circ$N & 10$^\circ$W--40$^\circ$E & 0.51 & 0.42 & 0.26 & 0.18 & 0.12 & 0.09 \\
N. Africa & 5$^\circ$N--35$^\circ$N & 15$^\circ$W--40$^\circ$E & 0.55 & 0.52 & 0.37 & 0.30 & 0.20 & 0.12 \\
S. Africa & 35$^\circ$S--5$^\circ$N & 12$^\circ$E--40$^\circ$E & 0.55 & 0.52 & 0.37 & 0.29 & 0.20 & 0.13 \\
SE Asia & 10$^\circ$S--22$^\circ$N & 95$^\circ$E--130$^\circ$E & 0.56 & 0.54 & 0.39 & 0.30 & 0.21 & 0.17 \\
E. Asia & 22$^\circ$N--55$^\circ$N & 100$^\circ$E--142$^\circ$E & 0.56 & 0.51 & 0.36 & 0.28 & 0.20 & 0.17 \\
Australia & 39$^\circ$S--12$^\circ$S & 114$^\circ$E--154$^\circ$E & 0.56 & 0.51 & 0.35 & 0.26 & 0.18 & 0.15 \\
\midrule
\multicolumn{9}{@{}l}{\textbf{(c) Major Ocean Basins}} \\
\midrule
N. Pacific & 20$^\circ$N--55$^\circ$N & 145$^\circ$E--130$^\circ$W & 0.54 & 0.49 & 0.33 & 0.24 & 0.15 & 0.12 \\
S. Pacific & 55$^\circ$S--20$^\circ$S & 180$^\circ$E--75$^\circ$W & 0.54 & 0.48 & 0.34 & 0.25 & 0.14 & 0.12 \\
N. Atlantic & 20$^\circ$N--58$^\circ$N & 55$^\circ$W--15$^\circ$W & 0.52 & 0.46 & 0.32 & 0.23 & 0.14 & 0.12 \\
S. Atlantic & 50$^\circ$S--10$^\circ$S & 35$^\circ$W--10$^\circ$E & 0.52 & 0.47 & 0.32 & 0.23 & 0.12 & 0.10 \\
Indian Ocean & 45$^\circ$S--5$^\circ$N & 50$^\circ$E--95$^\circ$E & 0.50 & 0.48 & 0.37 & 0.28 & 0.20 & 0.16 \\
\bottomrule
\end{tabular}
\end{table}

\FloatBarrier

\end{document}